\documentclass[final]{cvpr}

\usepackage{times}
\usepackage{epsfig}
\usepackage{graphicx}
\usepackage{amsmath}
\usepackage{amssymb}

\usepackage{color}
\usepackage[table]{xcolor}
\usepackage{multirow}
\usepackage{tabularx}
\usepackage{hhline}
\usepackage[para,online,flushleft]{threeparttable}

\usepackage{tikz,pgfplots}
\usepackage{environ}
\usetikzlibrary{shapes}
\usetikzlibrary{arrows,decorations.markings}
\pgfplotsset{compat=1.16}

\usepackage{MnSymbol,wasysym}

\usepackage{cases}
\usepackage{wrapfig}
\usepackage{sidecap}
\usepackage{overpic}
\usepackage{booktabs}
\usepackage{colortbl}
\usepackage[toc,page]{appendix}
\usepackage{placeins}
\usepackage{pifont}

\usepackage{rotating}
\usepackage[export]{adjustbox}

\usepackage[normalem]{ulem}
\usepackage{xspace}
\usepackage{pifont} 

\definecolor{red}{rgb}{0.95,0.4,0.4}
\definecolor{blue}{rgb}{0.4,0.4,0.95}
\definecolor{darkblue}{rgb}{0,0,0.8}
\definecolor{darkred}{rgb}{0.8,0,0}
\definecolor{darkgreen}{rgb}{0.15,0.6,0.15}
\definecolor{grey}{rgb}{0.6,0.6,0.6}
\definecolor{col1}{RGB}{232, 161, 148}
\definecolor{col2}{RGB}{148, 187, 232}

\definecolor{tgre}{rgb}{0.15,0.6,0.15}
\definecolor{tred}{rgb}{0.6,0.15,0.15}

\renewcommand{\paragraph}[1]{\noindent\textbf{#1}}

\newlength\savewidth

\graphicspath{{figs/matterport_results/}{figs/}}

\renewcommand*{\ie}{i.e.\@\xspace}
\renewcommand*{\eg}{e.g.\@\xspace}

\DeclareMathOperator*{\argmin}{argmin}

\newcommand{\teacherD}{\hat{D}_t}
\newcommand{\costvolD}{D_{\text{cv}}}
\newcommand{\posenetwork}{\theta_\text{pose}}
\newcommand{\ournetwork}{\theta_\text{depth}}
\newcommand{\teachernetwork}{\theta_\text{consistency}}

\usepackage[pagebackref=true,breaklinks=true,colorlinks,bookmarks=false]{hyperref}

\def\cvprPaperID{4851} 
\def\confYear{CVPR 2021}
\begin{document}

\makeatletter
\@namedef{ver@everyshi.sty}{}
\makeatother

\title{The Temporal Opportunist: Self-Supervised Multi-Frame Monocular Depth}

\author{Jamie Watson$^{1}$\hspace{15pt}Oisin Mac Aodha$^{2}$\hspace{15pt}Victor Prisacariu$^{1,3}$\hspace{15pt}Gabriel Brostow$^{1,4}$\hspace{15pt}Michael Firman$^{1}$\\$^{1}$Niantic \hspace{30pt} $^{2}$University of Edinburgh   \hspace{30pt}$^{3}$University of Oxford\hspace{30pt}  $^{4}$UCL \\\url{www.github.com/nianticlabs/manydepth}  }


\maketitle

\begin{abstract}
Self-supervised monocular depth estimation networks are trained to predict scene depth using nearby frames as a supervision signal during training.
However, for many applications, sequence information in the form of video frames is also available at test time.
The vast majority of monocular networks do not make use of this extra signal, thus ignoring valuable information that could be used to improve the predicted depth.
Those that do, either use computationally expensive test-time refinement techniques or off-the-shelf recurrent networks, which only indirectly make use of the geometric information that is inherently available.

We propose ManyDepth, an adaptive approach to dense depth estimation that can make use of sequence information at test time, when it is available.
Taking inspiration from multi-view stereo, we propose a deep end-to-end cost volume based approach that is trained using self-supervision only.
We present a novel consistency loss that encourages the network to ignore the cost volume when it is deemed unreliable, \eg in the case of moving objects, and an augmentation scheme to cope with static cameras.
Our detailed experiments on both KITTI and Cityscapes show that we outperform all published self-supervised baselines, including those that use single or multiple frames at test time.

\end{abstract}

\section{Introduction}
Knowing the depth to each pixel in an image has proved to be a useful and versatile tool, with applications ranging from augmented reality \cite{luo2020consistent}, autonomous driving \cite{Geiger2012CVPR}, through to 3D reconstruction \cite{newcombe2011kinectfusion}.
While specialist hardware can give per-pixel depth, \eg from structured light or Lidar sensors, a more attractive approach is to only require a single RGB camera.
Many recent monocular depth from RGB methods are trained using only self-supervision, which removes the need for expensive hardware to capture training depth data \cite{garg2016unsupervised,zhou2017unsupervised,godard2017unsupervised,gordon2019depth}.
While these approaches appear to be very promising, their test-time depth estimation performance is not yet on a par with specialist depth hardware or deep multi-view methods \cite{smolyanskiy2018importance}.

\begin{figure}
    \centering
    \includegraphics[width=1.0\columnwidth]{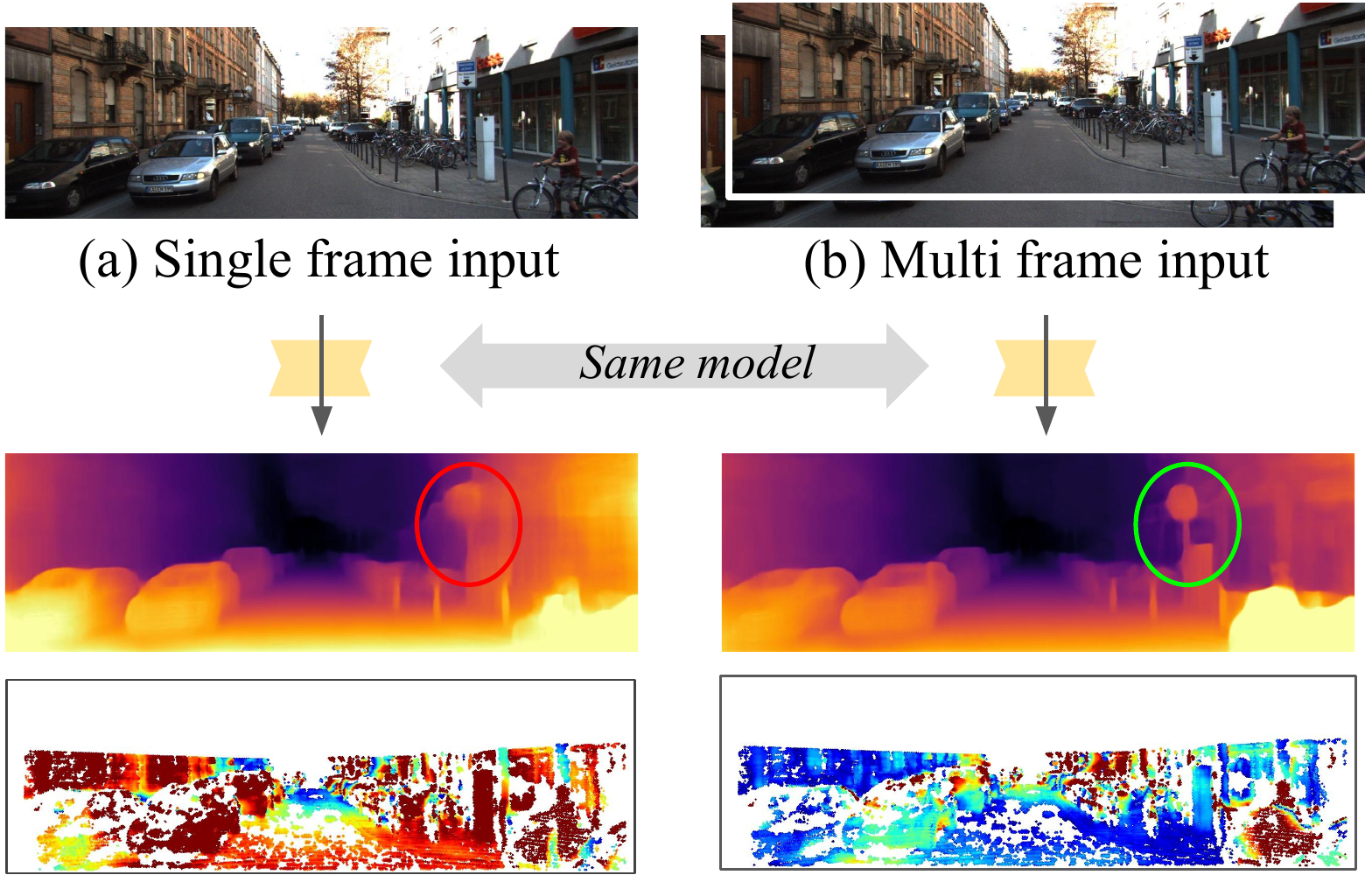}
    \caption{Trained using only self-supervision, our model not only predicts depth from single frames (a) but can also utilize multiple frames, when they are available, using the same model (b).
    This results in superior depth predictions at test time.
    Error maps on the bottom row show large depth errors as red, small as blue. 
    }
    \label{fig:teaser}
    \vspace{-6pt}
\end{figure}

In an attempt to close this performance gap, we observe that in most practical scenarios more than one frame is available at test time \eg from a camera on a moving vehicle or micro-baseline frames from a phone camera \cite{ha2016high,joshi2014micro}.
Yet these additional frames are typically not exploited by current monocular methods.
In this work, we use these additional frames at both training and test time, when they are available, to self-supervise a multi-frame depth estimation system.
We show that a straightforward application of self-supervised training to a multi-view plane-sweep stereo architecture produces poor results, significantly worse than self-supervised single frame networks.
To overcome this, we introduce several innovations to address issues caused by moving objects, scale ambiguity, and static cameras.
We call our resultant multi-frame system \textbf{ManyDepth}.

We make the following three contributions:
\vspace{-6pt}
\begin{enumerate}
    \setlength\itemsep{-4pt}
    \item A novel \emph{self-supervised} multi-frame depth estimation model that combines the strengths of monocular and multi-view depth estimation by making use of multiple frames at test time, when they are available.

    \item We show that moving objects and static scenes significantly impact self-supervised multi-view matching approaches, and we introduce efficient losses and training solutions to alleviate this problem.

    \item We propose an adaptive cost volume to overcome the scale ambiguity arising from self-supervised training on monocular sequences. To the best of our knowledge,  this is the first time cost volume extents have been learned from data rather than set as parameters.

\end{enumerate}

\noindent Our ManyDepth model outperforms existing single and multi-frame approaches on KITTI and Cityscapes. 

\section{Related work}

\subsection{Monocular depth estimation}
The goal of monocular depth estimation is to predict the depth of each pixel in a single input image.
Supervised approaches either make use of dense supervision from depth sensors \eg \cite{eigen2014depth,eigen2015predicting,fu2018deep} or sparse supervision from human annotations \eg \cite{chen2016single}.
Self-supervised methods remove the limitation of requiring ground truth depth supervision, instead training with image-reconstruction losses using stereo pairs \cite{xie2016deep3d,garg2016unsupervised,godard2017unsupervised} or monocular video sequences \cite{zhou2017unsupervised}.

Recent advances in self-supervised training have focused on addressing various challenges resulting from learning from images alone \eg more robust image reconstruction losses \cite{gordon2019depth, shu2020feature}, discrete rather than continuous depth predictions \cite{liu2019neural,gonzalez2020forget,johnston2020self}, feature space reconstruction losses \cite{zhanst2018,shu2020feature}, sparse automatically generated depth supervision \cite{klodt2018supervising,watson2019depthints}, occlusion handling \cite{gordon2019depth}, improved network architectures \cite{guizilini20203d}, and moving objects during training \cite{ranjan2018adversarial,godard2019digging,gordon2019depth,vijayanarasimhan2017sfm,geonet2018,chen2019self,bian2019unsupervised, tosi2020distilled, klingner2020self,li2020unsupervised} and at test time \cite{li2019mannequin}.
Our underlying monocular architecture is based on \cite{godard2019digging}, and could similarly benefit from many of the above enhancements.

\subsection{Multi-frame monocular depth estimation}
There is a growing number of works that extend existing self-supervised monocular models so that they can leverage temporal information at \emph{test time} to improve the quality of the predicted depth.
It is worth noting that there are also several non-deep-learning methods that also aim to produce consistent sequential depth estimates \eg~\cite{zhang2009consistent,karsch2014depth}, in addition to conventional SLAM based methods \cite{newcombe2011dtam,newcombe2010live,engel2014lsd,yang2018polarimetric}, and SLAM methods that integrate a monocular depth estimation network \cite{tateno2017cnn,bloesch2018codeslam,laidlow2019deepfusion}.
However, here we focus on state-of-the-art neural network based depth estimation.

Test-time refinement approaches adapt monocular methods to use sequence information at test time \eg \cite{casser2018depth,chen2019self,luo2020consistent,mccraith2020monocular, shu2020feature,kuznietsov2021comoda}.
As self-supervised training does not require any ground truth depth supervision, the same losses used during training can be applied to the test frames to update the network's parameters.
The downside is that this necessitates the additional computation of multiple forward and backward model passes for a set of test frames, potentially taking several seconds to perform per set \cite{mccraith2020monocular,luo2020consistent}, although one additional backward pass can be effective and fast~\cite{kuznietsov2021comoda}. 

A second broad group of approaches combine traditional monocular networks with recurrent layers to process sequences of frames \eg \cite{patil2020dont, wang2019recurrent,kumar2018depthnet, zhang2019exploiting}.
A related approach uses pairs of sequential frames at test time, sharing features between the pose and depth modules \cite{wang2020self} or computing depth-from-flow \cite{xie2019video}.
These approaches are much more efficient compared to test-time refinement, but they can be more computationally demanding during training due to the need to extract features from multiple frames in a sequence.
A further limitation of these methods is that they do not explicitly reason about geometry during inference;
they simply rely on the network having learned how to extract meaningful  temporal representations.

Our experiments show that our approach often outperforms test-time refinement in terms of accuracy while retaining the efficiency of recurrent methods at inference.

\begin{table}
    \centering
    \footnotesize
    \resizebox{1.0\columnwidth}{!}{
    \begin{tabular}{|l|c|c||c|c|c|c|}
    \cline{2-7}
    \multicolumn{1}{c|}{}  &   \multicolumn{2}{c||}{Train} & \multicolumn{4}{c|}{Test}   \\
    \cline{2-7} 
    \multicolumn{1}{c|}{} & \rotatebox[origin=c]{90}{Needs Depth} & \rotatebox[origin=c]{90}{Needs Pose} & \rotatebox[origin=c]{90}{Needs Pose} & \rotatebox[origin=c]{90}{\hspace{1pt} Object Motion \hspace{1pt}} & \rotatebox[origin=c]{90}{Single Frame} & \rotatebox[origin=c]{90}{Multi Frame}  \\
    \cline{2-7} \hline
    Two Frame \eg \cite{babu2018undemon} & \textcolor{tred}{Yes} & \textcolor{tred}{Yes} & \textcolor{tgre}{No} & \textcolor{tred}{No}  & \textcolor{tred}{No} & \textcolor{tgre}{Yes}  \\
    \hline
    Supervised MVS \eg \cite{huang2018DeepMVS} & \textcolor{tred}{Yes} & \textcolor{tred}{Yes} & \textcolor{tred}{Yes} & \textcolor{tred}{No} & \textcolor{tred}{No} & \textcolor{tgre}{Yes}  \\
    \hline
    Self-sup MVS \eg \cite{khot2019learning}& \textcolor{tgre}{No} & \textcolor{tred}{Yes} & \textcolor{tred}{Yes} & \textcolor{tred}{No} & \textcolor{tred}{No} & \textcolor{tgre}{Yes}  \\
    \hline
    Supervised MD \eg \cite{fu2018deep}& \textcolor{tred}{Yes} & \textcolor{tgre}{No} & \textcolor{tgre}{No} & \textcolor{tgre}{Yes} & \textcolor{tgre}{Yes} & \textcolor{tred}{No}  \\
    \hline
    Self-sup MD \eg \cite{zhou2017unsupervised}& \textcolor{tgre}{No} & \textcolor{tgre}{No} & \textcolor{tgre}{No} & \textcolor{tgre}{Yes} & \textcolor{tgre}{Yes} & \textcolor{tred}{No}  \\
    \hline \hline
    {\bf ManyDepth (Ours)}  & \textcolor{tgre}{\bf  No} & \textcolor{tgre}{\bf No} & \textcolor{tgre}{\bf No} & \textcolor{tgre}{\bf Yes} & \textcolor{tgre}{\bf Yes} & \textcolor{tgre}{\bf Yes}  \\
    \hline
    \end{tabular}}
    \vspace{-2pt}
    \caption{{Comparison of existing approaches that estimate depth from collections of images.}
    Our approach requires no ground-truth supervision and is robust to object motion.
    MVS stands for Multi-View Stereo, and MD stands for Monocular Depth.
    \label{tab:method_comparison}}
    \vspace{-2pt}
\end{table}

\subsection{Deep multi-view depth estimation}

Our problem of predicting depth from multiple frames is related to multi-view depth estimation.
While early deep stereo methods used mostly convolutional layers to map from images pairs to depth using ground-truth supervision \eg \cite{mayer2015large,ummenhofer2017demon,liang2018learning},  \cite{kendall2017end} showed that integrating a plane-sweep stereo cost volume significantly improved results.
Recent approaches improved the underlying architectures and contributed more effective ways of regularizing the cost volume \cite{chang2018pyramid,Zhang2019GANet,zhang2019domaininvariant,cheng2019learning}.
It is also possible to train stereo networks without ground truth supervision  \cite{zhong2017self,wang2019unos,aleotti2020reversing,huang2020m3vsnet}, but these models are typically outperformed by supervised variants.
Some works fuse conventional matching-based stereo estimation with monocular depth cues \cite{saxena2007depth, martins2018fusion, facil2017single}.
In contrast, we do not require stereo pairs during training or testing.

A more general version of the stereo-matching problem is multi-view stereo (MVS), which operates on unordered image collections.
Early deep MVS methods used memory-expensive 3D grid representations \eg \cite{ji2017surfacenet,kar2017learning}.
Current supervised approaches, \eg~\cite{huang2018DeepMVS,wang2018mvdepthnet,yao2018mvsnet,im2019dpsnet,long2020occlusion}, utilize cost volumes but assume ground truth depth and camera poses are available for training.
They often require camera poses at test time too; which can be refined from an initial estimate \cite{wei2019deepsfm,liu2019neural}.
Some methods can predict pose at test time, but they still need to be trained with supervision \eg \cite{ummenhofer2017demon}.

Similar to our approach, \cite{liu2019neural,hou2019multi} process sequences of frames at test time using cost volumes.
However, by using ground-truth depth supervision and provided camera poses they side-step the challenges associated with training from self-supervision alone.
\cite{wu2019spatial} predict depth from triplets of frames without requiring pose information, but their method cannot deal with variable numbers of frames at test time and is trained with ground truth depth.
Concurrent with our work, \cite{wimbauer2020monorec} learn depth from a cost volume, but they use stereo pairs and sparse supervised depth at training time and long sequences for pose estimation. 

Most related to us are self-supervised MVS methods that also do not require any ground truth depth \ie \cite{khot2019learning,dai2019mvs2}.
However, there are several reasons why these existing self-supervised and supervised MVS methods aren't applicable in many scenarios:
    (i) they need more than one input image at test time,
    (ii) they assume that the camera is not static,
    (iii) they typically require camera poses to be provided during training and sometimes also at test time, and
    (iv) they assume that there are no moving objects in the scene.
Our approach leverages the best of monocular and multi-view methods by making use of sequence information at test time, when it is available, while also being robust to scene motion --- see Table~\ref{tab:method_comparison}.

\begin{figure*}[!t]
    \centering
    \includegraphics[width=1.0\textwidth]{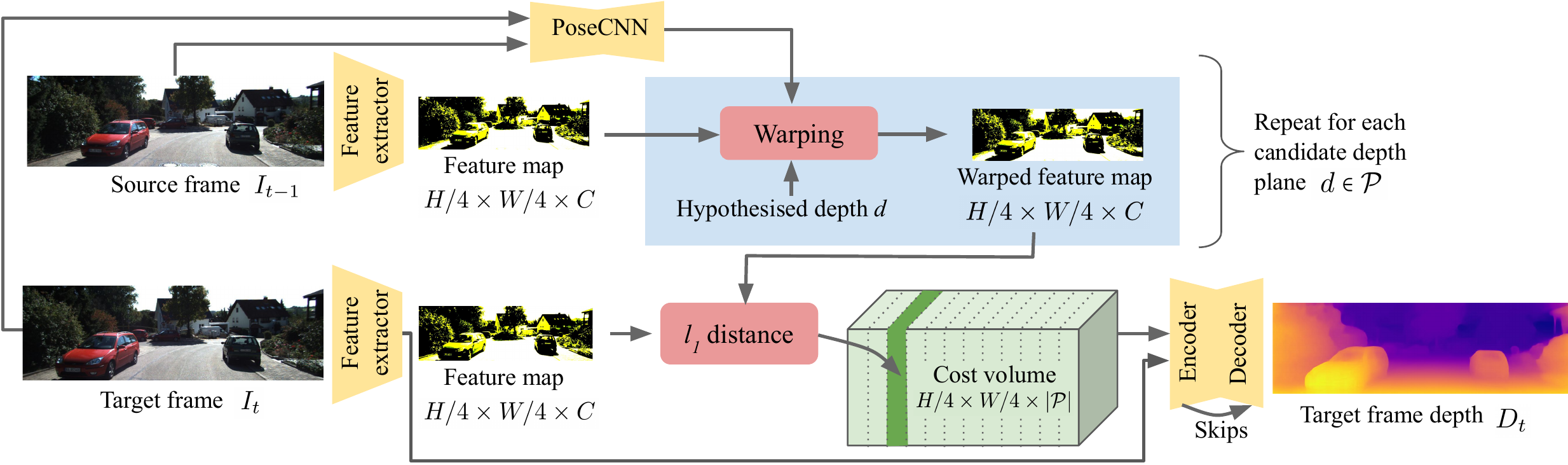}
    \caption{\textbf{Our cost volume based depth estimator.} Our depth network $\ournetwork$ has three main components: a feature extractor, an encoder, and a depth decoder.
    Our pose network $\posenetwork$ estimates the relative pose between pairs of images, which is then used to build a cost volume in the reference frame of the target image $I_t$  by warping features extracted from images at different time points.
    The encoder and depth decoder processes the cost volume to produce a depth image for $I_t$.
    \label{fig:cost_volume_building}}
    \vspace{-6pt}
\end{figure*}

\section{Problem setup}

The aim of depth estimation is to predict a depth map $D_t$, pixel-aligned with an input image $I_t$.
Conventional single-image depth estimation methods, \eg~\cite{eigen2015predicting,godard2017unsupervised,zhou2017unsupervised}, train a deep network $\ournetwork$ to map from $I_t$ to $D_t$,
\begin{align}
    D_t = \ournetwork(I_t).
\end{align}
In contrast, like other \emph{multi-frame} methods \cite{patil2020dont, wang2019recurrent}, our model accepts as input $N$ previous temporal video frames,
\begin{align}
    D_t = \ournetwork(I_t, I_{t-1}, \ldots,  I_{t-N}).
\end{align}
While our model makes use of information from multiple frames, it also can operate in the regime where only one frame is available at test time. 
Unlike similar works \eg \cite{kumar2018depthnet,zhang2019exploiting, chen2019self,luo2020consistent,casser2018depth}, we do not use \emph{future} frames at test time \eg $I_{t+1}$, the use of which would preclude online applications. 
At training time, we exploit previous \emph{and} future frames as a supervisory signal, and do not require stereo supervision.  
We do not assume access to the true relative camera pose between $I_t$ and the preceding frames; instead we learn to predict these poses $\{T_n\}_{n=1}^N$ at training and test time with a differentiable pose network $\posenetwork$, following \cite{zhou2017unsupervised}.
We also do not make use of any trained semantic models to mask moving objects \eg \cite{casser2018depth,gordon2019depth,guizilini2020semantically}.
We do, however, assume known fixed camera intrinsics $K$ --- though we could relax these requirements \cite{gordon2019depth, facil2019cam}.

\section{Method}
\label{sec:method}

Our method starts with two well established components: self-supervised reprojection based training, and a multi-view cost volume.
We then introduce three important innovations that enable cost volume matching to work with self-supervised training from monocular video:
    (1) adaptive cost volumes,
    (2) a method to prevent a failure mode we refer to as `cost volume overfitting', and
    (3) augmentation for static cameras and single frame inputs.

\subsection{Self-supervised monocular depth estimation}
\label{ref:subsec_reproj}

Following~\cite{zhou2017unsupervised}, we train a self-supervision depth network using only video frames that are temporally close to $I_t$.
We use the current estimated depth $D_t$ and the pose network $\posenetwork$'s estimate of relative camera pose $T_{t \to t+n}$ to synthesize the scene from the same viewpoint as $I_t$, but only using pixels from neighboring source frames \ie $\{ I_{t + n}, n \in \{-1, 1\} \, \}$.
The synthesized counterpart to $I_t$ is
\begin{align}
I_{t+n \to t} &= I_{t+n}\Big\langle \text{proj}(D_t, T_{t \to t+n}, K) \Big\rangle,
\end{align}
where $\langle\rangle$ is the sampling operator and $\text{proj}$ returns the 2D coordinates of the depths in $D_t$ when reprojected into the camera of $I_{t+n}$.
Note that while our cost volume, described later, only uses \emph{preceding} frames to enable online applications, at training-time our reprojection loss uses future frames too.
Following \cite{godard2019digging}, for each pixel we optimize the loss for the \emph{best matching source image}, by selecting the per pixel minimum  over the reconstruction loss $pe$,
\begin{align}
    L_p &= \min_{n} pe(I_t, I_{t+n \to t}).
    \label{eqn:reconstruction}
\end{align}
We set $pe$ as a combination of SSIM and $L_1$ losses, and we minimize this loss over all the pixels in the training images over four output scales; see \cite{godard2019digging} for more details.

\subsection{Building a cost volume}

To exploit multiple \emph{input} frames, inspired by \cite{collins1996space,kang2001handling,kendall2017end} we build a cost volume which measures the geometric compatibility at different depth values between the pixels from $I_t$ and nearby source frames from the input video.
This requires knowledge of relative pose $T$, which we estimate with the pose network $\posenetwork$, trained using a reprojection loss. 
We define a set of ordered planes $\mathcal{P}$, each perpendicular to the optical axis at $I_t$ and with depths linearly spaced between $d_{\min}$ and $d_{\max}$.
Each frame is encoded into a deep feature map $F_t$ and warped to the viewpoint of $I_t$ using each of the hypothesised alternative depths $d \in \mathcal{P}$ using the known camera intrinsics and estimated pose. This creates a warped feature map $F_{t+n \to t, d}$.
The final cost volume is constructed as the absolute difference between the warped features and the features from $I_t$, at each $d \in \mathcal{P}$.
This is averaged over all source images, following \cite{newcombe2011dtam}.
The cost volume effectively says `for pixel $(i, j)$, what is the likelihood of the correct depth being $d$, for each $d$ in $\mathcal{P}$?'.
Following \cite{fischer2015flownet}, the cost volume is concatenated with features $F_t$ and used as input to a convolutional decoder which regresses the depth $D_t$.
See Fig.~\ref{fig:cost_volume_building} for an overview.

Cost volumes have the benefit of allowing the network to leverage inputs from multiple viewing angles.
However, they typically require $d_{\min}$ and $d_{\max}$ to be chosen as hyperparameters, and they assume that world is static.
In the following sections we show how to relax these assumptions.

\subsection{Adaptive cost volumes}
Cost volume approaches have a problem of needing a known depth range \ie $d_{\min}$ and $d_{\max}$.
These are typically selected as hyperparameters in advance of training based on prior knowledge of the dataset \cite{dai2019mvs2} or from known camera poses \cite{huang2018DeepMVS}.
We are unable to do this, as self-supervised depth estimation trained on monocular sequences only estimates depth `up to scale'.
This means that while we assume that the final predicted depths, and corresponding poses from the pose network, will all end up in broad agreement with \emph{each other}, they will be different from real-world depth by an unknown scaling factor.

To solve this problem we introduce a novel \emph{adaptive cost volume}, by allowing $d_{\min}$ and $d_{\max}$ to be learned from the data, so they can adjust during training as the network finds its own scaling.
This is done using the current predictions from the network of $D_t$, whereby we compute the average min and max of each $D_t$ over a training batch. These are then used to update an exponential moving average estimates of $d_{\min}$ and $ d_{\max}$ with momentum $0.99$.
$d_{\min}$ and $d_{\max}$ are saved along with the model weights and then kept fixed at test time.
Our approach contrasts to \cite{gu2020cascade} who adapt $d_{\min}, d_{\max}$ at \emph{test time} in a coarse-to-fine manner.

\subsection{Addressing cost volume overfitting}

\begin{figure}
  \centering
 \footnotesize
\includegraphics[width=\columnwidth]{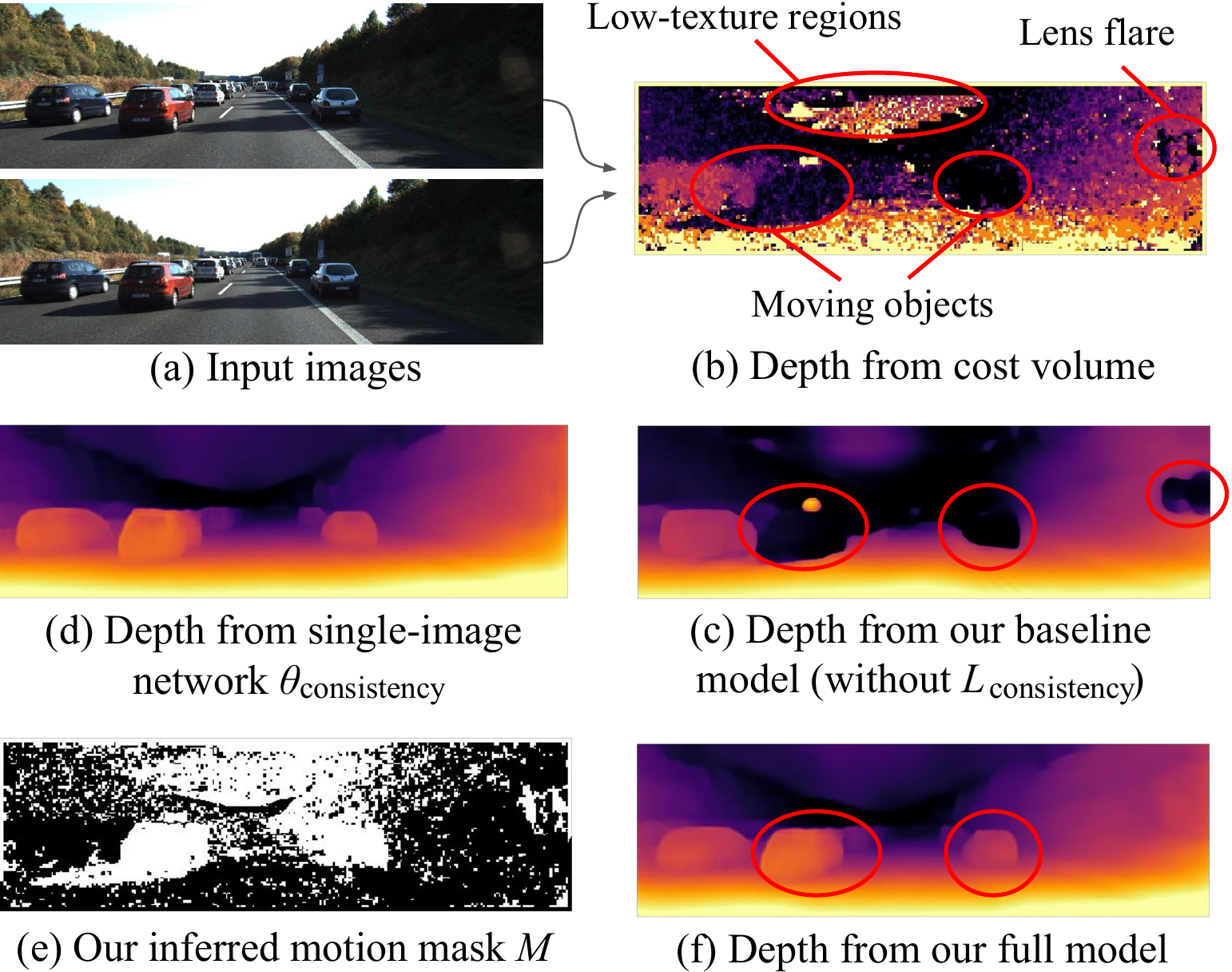}
  \vspace{-8pt}
  \caption{\textbf{We enable self-supervised training to work with a cost volume.}
  When we build a cost volume from a sequence (a), moving objects create incorrect depths in the cost volume (b).
  These errors are propagated through to our predicted depth maps (c).
  `Traditional' single-image depth estimation does not exhibit this failure mode, but can still produce biased depth (d).
  We therefore use a single-image network to help our model recover from this failure mode, but only on pixels identified to be unreliable by our mask $M$ (e).
  Our final prediction (f) is superior to both our baseline and the single-image network. 
  \label{fig:cost_volume_reliable}}
  \vspace{-6pt}
\end{figure}

We observe that our baseline cost volume model trained with monocular supervision suffers from severe artefacts, including large `holes' punched on moving objects.
These are similar to artefacts observed in monocular $I_t \rightarrow D_t$ models (see \cite{casser2018depth, godard2019digging} for a  description).
However, in our cost volume network they are far more severe (see Fig.~\ref{fig:cost_volume_reliable}~(c)).

\vspace{5pt}
\paragraph{Why does the monocular-trained cost volume fail?}
In theory, our model should do well. 
It is trained with a similar reprojection loss used to train state-of-the-art single-frame depth estimators, but it also has access to an additional source of information via the cost volume.
However, the information contained in the cost volume is only reliable in specific scenarios \eg in static regions with textured surfaces.
In regions where objects are moving, or where surfaces are untextured, the cost volume will be an unreliable source of depth information (Fig.~\ref{fig:cost_volume_reliable}~(b)).
For example, the moving car in Fig.~\ref{fig:cost_volume_reliable} results in a match in the cost volume at an incorrect depth and corresponds to a very low reprojection loss. 
During training, the network becomes over-reliant on the cost volume. 
Instead of ignoring the cost volume around moving objects, it trusts it too much.
Artefacts in the cost volume from moving objects are then introduced in the final depth map, at both training and test time.
Ultimately, the final predicted depths \emph{inherit the cost volume's mistakes}.
We introduce a method to correct this during training, by teaching the network not to trust the cost volume in these unreliable regions.

\vspace{5pt}
\paragraph{Using a separate network to regularize.}
We make the observation that \emph{single-image} depth networks do not have a cost volume, so are unaffected by `cost volume overfitting'.
While moving objects can still be a problem for these methods during training \cite{casser2018depth, godard2019digging, ranjan2018adversarial}, in general they make far less severe mistakes on moving objects.
We therefore use a monocular network at training time to help `teach' our cost volume network the right answer --- but only in regions we suspect the cost volume to be problematic.
This separate network $\teachernetwork$ produces a depth map $\teacherD$ for each training image, and is discarded after training.
$\teachernetwork$ shares $\posenetwork$ with our main network to help ensure scale-consistent predictions between $\ournetwork$ and $\teachernetwork$.
Potentially problematic pixels in our multi-frame output are identified by a binary mask $M$.
In these masked regions we apply an $L_1$ loss on $D_t$, encouraging the predictions to be similar to $\teacherD$,
\begin{align}
    L_\text{consistency} &= \sum M \, | D_t - \teacherD |.
\end{align}
Gradients to $\teacherD$ are blocked, ensuring knowledge only transfers from teacher to student and not vice versa.

\vspace{5pt}
\paragraph{Identifying unreliable pixels.}
Our binary mask $M$ is $1$ in regions considered to be unreliable, and $0$ otherwise.
To generate this mask we again make use of $\teacherD$.
We reason that in regions where the cost volume is \emph{reliable}, the depth represented by $\teacherD$ will be similar to the depths represented by the $\argmin$ of the cost volume.
We therefore compare the depth represented by the $\argmin$ of the cost volume (\ie $\costvolD$, not $D_t$) to the depth $\teacherD$ predicted by $\teachernetwork$.
The mask $M$ is set to $1$ only in regions where $\teacherD$ and $\costvolD$ differ significantly, so
\begin{align}
    M &= \max
    \Big(
        \frac{\costvolD - \teacherD}{\teacherD}, \frac{\teacherD - \costvolD}{\costvolD}
    \Big) > 1.
\end{align}

The idea of using a separate `disposable' network to help to regularize training is not new, \eg~\cite{ranjan2018adversarial, zhou2017unsupervised}. 
Our novelty is in using a \emph{single-frame} depth network to improve a \emph{multi-frame} system.
Our approach is also less costly and less constrained than using offline semantic segmentation \cite{gordon2019depth}, and makes fewer assumptions than RANSAC-based filtering \cite{guizilini2020semantically}.
In our experiments we compare to two alternative masking schemes from \cite{ranjan2018adversarial} and \cite{godard2019digging}, and show that our approach is superior.

\begin{figure*}[!t]
    \centering
    \resizebox{1.0\textwidth}{!}{\newcommand{\turnwidthnew}{0.6\columnwidth}

\hspace{-10pt}
\centering


\newcommand{\imlabel}[2]{\includegraphics[width=\turnwidthnew]{#1}%
\raisebox{35pt}{\makebox[2pt][r]{\small #2}}}


\begin{tabular}{@{\hskip 2mm}c@{\hskip 2mm}c@{\hskip 2mm}c@{\hskip 2mm}c@{\hskip 2mm}c@{}}

{\rotatebox{90}{\hspace{1mm}\footnotesize Input \& GT}} &
\includegraphics[width=\turnwidthnew]{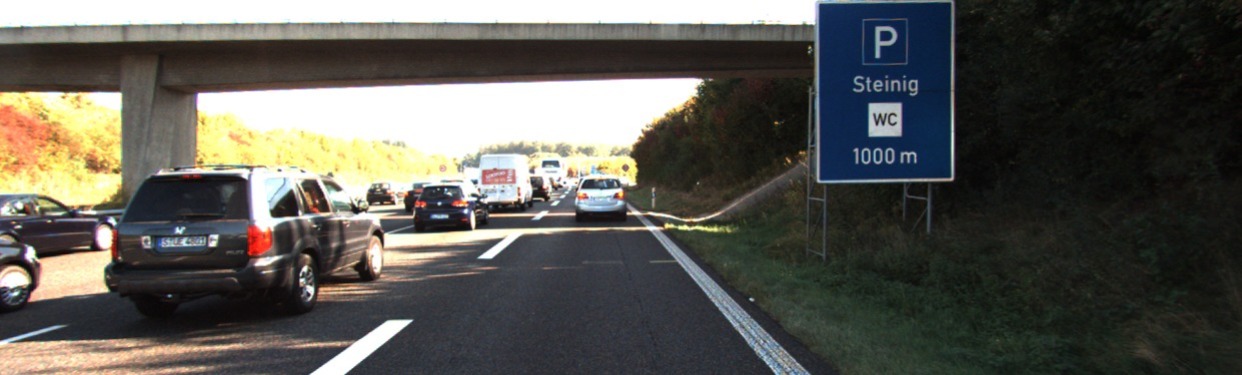} &
\imlabel{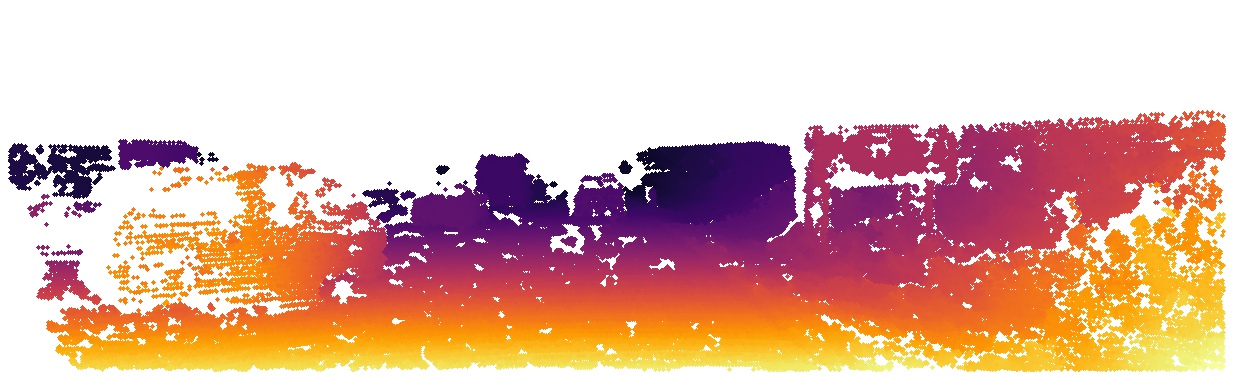}{Ground truth depth} &
\includegraphics[width=\turnwidthnew]{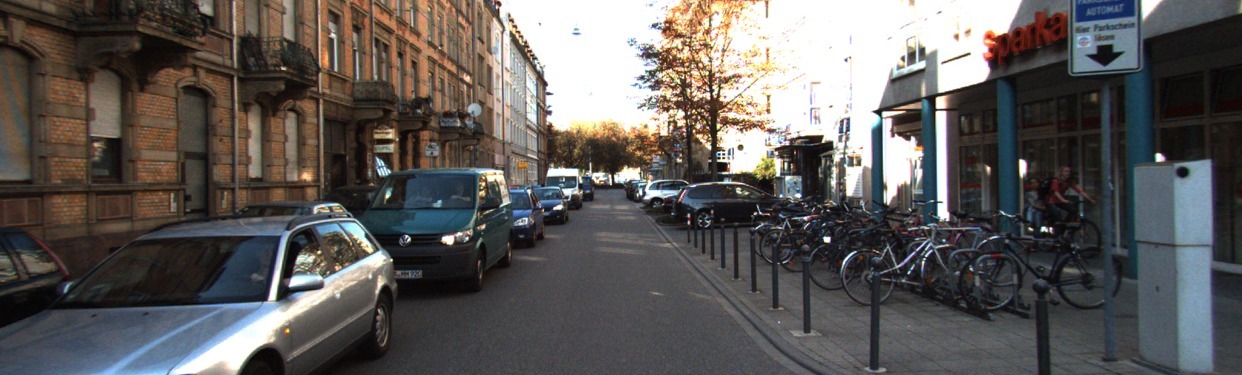} &
\imlabel{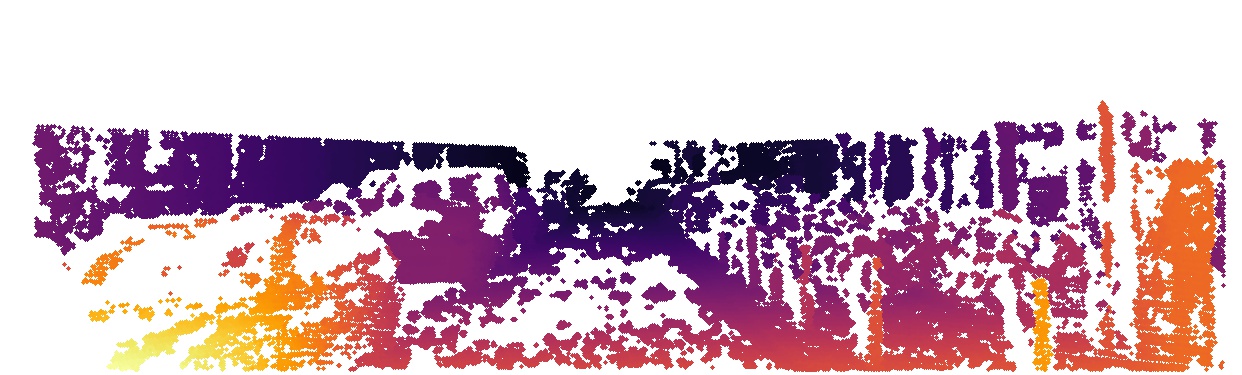}{Ground truth depth} \\

{\rotatebox{90}{\hspace{4mm}\footnotesize \cite{godard2019digging} MR}} &
\includegraphics[width=\turnwidthnew]{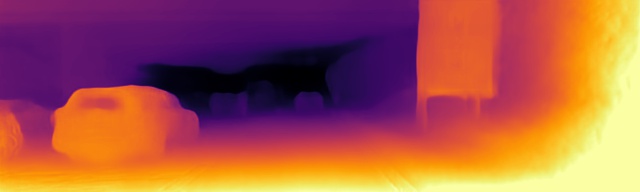} &
\imlabel{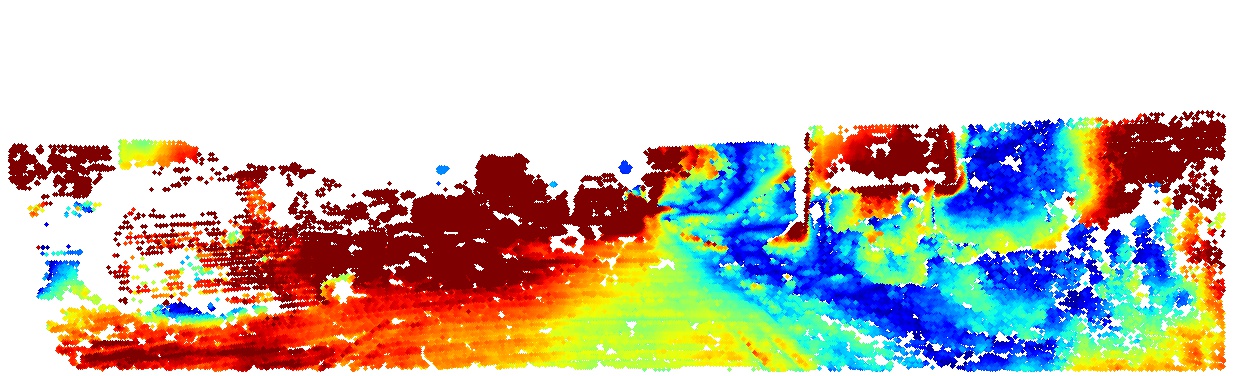}{Error map} &
\includegraphics[width=\turnwidthnew]{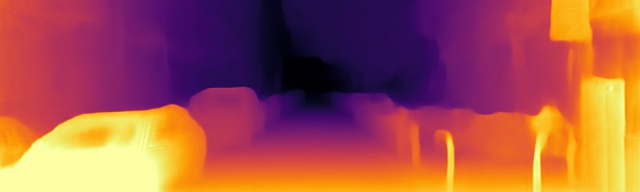} &
\imlabel{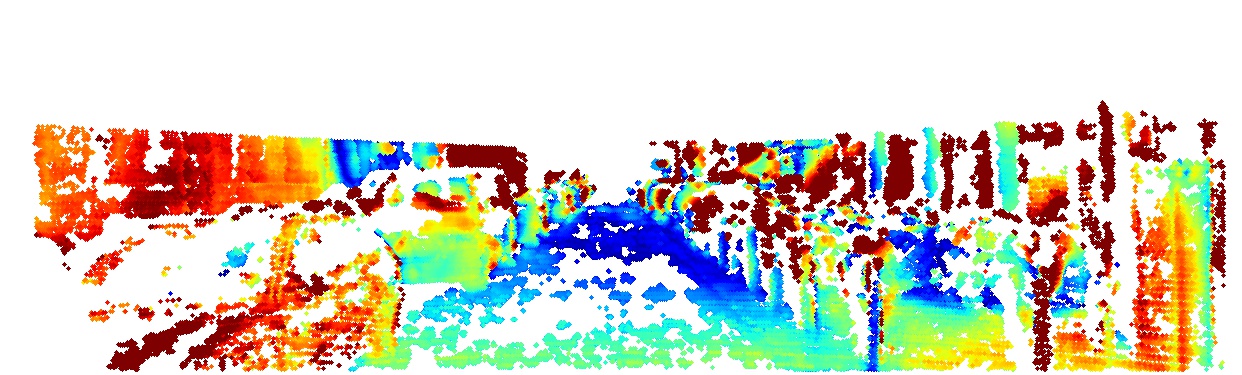}{Error map}  \\

{\rotatebox{90}{\hspace{4mm} \cite{guizilini20203d}}} &
\includegraphics[width=\turnwidthnew]{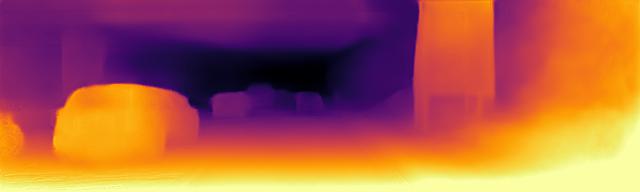} &
\includegraphics[width=\turnwidthnew]{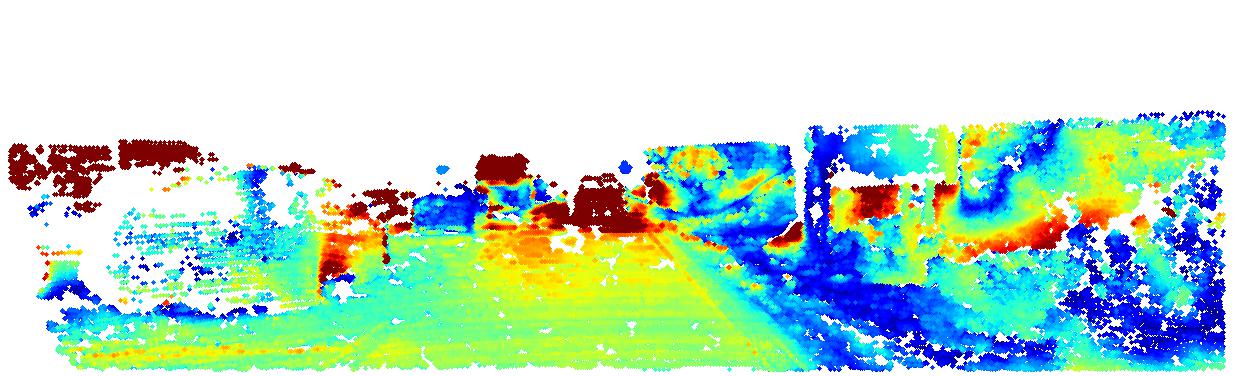} &
\includegraphics[width=\turnwidthnew]{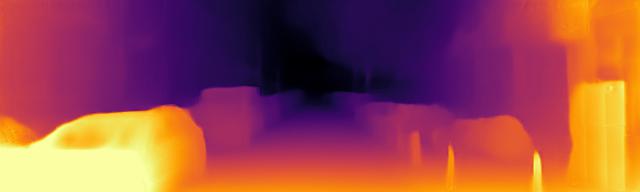} &
\includegraphics[width=\turnwidthnew]{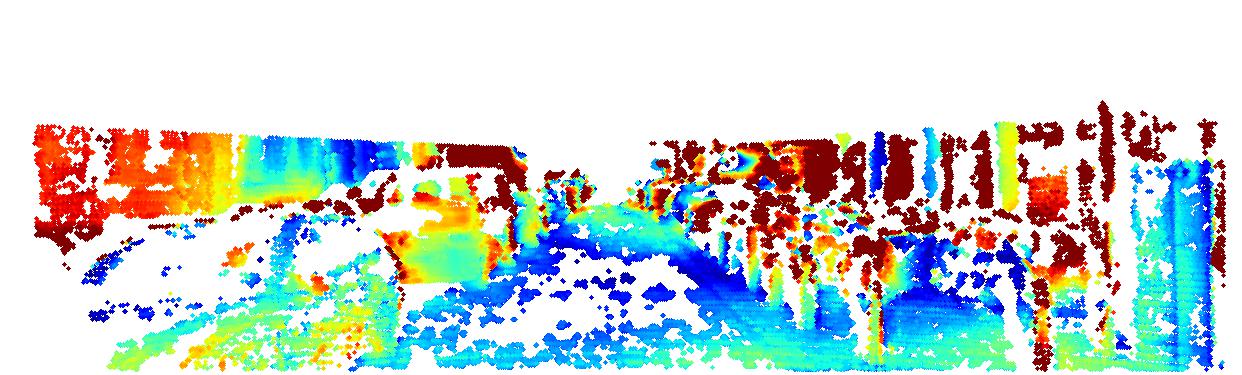}\\

{\rotatebox{90}{\hspace{4mm}\cite{patil2020dont}}} &
\includegraphics[width=\turnwidthnew]{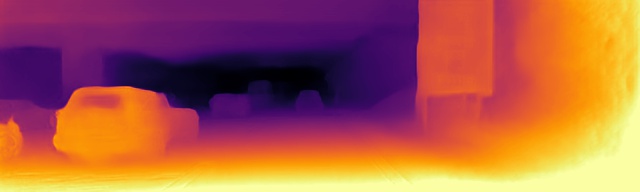} &
\includegraphics[width=\turnwidthnew]{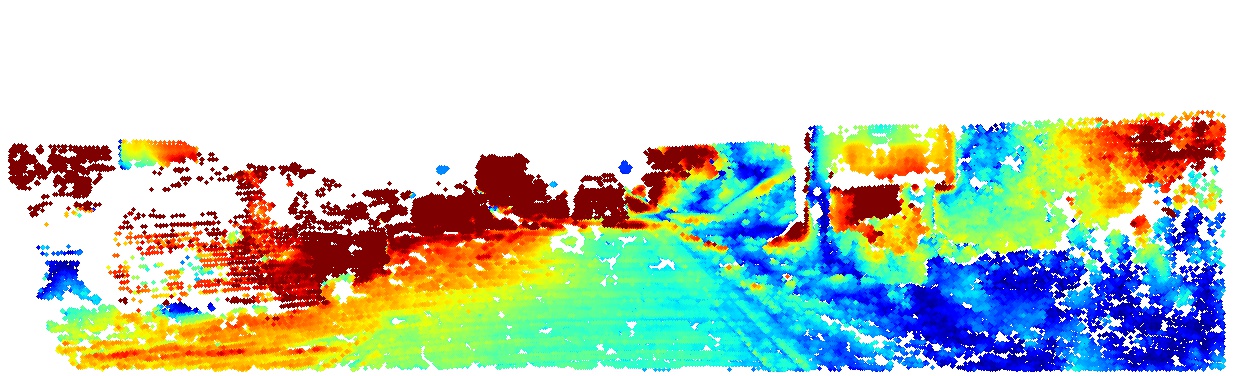} &
\includegraphics[width=\turnwidthnew]{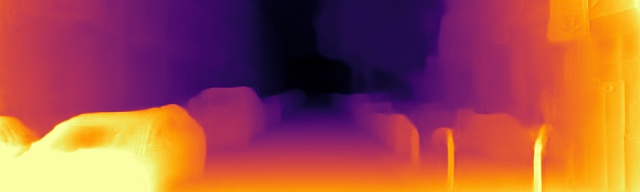} &
\includegraphics[width=\turnwidthnew]{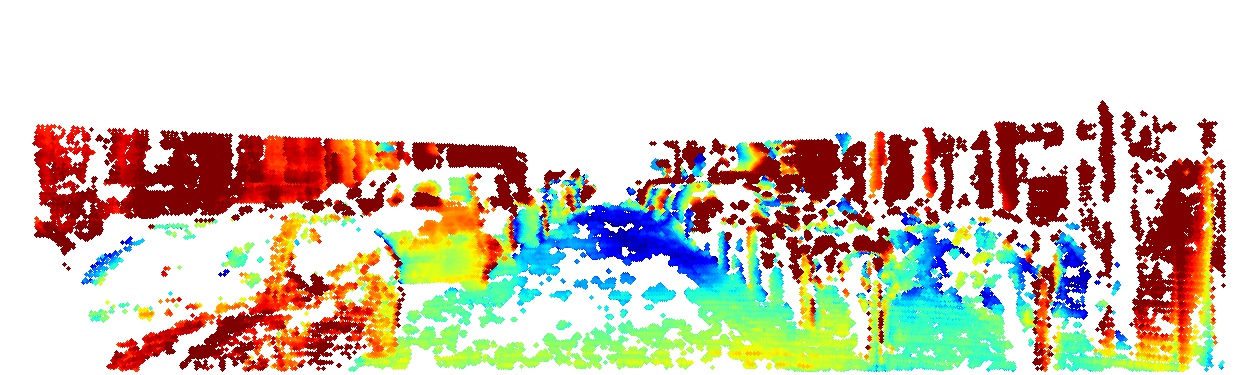}\\

{\rotatebox{90}{\hspace{2mm}\footnotesize Ours MR}} &
\includegraphics[width=\turnwidthnew]{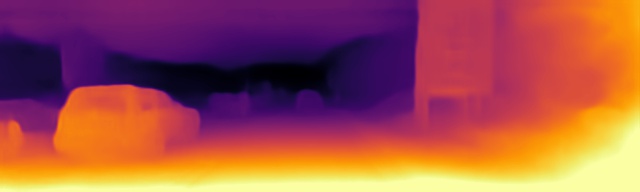} &
\includegraphics[width=\turnwidthnew]{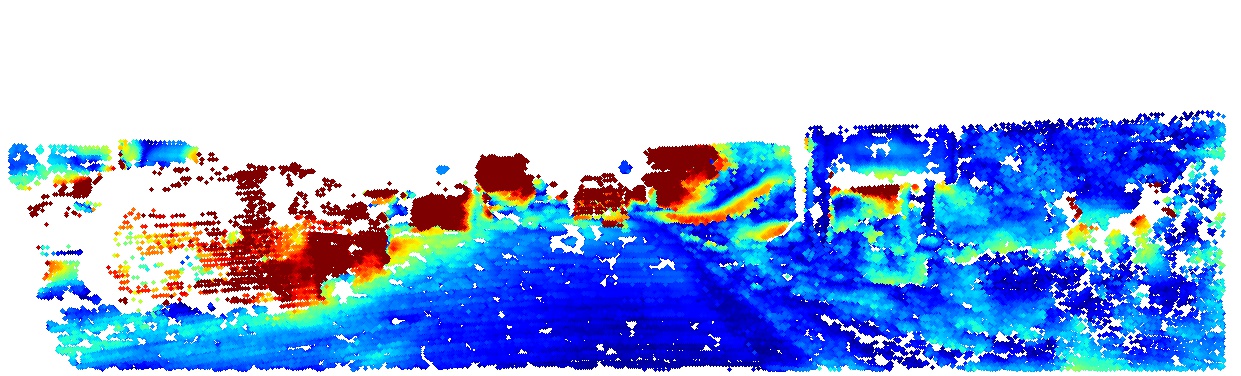} &
\includegraphics[width=\turnwidthnew]{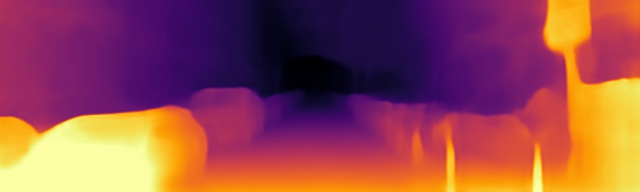} &
\includegraphics[width=\turnwidthnew]{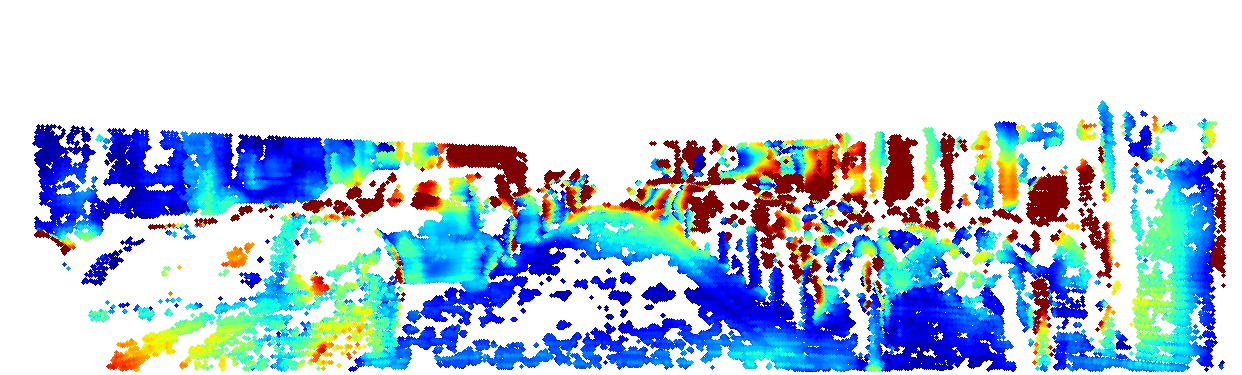}\\

{\rotatebox{90}{\hspace{1mm}\footnotesize Ours + TTR}} &
\includegraphics[width=\turnwidthnew]{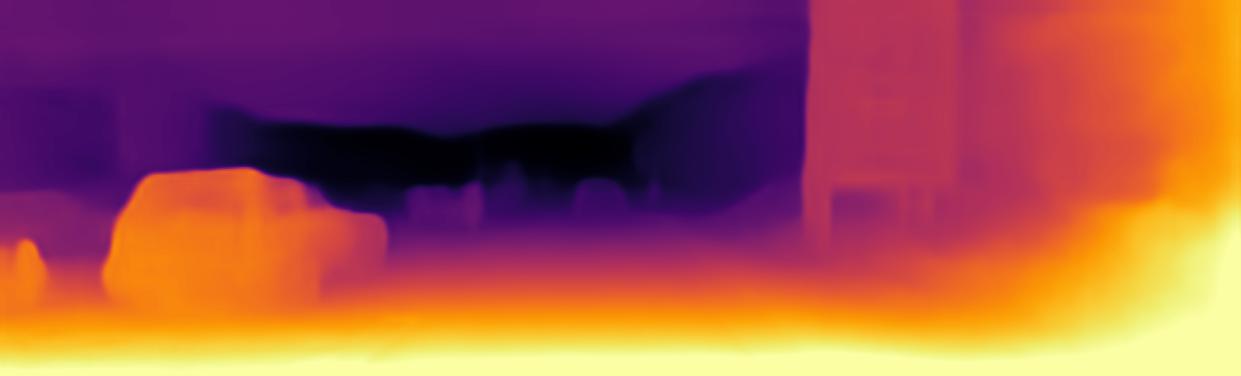} &
\includegraphics[width=\turnwidthnew]{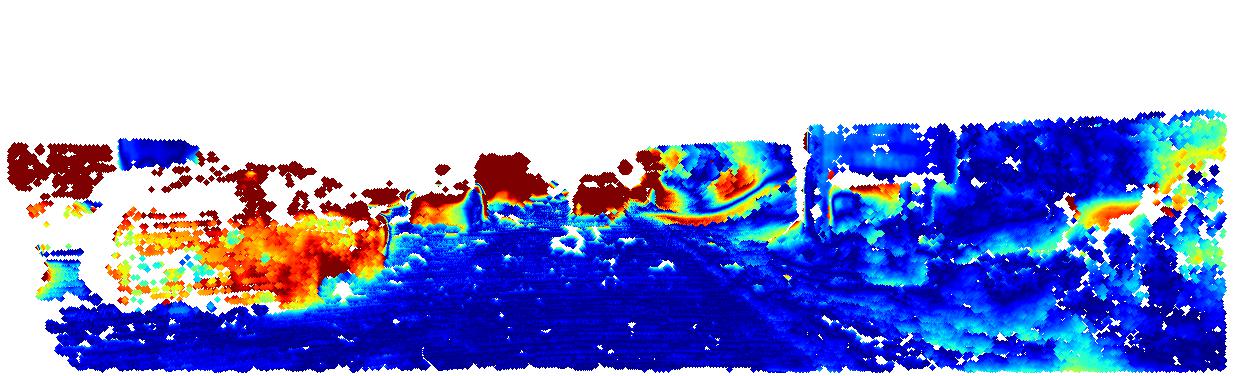} &
\includegraphics[width=\turnwidthnew]{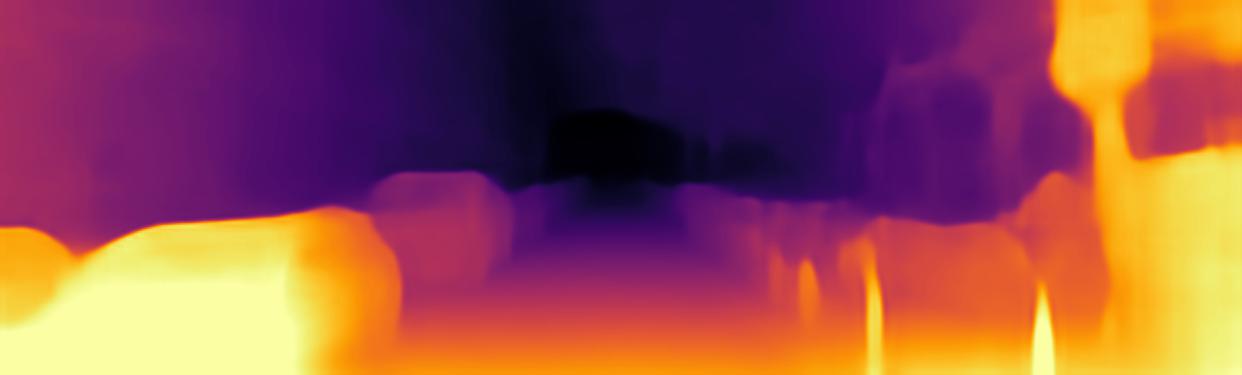} &
\includegraphics[width=\turnwidthnew]{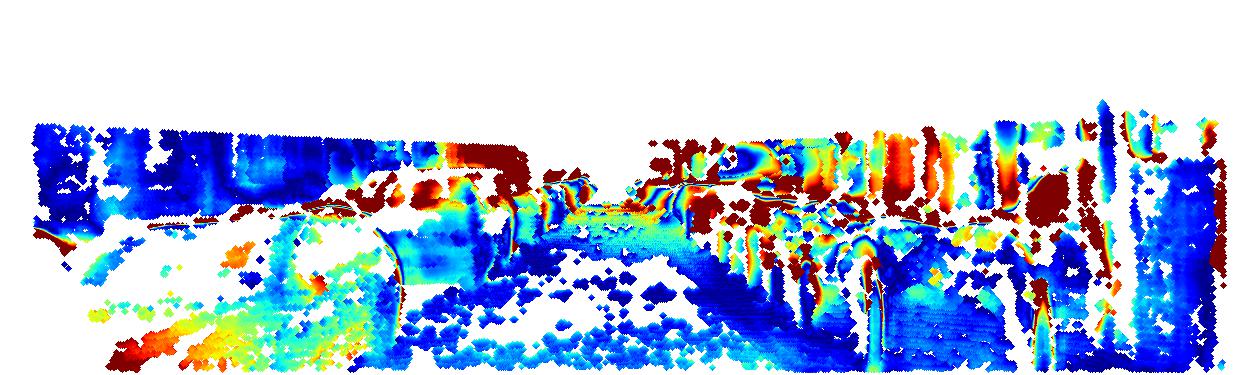}\\

\end{tabular}
}
    \vspace{2pt} 
    \caption{\textbf{Qualitative results on KITTI.} Error maps in columns 2 and 4 show the abs.~rel.~error compared to new ground truth \cite{uhrig2017sparse}, from good (blue) to bad (red). All error maps are colormapped equivalently. While depth maps look qualitatively similar between our multi-frame predictions (bottom rows) and the baselines, the error maps reveal the large `hidden mistakes' made by methods which only have access to a single test-time image. This is particularly apparent in ambiguous regions, \eg in the dark embankment on the right of the freeway.
    Additionally, note that \cite{patil2020dont} also has access to multiple frames at test time, however their method does not explicitly utilize geometry. This results in an improvement over the single frame \cite{godard2019digging}, but there is noticeably higher error than for our approach.   
    \label{fig:qualitative_results_kitti}}
    \vspace{-5pt}
\end{figure*}

\subsection{Static cameras and start-of-sequences}
\label{sec:augmentations}

Using multiple frames at test time introduces two potential challenges for our method.  
The first issue is when $I_{t-1}$ does not exist, \ie when predicting depth for a single image or those at the start of a sequence.
This case is trivially handled by monocular methods as they only require single frames as input. 
However, MVS approaches fail in these situations.
To address this problem, during training with probability $p$, we replace the cost volume with a tensor of zeros.
For these images, this encourages the network to learn to rely only on features directly from $I_t$.
Then at test time, when $I_{t-1}$ does not exist, we simply replace the cost volume with zeros.
The second case arises when the camera does not move between $I_{t-1}$ and $I_t$, \eg a car stopped at traffic lights.
Again this is another failure case for MVS methods.
To address this at training time, with probability $q$, we replace the $I_{t-1}$ input to the cost volume with a color-augmented version of $I_t$, but still supervise with the `real' $I_{t-1}, I_{t+1}$ in Eqn.~\ref{eqn:reconstruction}.
This enables the network to predict plausible depths even when the cost volume is constructed from images with no camera baseline.

\vspace{3pt}

Our final loss is $L = (1 - M)L_p + L_\text{consistency}+ L_{\text{smooth}}$, where $L_{\text{smooth}}$ is the smoothness loss from \cite{godard2017unsupervised}.

\section{Implementation details}
We use training-time color and flip augmentations on images being fed to the depth and pose networks, using the settings from~\cite{godard2019digging}.
Unless otherwise stated, all our models are trained with an input and output resolution of $640 \times 192$, and we fix $N=1$, so the cost volume is constructed with frames $\{I_t, I_{t-1}\}$, at both training and test time.
In all cases self-supervision during training is from frames $\{I_{t-1},I_t, I_{t+1}\}$.
We train with Adam \cite{kingma2014adam} for 20 epochs with a learning rate of $10^{-4}$, dropping by a factor of 10 for the final 5 epochs. 
After $Q$ epochs, we fix $d_{min}$ and $d_{max}$ and the weights of $\posenetwork$ and $\teachernetwork$.
This allows $\ournetwork$ to finetune with a non-moving target.
We set $Q=15$ for KITTI, and $Q=5$ for Cityscapes to account for the larger number of images in the Cityscapes training set.
The feature extractor in $\ournetwork$ comprises the first five ResNet18 layers \cite{he2016deep}.
These features are aggregated into a cost volume, the result of which is concatenated with our input image features, and followed by the remaining ResNet18 convolutional layers.
We use the depth decoder from \cite{godard2019digging}.
Our pose network $\posenetwork$ and skip connections for $\ournetwork$ are the same as  \cite{godard2019digging}.
$\teachernetwork$ uses the standard architecture from \cite{godard2019digging} with no modifications.
Full architecture details are in Section~\ref{sec:architecture_details}.
Following \cite{godard2019digging, watson2019depthints, wang2020self, guo2018learning}, we use weights pretrained on ImageNet~\cite{russakovsky2015imagenet}, but provide results trained from scratch in the supplementary material, see Table~\ref{tab:kitti_nopt}.
For all our experiments we set $p=q=0.25$ during training.

\section{Experiments}
\label{sec:evaluation}

Here we evaluate our ManyDepth model and
(1) show that it gives SOTA results by comparing, in a standardized way, to both single-frame and multi-frame depth estimation and
(2) validate our design decisions via ablations.
Additional results are provided in the supplementary material.

\newcommand{\x}{ x }
\newcommand{\midline}{  }

\newcommand{\splitline}{\arrayrulecolor{black}\hhline{~------------}}

\newcommand*\rot{\rotatebox{90}}

\definecolor{Asectioncolor}{RGB}{255, 200, 200}
\definecolor{Bsectioncolor}{RGB}{255, 228, 196}
\definecolor{Csectioncolor}{RGB}{235, 255, 235}
\definecolor{Dsectioncolor}{RGB}{235, 235, 255}

\begin{table*}[t]
  \centering
  \footnotesize
  \resizebox{1.0\textwidth}{!}{
    \begin{tabular}{|l|c|l|c|c|c||c|c|c|c|c|c|c|}
        \arrayrulecolor{black}\hline
          &  TTR & Method & Test frames  & Semantics & WxH & \cellcolor{col1}Abs Rel & \cellcolor{col1}Sq Rel & \cellcolor{col1}RMSE  & \cellcolor{col1}RMSE log & \cellcolor{col2}$\delta < 1.25 $ & \cellcolor{col2}$\delta < 1.25^{2}$ & \cellcolor{col2}$\delta < 1.25^{3}$ \\
         
        \hline\hline
        \parbox[b]{2mm}{\multirow{16}{*}{\rotatebox[origin=c]{90}{Low and medium resolution}}} & 
        \cellcolor{Asectioncolor} & Ranjan \etal \cite{ranjan2018adversarial}  & 1 & & 832\x256 & 0.148 & 1.149 & 5.464 & 0.226 & 0.815 & 0.935 & 0.973\\
        \midline
        & \cellcolor{Asectioncolor} &  EPC++ \cite{luo2019every} & 1  && 832\x256 & 0.141 & 1.029 & 5.350 & 0.216 & 0.816 & 0.941 & 0.976\\
        \midline
        & \cellcolor{Asectioncolor}&  Struct2depth (M) \cite{casser2018depth}  & 1  &$\bullet$ &416\x128& 0.141 & {1.026} & 5.291 &  0.215 & 0.816 & 0.945 & {0.979}\\
        \midline
        & \cellcolor{Asectioncolor}&  Videos in the wild \cite{gordon2019depth} & 1 &$\bullet$ &  416\x128& 
        0.128 & 0.959  & 5.230 & 0.212 & 0.845 & 0.947 & 0.976 \\
        
        
        \midline
        &  \cellcolor{Asectioncolor} &Guizilini \etal \cite{guizilini2020semantically} & 1 &$\bullet$ & 640\x192 & \underline{0.102} & \textbf{0.698} & \textbf{4.381} & \underline{0.178} & \underline{0.896} & \underline{0.964} & \textbf{0.984} \\
        \midline
        &  \cellcolor{Asectioncolor}& Johnston \etal~\cite{johnston2020self} & 1 && 640\x192 & 0.106 & 0.861 & 4.699 & 0.185 & 0.889 & 0.962 & 0.982 \\
        \midline
        &  \cellcolor{Asectioncolor}& Monodepth2 \cite{godard2019digging} & 1  &&  640\x192 &
         {0.115} &   {0.903} &   {4.863} &   {0.193} &   {0.877} &   {0.959} &   {0.981} \\ 
         \midline
        &  \cellcolor{Asectioncolor}& Packnet-SFM \cite{guizilini20203d} & 1 && 640\x192 & 0.111 & 0.785 & 4.601 & 0.189 & 0.878 & 0.960 & 0.982 \\
        \midline
        &  \cellcolor{Asectioncolor}& Li \etal \cite{li2020unsupervised}  & 1    & & 416\x128 &
        0.130 & 0.950 & 5.138 & 0.209 & 0.843 & 0.948 & 0.978 \\
        \midline
        &  \cellcolor{Asectioncolor}& Patil \etal \cite{patil2020dont}  & N    & & 640\x192  & 0.111  & 0.821  & 4.650  & 0.187  & 0.883  & 0.961  & 0.982 \\
        \midline
        &  \cellcolor{Asectioncolor}& Wang \etal \cite{wang2020self} & 2 (-1, 0)    & & 640\x192  & 0.106  & 0.799  & 4.662  & 0.187  & 0.889  & 0.961  & 0.982 \\
        \midline

        
         & \cellcolor{Asectioncolor} &\textbf{ManyDepth (MR)} & 2 (-1, 0) & & 640\x192 &   \textbf{0.098}  &   \underline{0.770}  &   \underline{4.459}  &   \textbf{0.176}  &   \textbf{0.900}  &   \textbf{0.965}  &   \underline{0.983} \\
        
        \splitline

         & $\bullet$  \cellcolor{Bsectioncolor} &GLNet \cite{chen2019self} & 3 (-1,  0, +1)   & & 416\x128  & 0.099  & 0.796  & 4.743  & 0.186  & 0.884  & 0.955  & 0.979 \\
        \midline
        & $\bullet$  \cellcolor{Bsectioncolor} &Luo \etal \cite{luo2020consistent} & N && 384\x112 &  0.130 &  2.086 & 4.876 & 0.205 & 0.878 & 0.946 &  0.970\\
        \midline
        & $\bullet$ \cellcolor{Bsectioncolor} & CoMoDA \cite{kuznietsov2021comoda} &  N & $\bullet$ &  640\x192 & 0.103 & 0.862 & 4.594 &  0.183  & 0.899 &  0.961 &  0.981 \\

        \midline
         & $\bullet$ \cellcolor{Bsectioncolor} &McCraith \etal \cite{mccraith2020monocular} & 2 (0, +1)&& 640\x192 & \textbf{0.089} & \underline{0.747} & \underline{4.275}  & \underline{0.173} & \underline{0.912} & \underline{0.964} & \underline{0.982} \\
        \midline
        & $\bullet$ \cellcolor{Bsectioncolor} & Struct2depth (M+R) \cite{casser2018depth}  &  3 (-1,  0,  +1)       &$\bullet$  & 416\x128  & 0.109  & 0.825  & 4.750  & 0.187  & 0.874  & 0.958  & \textbf{0.983} \\
        & $\bullet$ \cellcolor{Bsectioncolor} & \textbf{{ManyDepth (MR + TTR)}} & 2 (-1, 0) & & 640\x192 &   \underline{0.090}  &   \textbf{0.713}  &   \textbf{4.261}  &   \textbf{0.170}  &   \textbf{0.914}  &   \textbf{0.966}  &   \textbf{0.983} \\
            
        \arrayrulecolor{black}\hline\hline
                
        \parbox[b]{2mm}{\multirow{10}{*}{\rotatebox[origin=c]{90}{High resolution}}} 
        & \cellcolor{Csectioncolor} &Monodepth2 \cite{godard2019digging}   & 1 &  & 1024\x320 &
         {0.115} &   {0.882} &   {4.701} &   {0.190} &   {0.879} &   {0.961} &   {0.982} \\ 
        \midline
         &\cellcolor{Csectioncolor}& Packnet-SFM \cite{guizilini20203d} & 1 &&  1280\x384 &  0.107 & 0.802 & 4.538 & 0.186 & 0.889 & 0.962 & 0.981 \\
         \midline
         & \cellcolor{Csectioncolor}& Guizilini \etal \cite{guizilini2020semantically} & 1 &$\bullet$ & 1280\x384 & 0.100 & 0.761 & 4.270 & 0.175 & 0.902 & 0.965 & 0.982 \\
         \midline
         &  \cellcolor{Csectioncolor}& Shu \etal~\cite{shu2020feature} (ResNet50) & 1 && 1024\x320 & 0.104 & 0.729 & 4.481 & 0.179 & 0.893 & 0.965 & \textbf{0.984} \\
         \midline
        & \cellcolor{Csectioncolor} &Wang \etal \cite{wang2020self} & 2 (-1, 0)    & & 1024\x320  & 0.106  & 0.773  & 4.491  & 0.185  & 0.890  & 0.962  & 0.982 \\
        \midline
        

        
        & \cellcolor{Csectioncolor}& \textbf{ManyDepth (HR)} & 2 (-1, 0) && 1024\x320 &   \underline{0.093}  &   \underline{0.715}  &   \textbf{4.245}  &   \underline{0.172}  &   \underline{0.909}  &   \underline{0.966}  &   \underline{0.983}  \\
        
        \midline
        
        & \cellcolor{Csectioncolor} & \textbf{ManyDepth (HR ResNet50)} & 2 (-1, 0) && 1024\x320 &    \textbf{0.091}  &   \textbf{0.694}  &   \textbf{4.245}  &   \textbf{0.171}  &   \textbf{0.911}  &   \textbf{0.968}  &   \underline{0.983}  \\
        
        \splitline 

        & $\bullet$ \cellcolor{Dsectioncolor} & McCraith \etal \cite{mccraith2020monocular} & 2 (0, +1)  & & 1024\x320 & 0.089 & 0.756 & 4.228 & 0.170 & 0.917 & 0.967 & \textbf{0.983} \\
        
        \midline
        & $\bullet$ \cellcolor{Dsectioncolor} &Shu \etal~\cite{shu2020feature}  (ResNet50) & 3 (-1, 0, +1)  &  & 1024\x320 & 0.088 &  0.712 & \textbf{4.137} & 0.169 & 0.915 & 0.965 & 0.982 \\
        
        \midline
        
        & $\bullet$ \cellcolor{Dsectioncolor} &\textbf{ManyDepth (HR + TTR)} & 2 (-1, 0) && 1024\x320 &   \textbf{0.087}  &   \underline{0.696}  &   4.183  &   \textbf{0.167}  &   \underline{0.918}  &   \textbf{0.968}  &   \textbf{0.983}  \\
        
        & $\bullet$ \cellcolor{Dsectioncolor} &\textbf{ManyDepth (HR R50 + TTR)} & 2 (-1, 0) && 1024\x320 &   \textbf{0.087}  &   \textbf{0.685}  &   \underline{4.142}  &   \textbf{0.167}  &   \textbf{0.920}  &   \textbf{0.968}  &   \textbf{0.983}  \\

        \arrayrulecolor{black}\hline

    \end{tabular}
  }  
  \vspace{-1pt}
  \caption{\textbf{Comparison of our method to existing self-supervised approaches on the KITTI~\cite{Geiger2012CVPR} Eigen split.}   
At the top we compare medium and low resolution results {\setlength{\fboxsep}{0pt}\colorbox{Asectioncolor}{without}}  and {\setlength{\fboxsep}{0pt}\colorbox{Bsectioncolor}{with}} test-time refinement (TTR).
  At bottom we compare high resolution results {\setlength{\fboxsep}{0pt}\colorbox{Csectioncolor}{without}}  and {\setlength{\fboxsep}{0pt}\colorbox{Dsectioncolor}{with}} TTR. 
   The best results in each subsection are in \textbf{bold}; second best are \underline{underlined}.
   Our method outperforms all previous methods in all subsections across most metrics, whether or not the baselines use multiple frames at test time. 
   We indicate if a method uses semantic supervision (Semantics) and methods indicated by N take a long sequence of frames as input for each test image (\eg the preceding frames or frames before and after in time).
    \label{tab:kitti_eigen}} 
    
    \vspace{-5pt}

\end{table*}

We evaluate on two challenging depth estimation datasets, both of which exhibit moving objects.
For both, we use the standard depth evaluation metrics from  \cite{eigen2015predicting, eigen2014depth}.\\
\textbf{(a)~KITTI}~\cite{Geiger2012CVPR}.
We use the Eigen split from \cite{eigen2015predicting}.
This is commonly used for single frame depth estimation, but is more recently also used for multi-frame approaches \eg ~\cite{wang2019recurrent,patil2020dont}.
22 frames in the KITTI Eigen test set are at the start of a sequence, and do not have a previous frame.
We still include these images in the evaluation.
For these images, the network does not have access to any other frames and thus makes a prediction based on one frame only.
In Table~\ref{tab:kitti_improved_ground_truth}, we additionally include models evaluated on the improved KITTI ground truth~\cite{uhrig2017sparse}.\\
\textbf{(b)~Cityscapes}~\cite{Cordts2016Cityscapes}. Following \cite{zhou2017unsupervised, yang2018lego, geonet2018}, we train on 69,731 images from the monocular sequences, which we preprocess into triples using the scripts from \cite{zhou2017unsupervised}.
We do not use stereo pairs or semantics.
We evaluate on the 1,525 test images using the provided SGM~\cite{hirschmuller2007stereo} disparity maps.
As with KITTI, we clip predicted depths at 80m, and only evaluate on ground truth depths less than 80m.

\subsection{KITTI results}
In Table~\ref{tab:kitti_eigen} we compare to multi-frame approaches, some of which, \eg \cite{chen2019self,casser2018depth,patil2020dont,wang2020self,mccraith2020monocular}, see more frames than ours or also use future frames \eg  \cite{chen2019self,casser2018depth,mccraith2020monocular}.
We do not include results from \cite{liu2019neural} as they do not provide their scores on the KITTI Eigen split (see \cite{xie2019video}).
We additionally compare to the best-performing self-supervised monocular depth estimation approaches.
To control for resolution, we separate low and high resolution models, and we also split methods which use expensive multi-pass test-time refinement into separate sections.
We observe that our approach outperforms all previously published self-supervised methods that do not use semantic supervision on most metrics.
We also implement the test-time refinement scheme of \cite{mccraith2020monocular} on our model, updating the weights of the depth and pose encoders using sequential pairs of images from the test set, for 50 steps.
Not surprisingly, this further improves our results, and we outperform other test-time refinement methods.

Qualitative results are presented in Fig.~\ref{fig:qualitative_results_kitti}.
In some cases the predicted depth maps looks qualitatively similar to the monocular only models, but the error maps show the high magnitude of mistakes which can be present.

\definecolor{cyan}{rgb}{0.0, 1.0, 1.0}

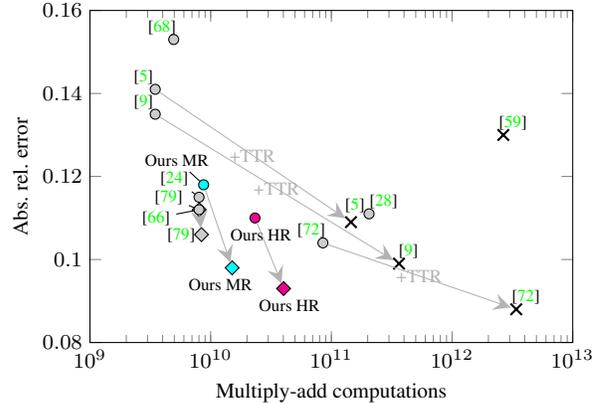
\begin{figure}
  \centering
 \footnotesize
\vspace{5pt}
\pgfplotsset{
    yticklabel style={/pgf/number format/fixed},  
}
\begin{tikzpicture}[trim axis right]
    \tikzset{myptr/.style={black!30,decoration={markings,mark=at position 1 with %
        {\arrow[scale=2,>=stealth]{>}}},postaction={decorate}}}
    \begin{axis}[
        xmode=log, 
      height=6cm, width=8cm, 
    xmin=1000000000, xmax=10000000000000,
    ymin=0.08, ymax=0.16,
      samples=200,
      xlabel = Multiply-add computations, ylabel = Abs.~rel.~error
    ]

\node[circle, fill=black!20, scale=0.5,  draw]
    (casser2018depth_single) at (axis cs: 3473193984.0, 0.141) {};
\node[scale=0.8]
    (casser2018depth_single_label) at (axis cs: 2778555187.2000003, 0.144) {\cite{casser2018depth}};
\node[cross out, thick, scale=0.7,  draw]
    (casser2018depth_multi) at (axis cs: 145418343360.0, 0.109) {};
\node[scale=0.8]
    (casser2018depth_multi_label) at (axis cs: 157051810828.80002, 0.113) {\cite{casser2018depth}};
\node[cross out, thick, scale=0.7,  draw]
    (luo2020consistent_multi) at (axis cs: 2659085184000.0, 0.13) {};
\node[scale=0.8]
    (luo2020consistent_multi_label) at (axis cs: 3190902220800.0, 0.133) {\cite{luo2020consistent}};
\node[circle, fill=black!20, scale=0.5,  draw]
    (shu2020feature_single) at (axis cs: 85107616000.0, 0.104) {};
\node[scale=0.8]
    (shu2020feature_single_label) at (axis cs: 68086092800.0, 0.107) {\cite{shu2020feature}};
\node[cross out, thick, scale=0.7,  draw]
    (shu2020feature_multi) at (axis cs: 3404304640000.0, 0.088) {};
\node[scale=0.8]
    (shu2020feature_multi_label) at (axis cs: 4085165568000.0, 0.091) {\cite{shu2020feature}};
\node[circle, fill=black!20, scale=0.5,  draw]
    (godard2019digging_single) at (axis cs: 8015063039.999999, 0.115) {};
\node[scale=0.8]
    (godard2019digging_single_label) at (axis cs: 5001399336.96, 0.11950000000000001) {\cite{godard2019digging}};
\draw[black]
    (godard2019digging_single) -- (godard2019digging_single_label);
\node[diamond, fill=black!20, scale=0.5,  draw]
    (patil2020dont_multi) at (axis cs: 8045189119.999996, 0.112) {};
\node[scale=0.8]
    (patil2020dont_multi_label) at (axis cs: 3536151295.999996, 0.110) {\cite{patil2020dont}};
\draw[black] 
    (patil2020dont_multi) -- (patil2020dont_multi_label);
\node[circle, fill=black!20, scale=0.5,  draw]
    (guizilini20203d_single) at (axis cs: 205318296960.0, 0.111) {};
\node[scale=0.8]
    (guizilini20203d_single_label) at (axis cs: 271020151987.2, 0.114) {\cite{guizilini20203d}};
\node[circle, fill=black!20, scale=0.5,  draw]
    (poggi2018towards_single) at (axis cs: 4940000000.0, 0.153) {};
\node[scale=0.8]
    (poggi2018towards_single_label) at (axis cs: 3952000000.0, 0.156) {\cite{poggi2018towards}};
\node[circle, fill=black!20, scale=0.5,  draw]
    (chen2019self_single) at (axis cs: 3473193984.0, 0.135) {};
\node[scale=0.8]
    (chen2019self_single_label) at (axis cs: 2778555187.2000003, 0.138) {\cite{chen2019self}};
\node[cross out, thick, scale=0.7,  draw]
    (chen2019self_multi) at (axis cs: 363545858400.0, 0.099) {};
\node[scale=0.8]
    (chen2019self_multi_label) at (axis cs: 436255030080.0, 0.10200000000000001) {\cite{chen2019self}};
\node[circle, fill=cyan, scale=0.5,  draw]
    (ours_single) at (axis cs: 8738866560.0, 0.118) {};
\node[scale=0.8]
    (ours_single_label) at (axis cs: 5243319936.0, 0.124) {Ours MR};
\draw[black]
    (ours_single) -- (ours_single_label);
\node[diamond, fill=cyan, scale=0.5,  draw]
    (ours_multi) at (axis cs: 15068508960.0, 0.098) {};
\node[scale=0.8]
    (ours_multi_label) at (axis cs: 12054807168.0, 0.094) {Ours MR};
\node[circle, fill=magenta, scale=0.5,  draw]
    (ours_hr_single) at (axis cs: 23303644160.0, 0.11) {};
\node[scale=0.8]
    (ours_hr_single_label) at (axis cs: 26100081459.199997, 0.106) {Ours HR};
\node[diamond, fill=magenta, scale=0.5,  draw]
    (ours_hr_multi) at (axis cs: 40182690560.0, 0.093) {};
\node[scale=0.8]
    (ours_hr_multi_label) at (axis cs: 45004613427.2, 0.089) {Ours HR};
\node[circle, fill=black!20, scale=0.5,  draw]
    (wang2020self_single) at (axis cs: 8015063039.999999, 0.112) {};
\node[scale=0.8]
    (wang2020self_single_label) at (axis cs: 4488435302.4, 0.115) {\cite{wang2020self}};
\draw[black]
    (wang2020self_single) -- (wang2020self_single_label);
\node[diamond, fill=black!20, scale=0.5,  draw]
    (wang2020self_multi) at (axis cs: 8388613180.0, 0.106) {};
\node[scale=0.8]
    (wang2020self_multi_label) at (axis cs: 5704256962.4, 0.106) {\cite{wang2020self}};
\draw[myptr]
    (casser2018depth_single) -- node[midway,font=\scriptsize] {+TTR} (casser2018depth_multi);
\draw[myptr]
    (shu2020feature_single) -- node[midway,font=\scriptsize] {+TTR} (shu2020feature_multi);
\draw[myptr]
    (chen2019self_single) -- node[midway,font=\scriptsize] {+TTR} (chen2019self_multi);
\draw[myptr]
    (ours_single) -- node[midway,font=\scriptsize] {} (ours_multi);
\draw[myptr]
    (ours_hr_single) -- node[midway,font=\scriptsize] {} (ours_hr_multi);
\draw[myptr]
    (wang2020self_single) -- node[midway,font=\scriptsize] {} (wang2020self_multi);

    \end{axis}
\end{tikzpicture}
  
  \vspace{2pt}
  \caption[\textbf{Our single-pass network is significantly more efficient than test-time optimisation methods.}] {
  \textbf{Our single-pass network is significantly more efficient than test-time optimisation.} We compare abs.~rel.~error ($y$-axis) against MACs ($x$-axis) on the KITTI Eigen test set. \newline
   \begin{tikz}\node[circle,fill=black!20,scale=0.5,draw]{};\end{tikz} -- Single frame models. These tend to have low MACs. 
   \newline
  \begin{tikz}\node[cross out, thick, scale=0.7,  draw]{};\end{tikz} -- Multi-frame models which use test-time refinement. \newline
   \begin{tikz}\node[diamond,fill=black!20,scale=0.5,draw]{};\end{tikz} -- Multi-frame models with a single forward pass at test time. \newline
   Methods which are more accurate than ours take over two orders of magnitude more time to compute (note the logarithmic scale on the $x$-axis.). Our MR multi-frame version (\begin{tikz} \node[diamond,fill=cyan,scale=0.5,draw]{};\end{tikz}) has better accuracy than \cite{wang2020self}, who has a similar runtime.
   
  \label{fig:kitti_eigen_runtime} }
  \vspace{5pt}
\end{figure}

\vspace{4pt}
\paragraph{Efficiency comparison.}
Fig.~\ref{fig:kitti_eigen_runtime} illustrates the runtime efficiency of our ManyDepth models
(640 x 192: \begin{tikz}\node[circle,fill=cyan,scale=0.5,draw]{};\end{tikz}, \begin{tikz}\node[diamond,fill=cyan,scale=0.5,draw]{};\end{tikz}
and 1024 x 320: \begin{tikz}\node[circle,fill=magenta,scale=0.5,draw]{};\end{tikz}, \begin{tikz}\node[diamond,fill=magenta,scale=0.5,draw]{};\end{tikz})
compared to other methods, including test-time refinement approaches (\begin{tikz}\node[cross out, thick, scale=0.7,  draw]{};\end{tikz}).
We report multiply-add computations (MACs) for each method and show that test-time refinement models which perform multiple forward-backward passes are too computationally demanding for use in real-time applications.
See Table~\ref{tab:macs} for a full results table, and Section~\ref{sec:macs} for additional details.

\renewcommand{\splitline}{}

\newcommand{\CSmid}{\arrayrulecolor{black}\hhline{----}}

\begin{table*}
  \centering

  \resizebox{0.9\textwidth}{!}{
  \begin{tabular}{|l|c|c|c||c|c|c|c|c|c|c|}
\arrayrulecolor{black}\hline
Method & Test frames &  Semantics & WxH & \cellcolor{col1}Abs Rel & \cellcolor{col1}Sq Rel & \cellcolor{col1}RMSE  & \cellcolor{col1}RMSE log & \cellcolor{col2}$\delta < 1.25 $ & \cellcolor{col2}$\delta < 1.25^{2}$ & \cellcolor{col2}$\delta < 1.25^{3}$ \\
\hline\hline

Struct2Depth 2 \cite{Casser_2019_CVPR_Workshops}  &  1  &  $\bullet$ &   416\x128 &
    0.145  & 1.737  & 7.280  &  0.205 & 0.813 & 0.942 & 0.976 \\

Pilzer \etal \cite{pilzer2018unsupervised} & 1 & &  512\x256 &
    0.240 & 4.264 & 8.049 & 0.334 & 0.710 &  0.871 & 0.937 \\
    
Monodepth2$\dagger$ \cite{godard2019digging} & 1  & & 416\x128 &
    0.129  &   1.569  &   6.876  &   0.187  &   0.849  &   0.957  &   0.983 \\

Videos in the Wild \cite{gordon2019depth} & 1 &$\bullet$ &  416\x128 &
    {0.127} & {1.330} & {6.960} & {0.195} & {0.830} & {0.947} & {0.981} \\

Li \etal \cite{li2020unsupervised} & 1 & & 416\x128 &
    0.119 &  1.290 & {6.980} &  {0.190} & {0.846} &  0.952 &  0.982 \\

\hline

Struct2Depth 2 \cite{Casser_2019_CVPR_Workshops}  &  3 (-1,  0,  +1)    &  & 416\x128 &
    0.222  & 5.737  & 8.613  & 0.258  & 0.774  & 0.908 & 0.954 \\

Struct2Depth 2 \cite{Casser_2019_CVPR_Workshops}  &  3 (-1,  0,  +1)  & $\bullet$ & 416\x128 &
    0.151 & 2.492 & 7.024 & 0.202 & 0.826 & 0.937 & 0.972 \\

{\bf ManyDepth} & 2 (-1, 0)  & &  416\x128 &
    \textbf{0.114}  &   \textbf{1.193}  &   \textbf{6.223}  &   \textbf{0.170}  &   \textbf{0.875}  &   \textbf{0.967}  &   \textbf{0.989} \\

    \arrayrulecolor{black}\hline
  \end{tabular}
  }
  \vspace{1pt}
  \caption{\textbf{Results on Cityscapes.}
  Our method gives superior performance to all competing models.
  $\dagger$ is trained by us with the authors' code, with preprocessing from \cite{zhou2017unsupervised}. See Section~\ref{sec:cityscapes_monodepth} for more details.
  Results from \cite{pilzer2018unsupervised} are their `Half-Cycle Mono' model, their only variant not requiring test-time stereo pairs.
  We evaluate using the cropping scheme of \cite{Casser_2019_CVPR_Workshops} following conversations with the authors; see Section~\ref{sec:cityscapes_cropping} for specifics.
  \label{tab:cs_train_test}}
  \vspace{-4pt}
\end{table*}


\subsection{KITTI ablation}
\label{sec_kitti_ab}
In Table~\ref{tab:kitti_eigen_ablations} we show the importance of the various components of our approach by turning them on and off in turn.
\begin{description}
\vspace{-8pt}
    \setlength{\itemsep}{-4pt}%

\item[ManyDepth w/o motion masking:]  We omit $L_\text{consistency}$ from our loss and set $M$ to zeros everywhere.

\item[ManyDepth w/o motion masking, w/o augmentation:] As above, but also omitting our augmentations.

\item[ManyDepth with motion masking but no teacher:] We remove $L_\text{consistency}$ but still use $M$ to mask $L_p$.

\item[Stack of 2 frames as input:]
A baseline which directly maps ($I_{t-1}, I_t$) to $D_t$. We modify \cite{godard2019digging}'s network to accept two images as input, and train using their loss.

\item[ManyDepth with motion masking from \cite{ranjan2018adversarial}:]
We use our full loss, but our mask $M$ is the same as \cite{ranjan2018adversarial}. We use their pretrained models to compute these masks offline for the entire training set.

\item[ManyDepth with motion masking from \cite{godard2019digging}:]
Here we use our full loss, but set the mask $M$ to an `automask' from  \cite{godard2019digging}.

\item[Khot \etal \cite{khot2019learning}:]
We trained this unsupervised MVS approach on KITTI, with the implementation from \cite{pytorch-mvsnet}.

\item[ManyDepth ($I_{t-2}, I_{t-1}, I_{t}$) \textnormal{and} ManyDepth ($I_{t-1}, I_t, I_{t+1}$):]
Retrained variants of our model which build the cost volume from three frames instead of just two. This improves some metrics but not all.
\end{description}

\vspace{4pt}
\paragraph{Benefit of our augmentations.}
In Table~\ref{tab:augmentation_ablations} we evaluate three different scenarios, comparing our model to a baseline which was trained without our augmentations from Section~\ref{sec:augmentations}. 
When evaluating in `standard' mode (\ie using the previous and current frames as input) on the entire KITTI test set, the difference between the two models is negligible.
This is partially because the KITTI test images are predominately from a moving camera.
However, when we evaluate in `start-of-sequence' mode (\ie the standard monocular setting using only $(I_t)$ as input) and `static camera' evaluation mode (\ie simulating a static camera with inputs $(I_t, I_t)$), our augmentation scheme is significantly better.

\subsection{Cityscapes results}
In Table \ref{tab:cs_train_test} we perform additional comparisons where we train and test on the Cityscapes dataset~\cite{Cordts2016Cityscapes}.
Again, we consistently outperform competing methods, even those that use semantic supervision.

\begin{table}
  \centering
  \vspace{-2pt}
  \resizebox{1.0\columnwidth}{!}{
  \begin{tabular}{|l||c|c|c|}
  \hline
     Ablation & \cellcolor{col1}Abs Rel & \cellcolor{col1}Sq Rel & \cellcolor{col1}RMSE \\
  \hline\hline
  ManyDepth full &   \textbf{0.098}  &   \textbf{0.770}  &   4.459  \\
  ManyDepth (w/o motion masking) &   0.113  &   1.354  &   5.228  \\
  ManyDepth (w/o motion masking, w/o aug.) & 0.284  &  11.240  &   8.516 \\
  ManyDepth (with motion masking, w/o teacher) &   0.154  &   2.682  &   6.573  \\
    Stack of 2 frames as input ($I_{t-1}, I_{t}$)  &   0.121  &   1.028  &   5.016 \\

    ManyDepth (with motion masking from \cite{ranjan2018adversarial}) & 0.099  &   0.783  &   \textbf{4.447}  \\

    ManyDepth (with motion masking from \cite{godard2019digging})  & 0.099  &   0.780  &   4.465  \\
    Khot \etal \cite{khot2019learning} reimplementation &   0.200   &        4.694   &    7.232 \\
    \hline
  ManyDepth with 3-frame input ($I_{t-2}, I_{t-1}, I_{t}$) &   0.098  &   0.780  &   \textbf{4.430}  \\
  ManyDepth with 3-frame input ($I_{t-1}, I_{t}, I_{t+1}$) &   \textbf{0.097}  &   \textbf{0.768}  &   4.431  \\

\hline
  \end{tabular}
  }
  \vspace{1pt}
  \caption{\textbf{Our contributions lead to better scores.} Here we ablate our ManyDepth method on KITTI 2015 \cite{Geiger2012CVPR} using the Eigen split. Full numbers are in Table~\ref{tab:kitti_eigen_ablations_sup_mat}.
  \label{tab:kitti_eigen_ablations}}
\end{table}

\begin{table}
  \centering
  \footnotesize
  \resizebox{1.0\columnwidth}{!}{

\begin{tabular}{|l|l||c|c|c|}
  \hline

     Test-time input & Model & \cellcolor{col1}Abs Rel & \cellcolor{col1}Sq Rel & \cellcolor{col1}RMSE  \\ 
     \hline\hline
  \multirow{2}{*}{Standard: $(I_{t-1}, I_t)$} & No augmentation & 0.100  &   0.794  &   \textbf{4.432}  \\
   & ManyDepth &   \textbf{0.098}  &   \textbf{0.770}  &   {4.459}  \\
  
  \hline
  \multirow{3}{*}{Start-of-sequence: $(I_t)$} 
  & Monodepth2 \cite{godard2019digging} & \textbf{0.115} & 0.903 & 4.863 \\
  & No augmentation & 0.148  &   1.076  &   5.161  \\
   & ManyDepth & {0.118}  &   \textbf{0.892}  &   \textbf{4.764}  \\
  
  \hline
  \multirow{2}{*}{Static camera: $(I_t, I_t)$} & No augmentation & 0.158  &   1.132  &   5.228  \\
   & ManyDepth & \textbf{0.117}  &   \textbf{0.886}  &   \textbf{4.754} \\
\hline
  \end{tabular}
  }
  \vspace{1pt}
  \caption{\textbf{Our augmentations help in static camera and start-of-sequence cases.}
  We compare two variants of our model, one trained with our novel data augmentations (`Ours') and one without.
  We create two artificial scenarios to test each model's performance on start-of-sequence images (where we just input $I_t$) and static cameras (where both input frames are the exact same).
  \label{tab:augmentation_ablations}}
  \vspace{-5pt}
\end{table}

\section{Conclusion}
We presented a fully self-supervised online method that predicts superior depths from a single image, or from multiple images when they are available.
We achieve the benefits of both multi-frame and monocular methods, while being more robust on moving objects and static cameras compared to a naive integration of a cost volume.  
We presented state-of-the-art results on both the KITTI and Cityscapes datasets.
While test-time refinement methods are close competitors in terms of depth accuracy, we have shown that our method is significantly more efficient during inference.
We expect that our results could be further improved via recent complimentary advances in monocular depth estimation \eg discretized output depths \cite{gonzalez2020forget} or feature based losses~\cite{shu2020feature}.

\vspace{4pt}
{
\paragraph{Acknowledgements.}
Thanks to Jamie Shotton; a conversation with him at a BMVA workshop motivated work in this area. Also to Daniyar Turmukhambetov and Sara Vicente for their valuable feedback, and to the authors of \cite{Casser_2019_CVPR_Workshops} for CityScapes evaluation help.
}

{\small
\bibliographystyle{ieee_fullname}
\bibliography{main}

\begin{thebibliography}{10}\itemsep=-1pt

\bibitem{aleotti2020reversing}
Filippo Aleotti, Fabio Tosi, Li Zhang, Matteo Poggi, and Stefano Mattoccia.
\newblock Reversing the cycle: self-supervised deep stereo through enhanced
  monocular distillation.
\newblock In {\em ECCV}, 2020.

\bibitem{babu2018undemon}
V~Madhu Babu, Kaushik Das, Anima Majumdar, and Swagat Kumar.
\newblock Undemon: Unsupervised deep network for depth and ego-motion
  estimation.
\newblock In {\em IROS}, 2018.

\bibitem{bian2019unsupervised}
Jiawang Bian, Zhichao Li, Naiyan Wang, Huangying Zhan, Chunhua Shen, Ming-Ming
  Cheng, and Ian Reid.
\newblock Unsupervised scale-consistent depth and ego-motion learning from
  monocular video.
\newblock In {\em NeurIPS}, 2019.

\bibitem{bloesch2018codeslam}
Michael Bloesch, Jan Czarnowski, Ronald Clark, Stefan Leutenegger, and Andrew~J
  Davison.
\newblock {CodeSLAM}—learning a compact, optimisable representation for dense
  visual {SLAM}.
\newblock In {\em CVPR}, 2018.

\bibitem{casser2018depth}
Vincent Casser, Soeren Pirk, Reza Mahjourian, and Anelia Angelova.
\newblock Depth prediction without the sensors: Leveraging structure for
  unsupervised learning from monocular videos.
\newblock In {\em AAAI}, 2019.

\bibitem{Casser_2019_CVPR_Workshops}
Vincent Casser, Soeren Pirk, Reza Mahjourian, and Anelia Angelova.
\newblock Unsupervised monocular depth and ego-motion learning with structure
  and semantics.
\newblock In {\em CVPR Workshops}, 2019.

\bibitem{chang2018pyramid}
Jia-Ren Chang and Yong-Sheng Chen.
\newblock Pyramid stereo matching network.
\newblock In {\em CVPR}, 2018.

\bibitem{chen2016single}
Weifeng Chen, Zhao Fu, Dawei Yang, and Jia Deng.
\newblock Single-image depth perception in the wild.
\newblock In {\em NeurIPS}, 2016.

\bibitem{chen2019self}
Yuhua Chen, Cordelia Schmid, and Cristian Sminchisescu.
\newblock Self-supervised learning with geometric constraints in monocular
  video: Connecting flow, depth, and camera.
\newblock In {\em ICCV}, 2019.

\bibitem{cheng2019learning}
Xinjing Cheng, Peng Wang, and Ruigang Yang.
\newblock Learning depth with convolutional spatial propagation network.
\newblock {\em PAMI}, 2019.

\bibitem{collins1996space}
Robert~T Collins.
\newblock A space-sweep approach to true multi-image matching.
\newblock In {\em CVPR}, 1996.

\bibitem{Cordts2016Cityscapes}
Marius Cordts, Mohamed Omran, Sebastian Ramos, Timo Rehfeld, Markus Enzweiler,
  Rodrigo Benenson, Uwe Franke, Stefan Roth, and Bernt Schiele.
\newblock The cityscapes dataset for semantic urban scene understanding.
\newblock In {\em CVPR}, 2016.

\bibitem{dai2019mvs2}
Yuchao Dai, Zhidong Zhu, Zhibo Rao, and Bo Li.
\newblock {MVS}$^2$: Deep unsupervised multi-view stereo with multi-view
  symmetry.
\newblock In {\em 3DV}, 2019.

\bibitem{eigen2015predicting}
David Eigen and Rob Fergus.
\newblock Predicting depth, surface normals and semantic labels with a common
  multi-scale convolutional architecture.
\newblock In {\em ICCV}, 2015.

\bibitem{eigen2014depth}
David Eigen, Christian Puhrsch, and Rob Fergus.
\newblock Depth map prediction from a single image using a multi-scale deep
  network.
\newblock In {\em NeurIPS}, 2014.

\bibitem{engel2014lsd}
Jakob Engel, Thomas Sch{\"o}ps, and Daniel Cremers.
\newblock {LSD-SLAM}: Large-scale direct monocular slam.
\newblock In {\em ECCV}, 2014.

\bibitem{facil2017single}
Jos{\'e}~M F{\'a}cil, Alejo Concha, Luis Montesano, and Javier Civera.
\newblock Single-view and multi-view depth fusion.
\newblock {\em IEEE Robotics and Automation Letters}, 2017.

\bibitem{facil2019cam}
Jose~M Facil, Benjamin Ummenhofer, Huizhong Zhou, Luis Montesano, Thomas Brox,
  and Javier Civera.
\newblock {CAM-Convs}: Camera-aware multi-scale convolutions for single-view
  depth.
\newblock In {\em CVPR}, 2019.

\bibitem{fischer2015flownet}
Philipp Fischer, Alexey Dosovitskiy, Eddy Ilg, Philip H{\"a}usser, Caner
  Haz{\i}rba{\c{s}}, Vladimir Golkov, Patrick van~der Smagt, Daniel Cremers,
  and Thomas Brox.
\newblock {FlowNet}: Learning optical flow with convolutional networks.
\newblock In {\em ICCV}, 2015.

\bibitem{fu2018deep}
Huan Fu, Mingming Gong, Chaohui Wang, Kayhan Batmanghelich, and Dacheng Tao.
\newblock Deep ordinal regression network for monocular depth estimation.
\newblock In {\em CVPR}, 2018.

\bibitem{garg2016unsupervised}
Ravi Garg, Vijay Kumar~BG, and Ian Reid.
\newblock Unsupervised {CNN} for single view depth estimation: Geometry to the
  rescue.
\newblock In {\em ECCV}, 2016.

\bibitem{Geiger2012CVPR}
Andreas Geiger, Philip Lenz, and Raquel Urtasun.
\newblock {Are we ready for Autonomous Driving? The KITTI Vision Benchmark
  Suite}.
\newblock In {\em CVPR}, 2012.

\bibitem{godard2017unsupervised}
Cl{\'e}ment Godard, Oisin Mac~Aodha, and Gabriel~J Brostow.
\newblock Unsupervised monocular depth estimation with left-right consistency.
\newblock In {\em CVPR}, 2017.

\bibitem{godard2019digging}
Cl{\'e}ment Godard, Oisin Mac~Aodha, Michael Firman, and Gabriel~J. Brostow.
\newblock Digging into self-supervised monocular depth estimation.
\newblock In {\em ICCV}, 2019.

\bibitem{gonzalez2020forget}
Juan~Luis Gonzalez and Munchurl Kim.
\newblock Forget about the {LiDAR}: Self-supervised depth estimators with {MED}
  probability volumes.
\newblock {\em NeurIPS}, 2020.

\bibitem{gordon2019depth}
Ariel Gordon, Hanhan Li, Rico Jonschkowski, and Anelia Angelova.
\newblock Depth from videos in the wild: Unsupervised monocular depth learning
  from unknown cameras.
\newblock In {\em ICCV}, 2019.

\bibitem{gu2020cascade}
Xiaodong Gu, Zhiwen Fan, Siyu Zhu, Zuozhuo Dai, Feitong Tan, and Ping Tan.
\newblock Cascade cost volume for high-resolution multi-view stereo and stereo
  matching.
\newblock In {\em CVPR}, 2020.

\bibitem{guizilini20203d}
Vitor Guizilini, Rares Ambrus, Sudeep Pillai, Allan Raventos, and Adrien
  Gaidon.
\newblock {3D} packing for self-supervised monocular depth estimation.
\newblock In {\em CVPR}, 2020.

\bibitem{guizilini2020semantically}
Vitor Guizilini, Rui Hou, Jie Li, Rares Ambrus, and Adrien Gaidon.
\newblock Semantically-guided representation learning for self-supervised
  monocular depth.
\newblock In {\em ICLR}, 2020.

\bibitem{pytorch-mvsnet}
Xiaoyang Guo.
\newblock {PyTorch} implementation of {MVSNet}.
\newblock \url{https://github.com/xy-guo/MVSNet_pytorch}, 2020.

\bibitem{guo2018learning}
Xiaoyang Guo, Hongsheng Li, Shuai Yi, Jimmy Ren, and Xiaogang Wang.
\newblock Learning monocular depth by distilling cross-domain stereo networks.
\newblock In {\em ECCV}, 2018.

\bibitem{ha2016high}
Hyowon Ha, Sunghoon Im, Jaesik Park, Hae-Gon Jeon, and In So~Kweon.
\newblock High-quality depth from uncalibrated small motion clip.
\newblock In {\em CVPR}, 2016.

\bibitem{he2016deep}
Kaiming He, Xiangyu Zhang, Shaoqing Ren, and Jian Sun.
\newblock Deep residual learning for image recognition.
\newblock In {\em CVPR}, 2016.

\bibitem{hirschmuller2007stereo}
Heiko Hirschmuller.
\newblock Stereo processing by semiglobal matching and mutual information.
\newblock {\em PAMI}, 2007.

\bibitem{hou2019multi}
Yuxin Hou, Juho Kannala, and Arno Solin.
\newblock Multi-view stereo by temporal nonparametric fusion.
\newblock In {\em ICCV}, 2019.

\bibitem{huang2020m3vsnet}
Baichuan Huang, Can Huang, Yijia He, Jingbin Liu, and Xiao Liu.
\newblock M$^{3}${VSNet}: Unsupervised multi-metric multi-view stereo network.
\newblock {\em arXiv:2005.00363}, 2020.

\bibitem{huang2018DeepMVS}
Po-Han Huang, Kevin Matzen, Johannes Kopf, Narendra Ahuja, and Jia-Bin Huang.
\newblock {DeepMVS}: Learning multi-view stereopsis.
\newblock In {\em CVPR}, 2018.

\bibitem{im2019dpsnet}
Sunghoon Im, Hae-Gon Jeon, Stephen Lin, and In~So Kweon.
\newblock {DPSNet}: End-to-end deep plane sweep stereo.
\newblock {\em ICLR}, 2019.

\bibitem{ji2017surfacenet}
Mengqi Ji, Juergen Gall, Haitian Zheng, Yebin Liu, and Lu Fang.
\newblock {SurfaceNet}: An end-to-end {3D} neural network for multiview
  stereopsis.
\newblock In {\em ICCV}, 2017.

\bibitem{johnston2020self}
Adrian Johnston and Gustavo Carneiro.
\newblock Self-supervised monocular trained depth estimation using
  self-attention and discrete disparity volume.
\newblock In {\em CVPR}, 2020.

\bibitem{joshi2014micro}
Neel Joshi and C~Lawrence Zitnick.
\newblock Micro-baseline stereo.
\newblock {\em Microsoft Research Technical Report}, 2014.

\bibitem{kang2001handling}
Sing~Bing Kang, Richard Szeliski, and Jinxiang Chai.
\newblock Handling occlusions in dense multi-view stereo.
\newblock In {\em CVPR}, 2001.

\bibitem{kar2017learning}
Abhishek Kar, Christian H{\"a}ne, and Jitendra Malik.
\newblock Learning a multi-view stereo machine.
\newblock In {\em NeurIPS}, 2017.

\bibitem{karsch2014depth}
Kevin Karsch, Ce Liu, and Sing~Bing Kang.
\newblock Depth transfer: Depth extraction from video using non-parametric
  sampling.
\newblock {\em PAMI}, 2014.

\bibitem{kendall2017end}
Alex Kendall, Hayk Martirosyan, Saumitro Dasgupta, Peter Henry, Ryan Kennedy,
  Abraham Bachrach, and Adam Bry.
\newblock End-to-end learning of geometry and context for deep stereo
  regression.
\newblock In {\em ICCV}, 2017.

\bibitem{khot2019learning}
Tejas Khot, Shubham Agrawal, Shubham Tulsiani, Christoph Mertz, Simon Lucey,
  and Martial Hebert.
\newblock Learning unsupervised multi-view stereopsis via robust photometric
  consistency.
\newblock In {\em CVPR Workshops}, 2019.

\bibitem{kingma2014adam}
Diederik~P Kingma and Jimmy Ba.
\newblock Adam: A method for stochastic optimization.
\newblock {\em arXiv:1412.6980}, 2014.

\bibitem{klingner2020self}
Marvin Klingner, Jan-Aike Term{\"o}hlen, Jonas Mikolajczyk, and Tim
  Fingscheidt.
\newblock Self-supervised monocular depth estimation: Solving the dynamic
  object problem by semantic guidance.
\newblock In {\em ECCV}, 2020.

\bibitem{klodt2018supervising}
Maria Klodt and Andrea Vedaldi.
\newblock Supervising the new with the old: learning {SFM} from {SFM}.
\newblock In {\em ECCV}, 2018.

\bibitem{kumar2018depthnet}
Arun~CS Kumar, Suchendra~M Bhandarkar, and Mukta Prasad.
\newblock {DepthNet}: A recurrent neural network architecture for monocular
  depth prediction.
\newblock In {\em CVPR Workshops}, 2018.

\bibitem{kuznietsov2021comoda}
Yevhen Kuznietsov, Marc Proesmans, and Luc Van~Gool.
\newblock {CoMoDA}: Continuous monocular depth adaptation using past
  experiences.
\newblock In {\em WACV}, 2021.

\bibitem{laidlow2019deepfusion}
Tristan Laidlow, Jan Czarnowski, and Stefan Leutenegger.
\newblock {DeepFusion}: real-time dense {3D} reconstruction for monocular
  {SLAM} using single-view depth and gradient predictions.
\newblock In {\em ICRA}, 2019.

\bibitem{li2020unsupervised}
Hanhan Li, Ariel Gordon, Hang Zhao, Vincent Casser, and Anelia Angelova.
\newblock Unsupervised monocular depth learning in dynamic scenes.
\newblock In {\em CoRL}, 2020.

\bibitem{li2019mannequin}
Zhengqi Li, Tali Dekel, Forrester Cole, Richard Tucker, Noah Snavely, Ce Liu,
  and William~T. Freeman.
\newblock Learning the depths of moving people by watching frozen people.
\newblock In {\em CVPR}, 2019.

\bibitem{liang2018learning}
Zhengfa Liang, Yiliu Feng, Yulan Guo, Hengzhu Liu, Wei Chen, Linbo Qiao, Li
  Zhou, and Jianfeng Zhang.
\newblock Learning for disparity estimation through feature constancy.
\newblock In {\em CVPR}, 2018.

\bibitem{liu2019neural}
Chao Liu, Jinwei Gu, Kihwan Kim, Srinivasa~G Narasimhan, and Jan Kautz.
\newblock Neural {RGB}$\rightarrow${D} sensing: Depth and uncertainty from a
  video camera.
\newblock In {\em CVPR}, 2019.

\bibitem{long2020occlusion}
Xiaoxiao Long, Lingjie Liu, Christian Theobalt, and Wenping Wang.
\newblock Occlusion-aware depth estimation with adaptive normal constraints.
\newblock In {\em ECCV}, 2020.

\bibitem{luo2019every}
Chenxu Luo, Zhenheng Yang, Peng Wang, Yang Wang, Wei Xu, Ram Nevatia, and Alan
  Yuille.
\newblock Every pixel counts++: Joint learning of geometry and motion with {3D}
  holistic understanding.
\newblock {\em PAMI}, 2019.

\bibitem{luo2020consistent}
Xuan Luo, Jia-Bin Huang, Richard Szeliski, Kevin Matzen, and Johannes Kopf.
\newblock Consistent video depth estimation.
\newblock In {\em ACM SIGGRAPH}, 2020.

\bibitem{martins2018fusion}
Diogo Martins, Kevin Van~Hecke, and Guido De~Croon.
\newblock Fusion of stereo and still monocular depth estimates in a
  self-supervised learning context.
\newblock In {\em ICRA}, 2018.

\bibitem{mayer2015large}
Nikolaus Mayer, Eddy Ilg, Philip H{\"a}usser, Philipp Fischer, Daniel Cremers,
  Alexey Dosovitskiy, and Thomas Brox.
\newblock A large dataset to train convolutional networks for disparity,
  optical flow, and scene flow estimation.
\newblock In {\em CVPR}, 2016.

\bibitem{mccraith2020monocular}
Robert McCraith, Lukas Neumann, Andrew Zisserman, and Andrea Vedaldi.
\newblock Monocular depth estimation with self-supervised instance adaptation.
\newblock {\em arXiv:2004.05821}, 2020.

\bibitem{newcombe2010live}
Richard~A Newcombe and Andrew~J Davison.
\newblock Live dense reconstruction with a single moving camera.
\newblock In {\em CVPR}, 2010.

\bibitem{newcombe2011kinectfusion}
Richard~A Newcombe, Shahram Izadi, and Otmar Hilliges.
\newblock Kinectfusion: Real-time dense surface mapping and tracking.
\newblock In {\em UIST}, 2011.

\bibitem{newcombe2011dtam}
Richard~A Newcombe, Steven~J Lovegrove, and Andrew~J Davison.
\newblock {DTAM}: Dense tracking and mapping in real-time.
\newblock In {\em ICCV}, 2011.

\bibitem{patil2020dont}
Vaishakh Patil, Wouter Van~Gansbeke, Dengxin Dai, and Luc Van~Gool.
\newblock Don't forget the past: Recurrent depth estimation from monocular
  video.
\newblock In {\em IEEE Robotics and Automation Letters}, 2020.

\bibitem{pilzer2018unsupervised}
Andrea Pilzer, Dan Xu, Mihai~Marian Puscas, Elisa Ricci, and Nicu Sebe.
\newblock Unsupervised adversarial depth estimation using cycled generative
  networks.
\newblock In {\em 3DV}, 2018.

\bibitem{poggi2018towards}
Matteo Poggi, Filippo Aleotti, Fabio Tosi, and Stefano Mattoccia.
\newblock Towards real-time unsupervised monocular depth estimation on {CPU}.
\newblock In {\em IROS}, 2018.

\bibitem{ranjan2018adversarial}
Anurag Ranjan, Varun Jampani, Kihwan Kim, Deqing Sun, Jonas Wulff, and
  Michael~J Black.
\newblock Competitive collaboration: Joint unsupervised learning of depth,
  camera motion, optical flow and motion segmentation.
\newblock In {\em CVPR}, 2019.

\bibitem{russakovsky2015imagenet}
Olga Russakovsky, Jia Deng, Hao Su, Jonathan Krause, Sanjeev Satheesh, Sean Ma,
  Zhiheng Huang, Andrej Karpathy, Aditya Khosla, Michael Bernstein, et~al.
\newblock Imagenet large scale visual recognition challenge.
\newblock {\em IJCV}, 2015.

\bibitem{saxena2007depth}
Ashutosh Saxena, Jamie Schulte, and Andrew~Y Ng.
\newblock Depth estimation using monocular and stereo cues.
\newblock In {\em IJCAI}, 2007.

\bibitem{shu2020feature}
Chang Shu, Kun Yu, Zhixiang Duan, and Kuiyuan Yang.
\newblock Feature-metric loss for self-supervised learning of depth and
  egomotion.
\newblock In {\em ECCV}, 2020.

\bibitem{smolyanskiy2018importance}
Nikolai Smolyanskiy, Alexey Kamenev, and Stan Birchfield.
\newblock On the importance of stereo for accurate depth estimation: An
  efficient semi-supervised deep neural network approach.
\newblock In {\em CVPR Workshops}, 2018.

\bibitem{tateno2017cnn}
Keisuke Tateno, Federico Tombari, Iro Laina, and Nassir Navab.
\newblock Cnn-slam: Real-time dense monocular slam with learned depth
  prediction.
\newblock In {\em CVPR}, 2017.

\bibitem{tosi2020distilled}
Fabio Tosi, Filippo Aleotti, Pierluigi~Zama Ramirez, Matteo Poggi, Samuele
  Salti, Luigi~Di Stefano, and Stefano Mattoccia.
\newblock Distilled semantics for comprehensive scene understanding from
  videos.
\newblock In {\em CVPR}, 2020.

\bibitem{uhrig2017sparse}
Jonas Uhrig, Nick Schneider, Lukas Schneider, Uwe Franke, Thomas Brox, and
  Andreas Geiger.
\newblock Sparsity invariant {CNNs}.
\newblock In {\em 3DV}, 2017.

\bibitem{ummenhofer2017demon}
Benjamin Ummenhofer, Huizhong Zhou, Jonas Uhrig, Nikolaus Mayer, Eddy Ilg,
  Alexey Dosovitskiy, and Thomas Brox.
\newblock {DeMoN}: Depth and motion network for learning monocular stereo.
\newblock In {\em CVPR}, 2017.

\bibitem{vijayanarasimhan2017sfm}
Sudheendra Vijayanarasimhan, Susanna Ricco, Cordelia Schmid, Rahul Sukthankar,
  and Katerina Fragkiadaki.
\newblock {SfM-Net}: Learning of structure and motion from video.
\newblock {\em arXiv:1704.07804}, 2017.

\bibitem{wang2020self}
Jianrong Wang, Ge Zhang, Zhenyu Wu, XueWei Li, and Li Liu.
\newblock Self-supervised joint learning framework of depth estimation via
  implicit cues.
\newblock {\em arXiv:2006.09876}, 2020.

\bibitem{wang2018mvdepthnet}
Kaixuan Wang and Shaojie Shen.
\newblock {MVDepthNet}: Real-time multiview depth estimation neural network.
\newblock In {\em 3DV}, 2018.

\bibitem{wang2019recurrent}
Rui Wang, Stephen~M Pizer, and Jan-Michael Frahm.
\newblock Recurrent neural network for (un-)supervised learning of monocular
  video visual odometry and depth.
\newblock In {\em CVPR}, 2019.

\bibitem{wang2019unos}
Yang Wang, Peng Wang, Zhenheng Yang, Chenxu Luo, Yi Yang, and Wei Xu.
\newblock {UnOS}: Unified unsupervised optical-flow and stereo-depth estimation
  by watching videos.
\newblock In {\em CVPR}, 2019.

\bibitem{watson2019depthints}
Jamie Watson, Michael Firman, Gabriel Brostow, and Daniyar Turmukhambetov.
\newblock Self-supervised monocular depth hints.
\newblock In {\em ICCV}, 2019.

\bibitem{wei2019deepsfm}
Xingkui Wei, Yinda Zhang, Zhuwen Li, Yanwei Fu, and Xiangyang Xue.
\newblock {DeepSFM}: Structure from motion via deep bundle adjustment.
\newblock In {\em ECCV}, 2020.

\bibitem{wimbauer2020monorec}
Felix Wimbauer, Nan Yang, Lukas von Stumberg, Niclas Zeller, and Daniel
  Cremers.
\newblock {MonoRec}: Semi-supervised dense reconstruction in dynamic
  environments from a single moving camera.
\newblock In {\em CVPR}, 2021.

\bibitem{wu2019spatial}
Zhenyao Wu, Xinyi Wu, Xiaoping Zhang, Song Wang, and Lili Ju.
\newblock Spatial correspondence with generative adversarial network: Learning
  depth from monocular videos.
\newblock In {\em ICCV}, 2019.

\bibitem{xie2016deep3d}
Junyuan Xie, Ross Girshick, and Ali Farhadi.
\newblock {Deep3D}: Fully automatic {2D-to-3D} video conversion with deep
  convolutional neural networks.
\newblock In {\em ECCV}, 2016.

\bibitem{xie2019video}
Jiaxin Xie, Chenyang Lei, Zhuwen Li, Li~Erran Li, and Qifeng Chen.
\newblock Video depth estimation by fusing flow-to-depth proposals.
\newblock In {\em IROS}, 2020.

\bibitem{yang2018polarimetric}
Luwei Yang, Feitong Tan, Ao Li, Zhaopeng Cui, Yasutaka Furukawa, and Ping Tan.
\newblock Polarimetric dense monocular {SLAM}.
\newblock In {\em CVPR}, 2018.

\bibitem{yang2018lego}
Zhenheng Yang, Peng Wang, Yang Wang, Wei Xu, and Ram Nevatia.
\newblock {LEGO}: Learning edge with geometry all at once by watching videos.
\newblock In {\em CVPR}, 2018.

\bibitem{yao2018mvsnet}
Yao Yao, Zixin Luo, Shiwei Li, Tian Fang, and Long Quan.
\newblock Mvsnet: Depth inference for unstructured multi-view stereo.
\newblock In {\em ECCV}, 2018.

\bibitem{geonet2018}
Zhichao Yin and Jianping Shi.
\newblock {GeoNet}: Unsupervised learning of dense depth, optical flow and
  camera pose.
\newblock In {\em CVPR}, 2018.

\bibitem{zhanst2018}
Huangying Zhan, Ravi Garg, Chamara~Saroj Weerasekera, Kejie Li, Harsh Agarwal,
  and Ian Reid.
\newblock Unsupervised learning of monocular depth estimation and visual
  odometry with deep feature reconstruction.
\newblock In {\em CVPR}, 2018.

\bibitem{Zhang2019GANet}
Feihu Zhang, Victor Prisacariu, Ruigang Yang, and Philip~HS Torr.
\newblock {GA-Net}: Guided aggregation net for end-to-end stereo matching.
\newblock In {\em CVPR}, 2019.

\bibitem{zhang2019domaininvariant}
Feihu Zhang, Xiaojuan Qi, Ruigang Yang, Victor Prisacariu, Benjamin Wah, and
  Philip Torr.
\newblock Domain-invariant stereo matching networks.
\newblock In {\em ECCV}, 2020.

\bibitem{zhang2009consistent}
Guofeng Zhang, Jiaya Jia, Tien-Tsin Wong, and Hujun Bao.
\newblock Consistent depth maps recovery from a video sequence.
\newblock {\em PAMI}, 2009.

\bibitem{zhang2019exploiting}
Haokui Zhang, Chunhua Shen, Ying Li, Yuanzhouhan Cao, Yu Liu, and Youliang Yan.
\newblock Exploiting temporal consistency for real-time video depth estimation.
\newblock In {\em ICCV}, 2019.

\bibitem{zhong2017self}
Yiran Zhong, Yuchao Dai, and Hongdong Li.
\newblock Self-supervised learning for stereo matching with self-improving
  ability.
\newblock {\em arXiv:1709.00930}, 2017.

\bibitem{zhou2017unsupervised}
Tinghui Zhou, Matthew Brown, Noah Snavely, and David Lowe.
\newblock Unsupervised learning of depth and ego-motion from video.
\newblock In {\em CVPR}, 2017.

\end{thebibliography}
}

\clearpage

\begin{appendices}
\section{Additional Qualitative Results}

\paragraph{YouTube videos.}
We use our model trained on KITTI to make predictions on frames from a YouTube video, downloaded from the Wind Walk Travel Videos channel\footnote{\url{https://www.youtube.com/channel/UCPur06mx78RtwgHJzxpu2ew}}.
We include some of these predictions in Figure~\ref{fig:wander} to demonstrate the ability of our KITTI-trained model to transfer between different datasets. 
We also include predictions from an equivalent KITTI-trained Monodepth2 model \cite{godard2019digging}.
This monocular only baseline fails to pick out some of the details, as it cannot take advantage of the multiple test-time viewpoints.

\paragraph{KITTI dataset.}
Additional qualitative results on the KITTI dataset are shown here in Figures~\ref{fig:extra_kitti_qual_1}, \ref{fig:extra_kitti_qual_2}, and \ref{fig:extra_kitti_qual_3}. 
Ground truth depth maps and the error maps use the improved, densified, ground truth from \cite{uhrig2017sparse}.
We also show a colorbar for the error maps in Figure~\ref{fig:colormap}.
The error maps show the absolute relative error in depth, computed as
\begin{align}
    \text{abs\_rel\_error} = \big| \frac{D_\text{pred} - D_\text{gt}}{D_\text{gt}} \big|.
    \label{eqn:abs_rel_error}
\end{align}


\newcommand{\itemwidth}{0.35\columnwidth}

\begin{figure*}
  \centering
  \resizebox{1.0\textwidth}{!}{
        \begin{tabular}{@{\hskip 2mm}c@{\hskip 2mm}c@{\hskip 2mm}c@{\hskip 2mm}c@{}}
        
        {\scriptsize Input} &  {\scriptsize ManyDepth} & {\scriptsize Monodepth2 \cite{godard2019digging}} \\
        \includegraphics[width=\itemwidth]{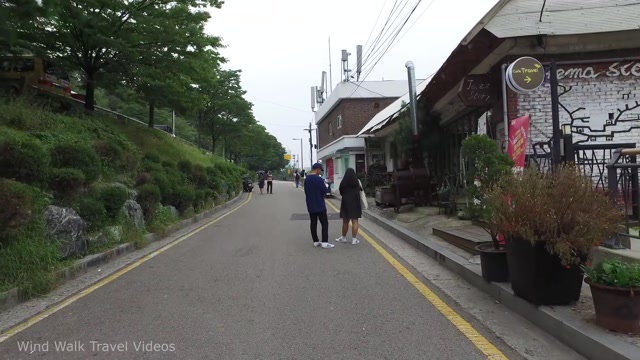} &
        \includegraphics[width=\itemwidth]{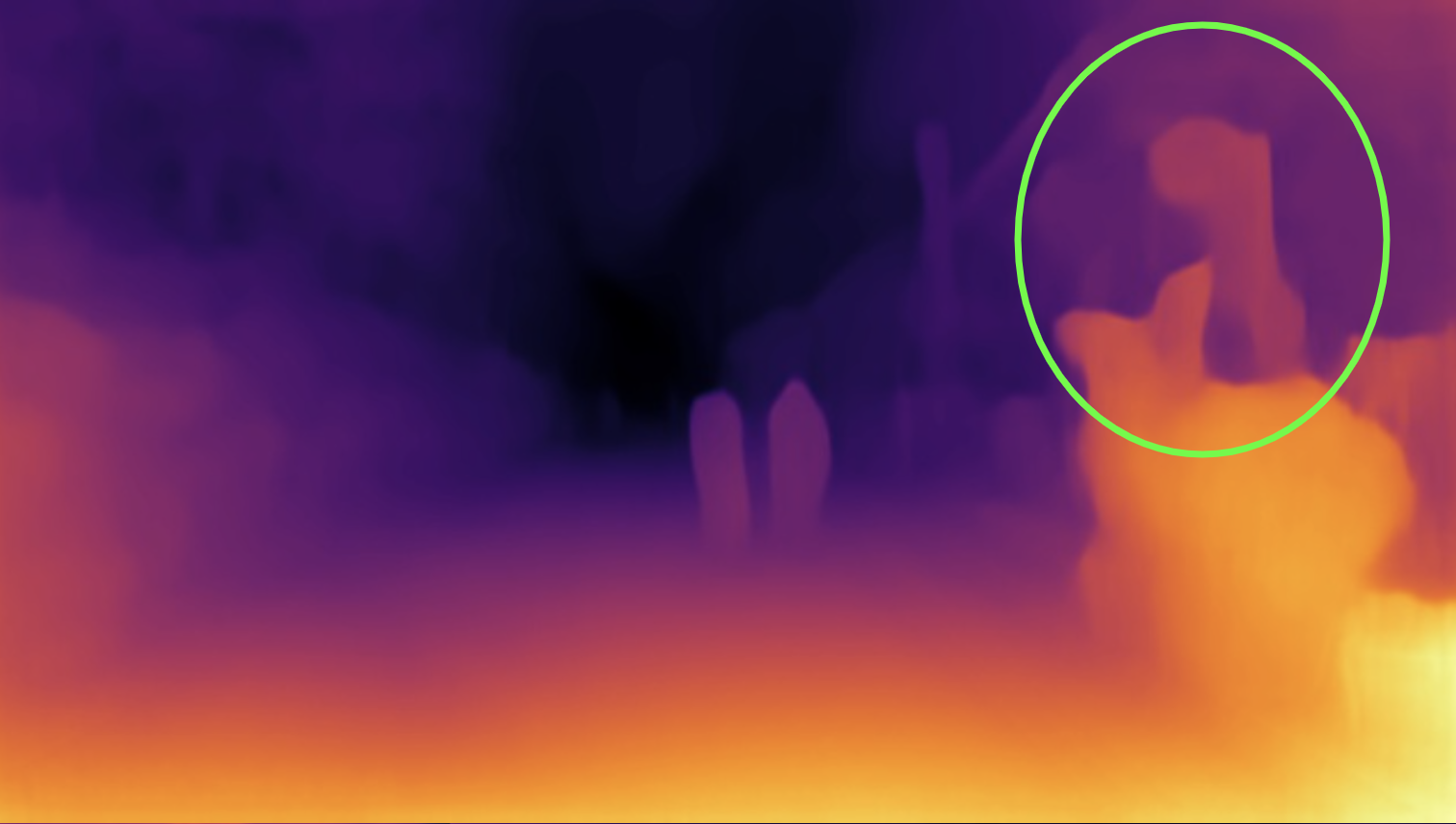} &
        \includegraphics[width=\itemwidth]{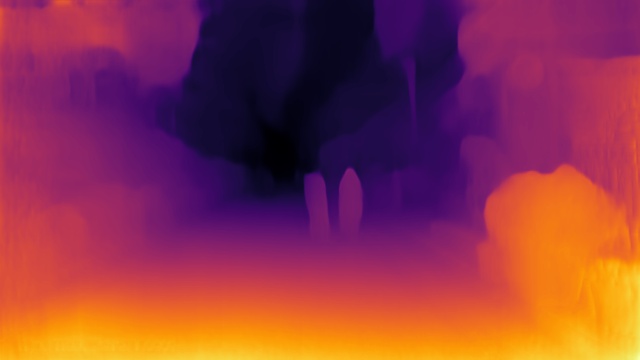} \\
        
        \includegraphics[width=\itemwidth]{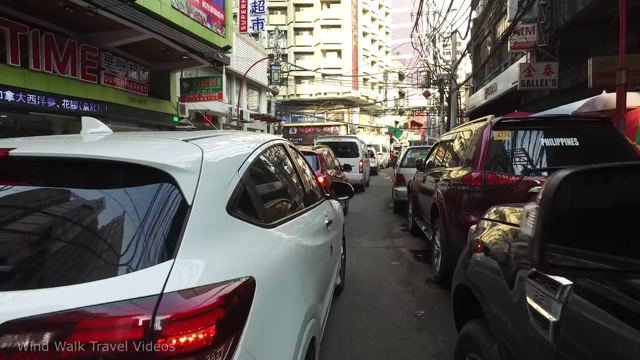} &
        \includegraphics[width=\itemwidth]{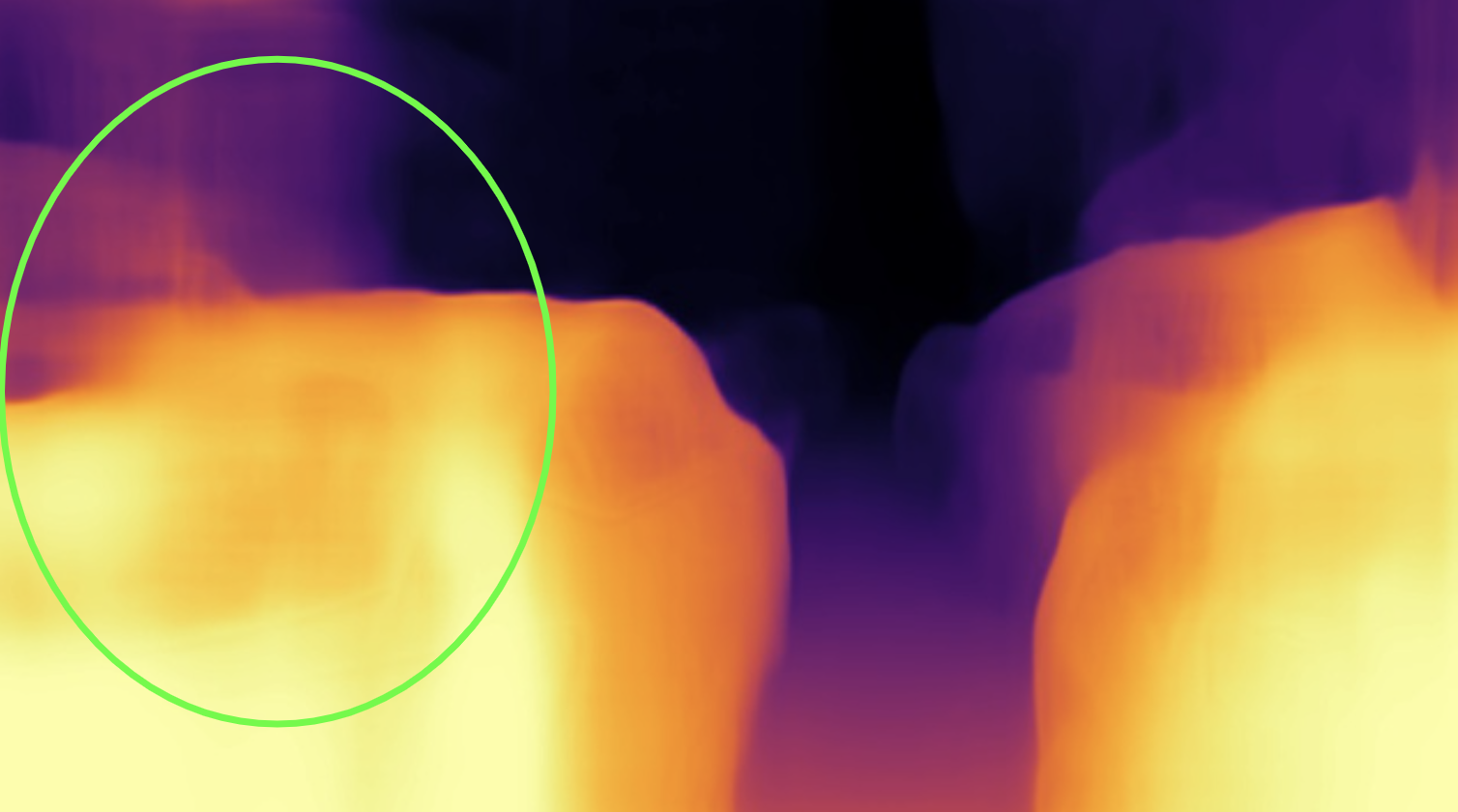} &
        \includegraphics[width=\itemwidth]{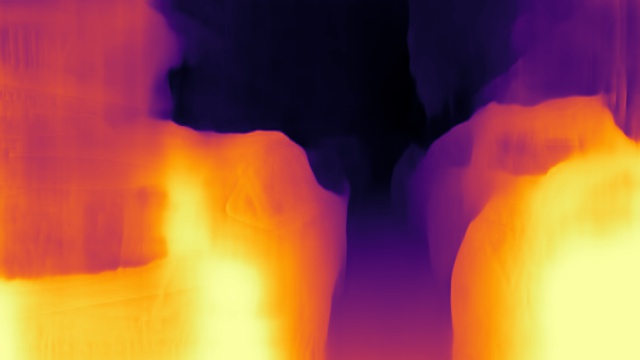} \\
        
        \includegraphics[width=\itemwidth]{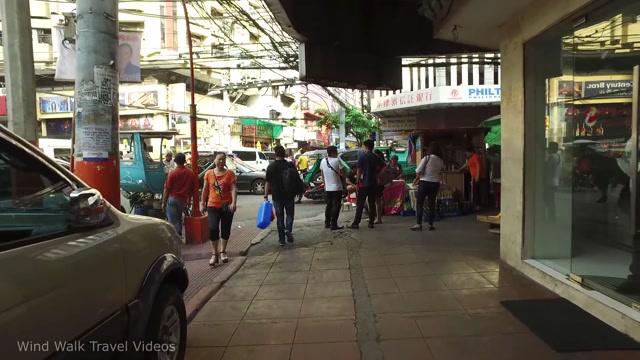} &
        \includegraphics[width=\itemwidth]{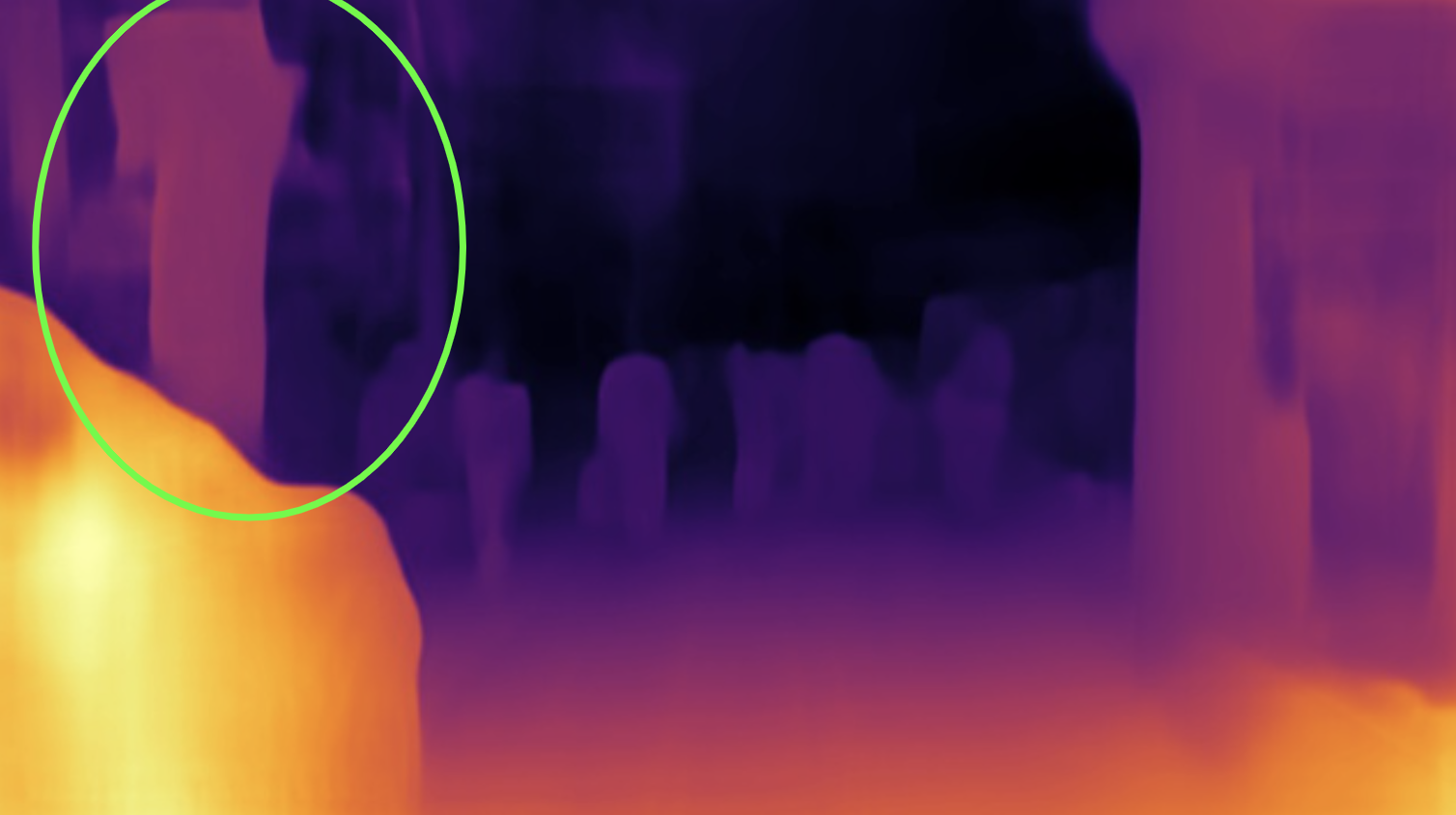} &
        \includegraphics[width=\itemwidth]{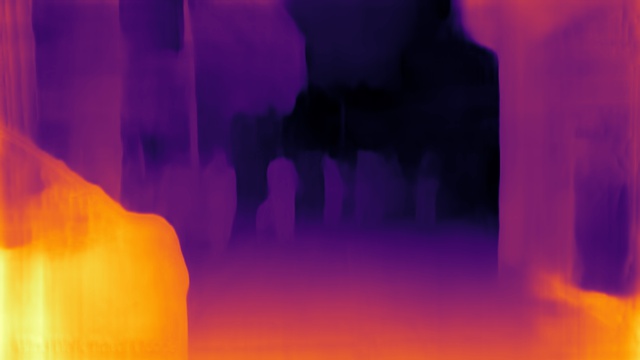} \\
        
        \includegraphics[width=\itemwidth]{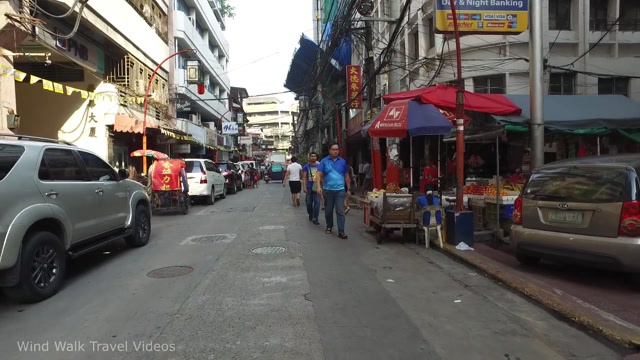} &
        \includegraphics[width=\itemwidth]{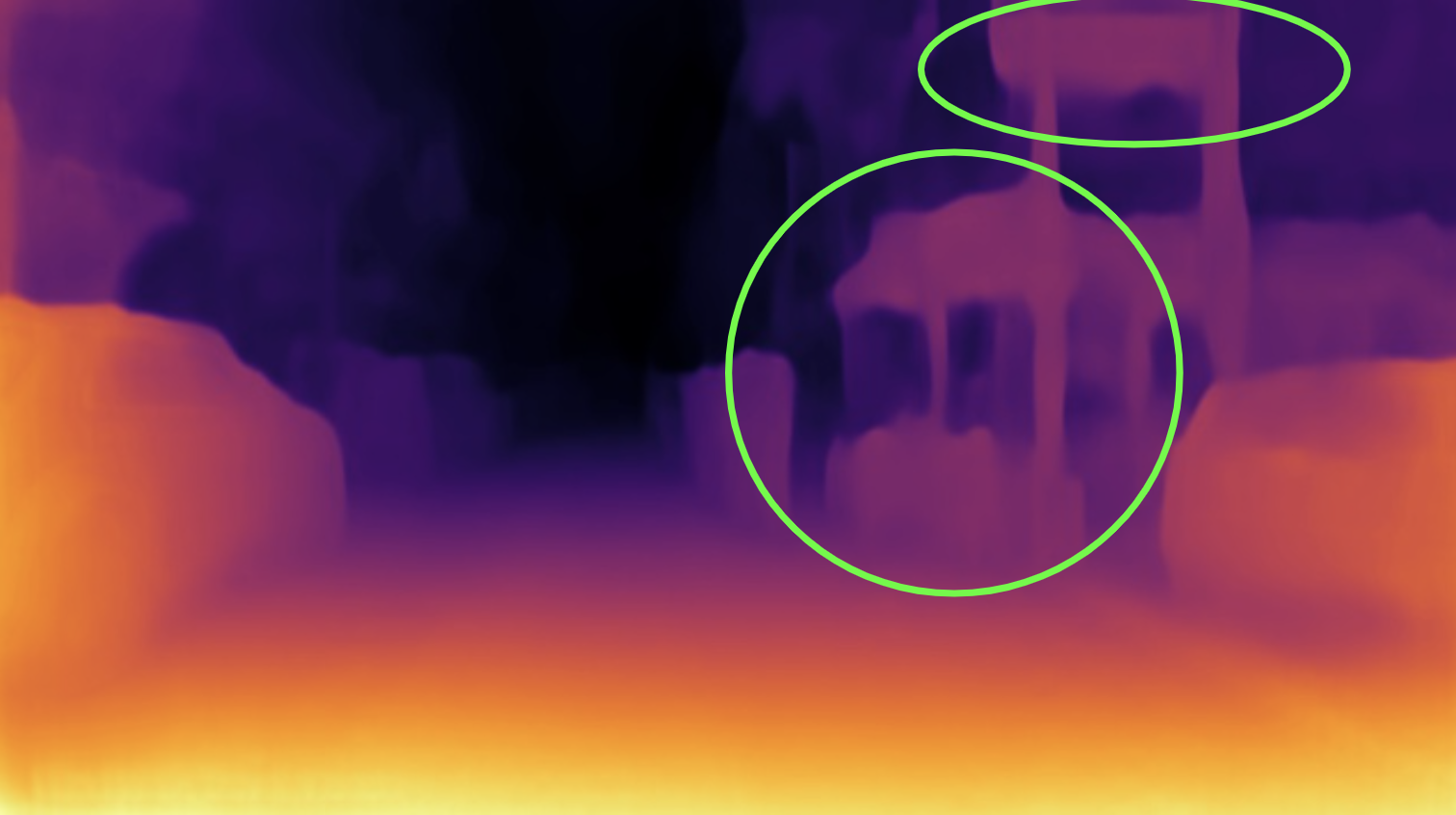} &
        \includegraphics[width=\itemwidth]{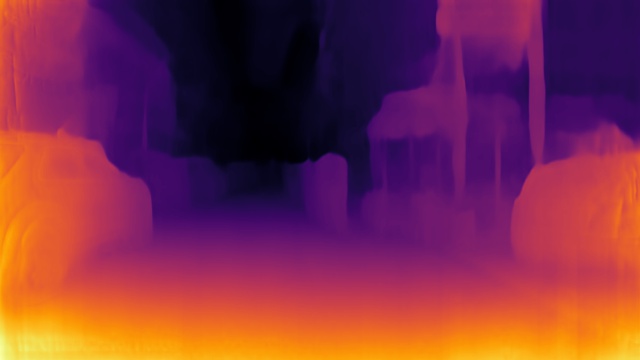} \\
        
        \includegraphics[width=\itemwidth]{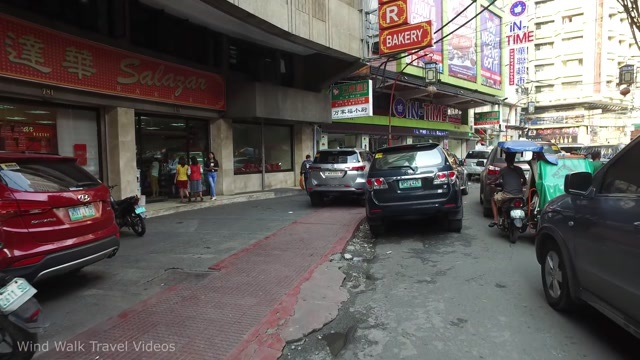} &
        \includegraphics[width=\itemwidth]{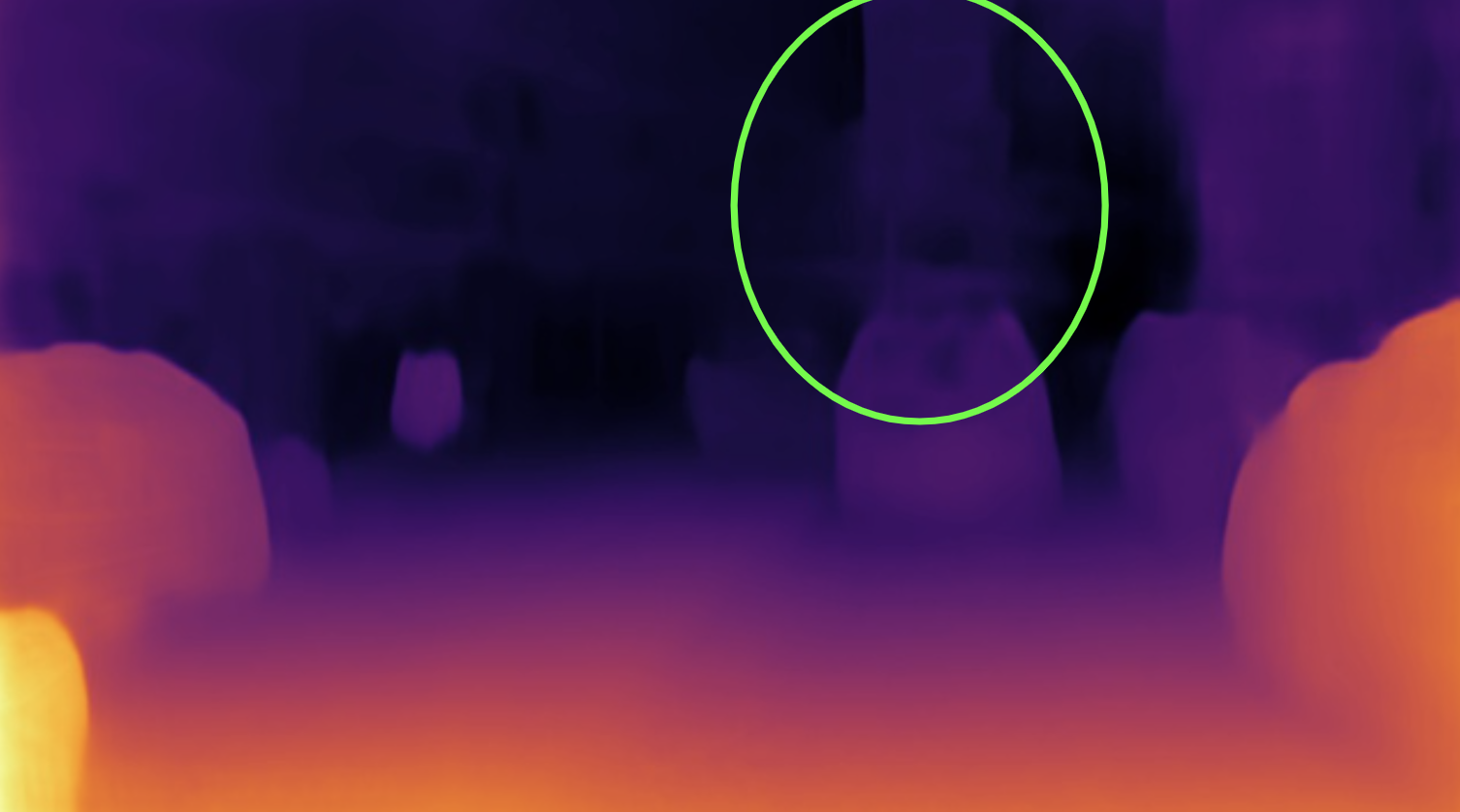} &
        \includegraphics[width=\itemwidth]{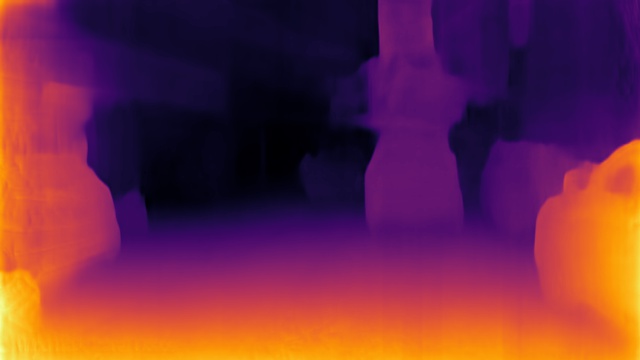} \\
        
    \end{tabular}
    }
    \caption{\textbf{Qualitative results on YouTube videos}. All predictions are made with a model trained on the KITTI dataset. Note how our method is able to correctly estimate the shape of the sign in the top row, the car on the left in the second row, the pillars in the third row, and the signs in the fourth and fifth rows. \label{fig:wander}}
    
\end{figure*}

\section{Architecture details}
\label{sec:architecture_details}

Here we describe the architecture of our model. 
Unless otherwise specified, we use a ResNet18~\cite{he2016deep} as the basis for our depth network. 
See Figure~\ref{fig:cost_volume_building} from the main paper for a graphical depiction. 
Our feature extractor evaluates up until the output of the first residual layer of the ResNet18, resulting in a tensor which is a quarter of the input resolution with 64 channels. 
These features are then aggregated into a cost volume for each input frame, using 96 depth bins.
The output of the cost volume is then concatenated with the output of the feature extractor for the image at $t=0$, resulting in a residual like combination. 
The depth decoder evaluates the remaining residual layers and then decodes the output back to the input resolution over the course of several convolution and upsampling layers while making use of skip connections. 
We use the same pose network used in \cite{godard2019digging}. 
A full description of our architecture is in Table~\ref{tab:network_arch}.
Finally, our consistency network $\teachernetwork$ uses the standard architecture from \cite{godard2019digging} with no additional modifications.

\begin{table}[!h]
  \centering
  \resizebox{\linewidth}{!}{
  \begin{tabular}[t]{|l|l|l|l|l|l|l|}
\hline
\textbf{layer} & \textbf{k} & \textbf{s} & \textbf{chns} & \textbf{res} & \textbf{input}   & \textbf{activation}    \\ \hline
\hline
\multicolumn{7}{|l|}{\textbf{Feature Extractor}} \\
\hline
conv1      &   7    &  2     &     64  &   2  &        image at time t             & ReLU  \\

max\_pool1      &       & 2     &       & 4   &        conv1              &   \\
resnet\_layer1      &       &       &     64  &   4  &        max\_pool1              &   \\ 

\hline
\multicolumn{7}{|l|}{} \\ \hline
\multicolumn{7}{|l|}{\textbf{Cost Volume}} \\
\hline
cost\_vol      &       &       &  96   & 4 &  resnet\_layer1 for t $\in \{0, -1\}$                        &   \\
conv\_reduce      &     3  &    1   & 64 &    4   &  resnet\_layer1\_t=0, cost\_vol &  ReLU                    \\

\hline
\multicolumn{7}{|l|}{} \\ \hline
\multicolumn{7}{|l|}{\textbf{Depth Decoder}} \\
\hline
resnet\_layer2      &       &       &     128  &   8  &       conv\_reduce              &   \\
resnet\_layer3      &       &       &     256  &   16  &        resnet\_layer2              &   \\
resnet\_layer4      &       &       &     512  &   32  &        resnet\_layer3              &   \\ \hline

upconv5       & 3      & 1      & 256      & 32    & resnet\_layer4                     & ELU \\
iconv5        & 3      & 1      & 256      & 16    & $\uparrow$upconv5, resnet\_layer3 & ELU \\ \hline

upconv4       & 3      & 1      & 128      & 16    & iconv5                    & ELU \\
iconv4        & 3      & 1      & 128      & 8     & $\uparrow$upconv4, resnet\_layer2 & ELU \\
disp4         & 3      & 1      & 1        & 1     & iconv4                    & Sigmoid \\ \hline

upconv3       & 3      & 1      & 64       & 8     & iconv4                    & ELU \\
iconv3        & 3      & 1      & 64       & 4     & $\uparrow$upconv3, resnet\_layer1 & ELU \\
disp3         & 3      & 1      & 1        & 1     & iconv3                   & Sigmoid  \\ \hline

upconv2       & 3      & 1      & 32       & 4     & iconv3                    & ELU \\
iconv2        & 3      & 1      & 32       & 2     & $\uparrow$upconv2, conv1 & ELU \\
disp2         & 3      & 1      & 1        & 1     & iconv2                    & Sigmoid \\ \hline

upconv1       & 3      & 1      & 16       & 2     & iconv2                    & ELU \\
iconv1        & 3      & 1      & 16       & 1     & $\uparrow$upconv1         & ELU \\
disp1         & 3      & 1      & 1        & 1     & iconv1                    & Sigmoid \\ \hline
\end{tabular} 
 }
  \vspace{0pt}
  \caption{\textbf{Our network architecture.} Here \textbf{k} is the kernel size, \textbf{s} the stride, \textbf{chns} the number of output channels for each layer, \textbf{res} is the downscaling factor for each layer relative to the input image, and \textbf{input} corresponds to the input of each layer where $\uparrow$ is a $2\times$ nearest-neighbor upsampling of the layer.
  resnet\_layer are the standard residual layers as defined in \cite{he2016deep}.
  The feature extractor is applied to each of the input frames and the resulting features are combined in the cost volume. 
    \label{tab:network_arch} 
  }
  \vspace{-5pt}
\end{table}

\section{Efficiency computations}
\label{sec:macs}

In Table~\ref{tab:macs} we show the computations used to generate the graph in Figure~5 in the main paper.
Multiply-add computations (MACs) were generated with the THOP library~\footnote{\url{https://github.com/Lyken17/pytorch-OpCounter}}, using PyTorch version 1.5.1, on a GeForce GTX 1080Ti GPU.
To benchmark the models we used authors' released code where available, or generated equivalent models using descriptions from the paper.
Test-time refinement methods involve multiple passes through the networks.
In these cases we count one update step as two passes through the relevant networks: a forward and a backward pass.
This means that a model which makes `20 steps of test-time optimization' makes a total of 40 passes through the encoder, decoder and pose network.
\cite{luo2020consistent} doesn't use a pose network, nor does it use image reconstruction losses, so we give it $0$ MACs for those categories.
Patil \etal~\cite{patil2020dont} uses an LSTM and so relies on sequences of frames.
For this method we conservatively ignore the computations required by the LSTM modules, as code from \cite{patil2020dont} was not available at the time of submission.


\section{Additional results on KITTI dataset}

\subsection{KITTI benchmark scores}

In Table~\ref{tab:kitti_improved_ground_truth} we present KITTI results evaluated using the improved dense ground truth from \cite{uhrig2017sparse}, in the manner described in \cite{godard2019digging}.
Only a subset of baselines are evaluated here, as many baselines do not report these numbers, nor do they provide precomputed depth predictions on the KITTI dataset.

\subsection{Models trained from scratch}

Table~\ref{tab:kitti_nopt} shows KITTI results trained without ImageNet pretraining.
Scores are slightly worse than our `with pretraining' results, but are still competitive.

\subsection{Additional KITTI ablation scores}

In Table~\ref{tab:kitti_eigen_ablations_sup_mat} we show full metrics for ablations, performed on the KITTI Eigen test set.
These experiments are described in detail in the main paper.
We note that while the motion masking method of \cite{godard2019digging} is competitive with our approach on KITTI, we show in our Cityscapes ablations (Table~\ref{tab:cs_ablations_sup_mat}) how, when more moving objects are present, our masking scheme is significantly better than that of \cite{godard2019digging}.

\subsection{Full scores for augmentation ablation}

Table~\ref{tab:kitti_augmentation_ablations_sup_mat} we show the full set of metrics for the experiment described in Section~\ref{sec_kitti_ab} of the main paper, and presented in Table~\ref{tab:kitti_eigen_ablations} of the main paper.

\section{Additional results on Cityscapes dataset}

\subsection{Alternative crop evaluation schemes}
\label{sec:cityscapes_cropping}
Past literature has used different cropping schemes for evaluating on the Cityscapes dataset.
In Table~\ref{tab:cs_train_test_sup_mat} we show our Cityscapes results, evaluated with these two different schemes:
\begin{description}
    \item[Evaluation cropping scheme `A':] Here, we crop the center 50\% of pixels vertically, reducing the image size from $2048\times1024$ to $2048\times512$. Next we remove a 192 pixels from the left and the right of the image, giving a final evaluation region of $1664\times512$. This is the evaluation scheme used by \cite{ Casser_2019_CVPR_Workshops, li2020unsupervised, gordon2019depth}, following conversations with the authors.
    \item[Evaluation cropping scheme `B':] Here, we only evaluate the top 75\% of the image, ignoring the bottom 25\% of pixels. This leaves an evaluation region of size $2048\times768$. This effectively crops out the `car' from the bottom of the image.  This evaluation method is the suitable for evaluating works which also train with this same cropping scheme, \eg \cite{zhou2017unsupervised, yang2018lego, geonet2018}.
\end{description}

\subsection{Cityscapes ablation}
In Table~\ref{tab:cs_ablations_sup_mat} we show an ablation of our masking schemes on Cityscapes~\cite{Cordts2016Cityscapes}, mirroring similar experiments on KITTI in the main paper.
Again we see that our contributions help. 
We do not use the masking scheme of \cite{ranjan2018adversarial}, as they do not provide trained models on Cityscapes.
The improvements that our masking provides, compared to the baselines and \cite{godard2019digging}, is superior on this dataset compared with KITTI.
Cityscapes contains more moving objects, so benefits more from our training scheme.

\section{Note on Cityscapes training of Monodepth2}
\label{sec:cityscapes_monodepth}

In Table~\ref{tab:cs_train_test} of the main paper, we show the scores on the Cityscapes dataset for our implementation of Monodepth2~\cite{godard2019digging}. To obtain these, we trained for 5 epochs with a batch size of 12, mirroring the training procedure of our consistency network. This is to account for the far larger number of images in the Cityscapes training dataset compared to KITTI. We note that when trained for a full 20 epochs, the resulting scores are significantly worse. 

\begin{figure}[b]
    \centering
    \includegraphics[width=0.8\columnwidth]{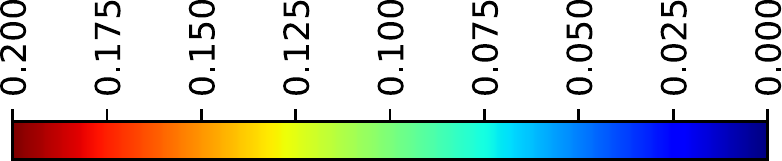}
    \caption{Color scale used for all error plots in the paper, supplementary material, and  video. Values on the color axis are units of absolute relative error, applied after median scaling of the predicted depth. See Eqn.~\ref{eqn:abs_rel_error} for details on how to compute this error.}
    \label{fig:colormap}
\end{figure}

\begin{table*}
  \centering
  \vspace{-2pt}
  \footnotesize
  \begin{tabular}{|l||c|c|c|c|c|c|c|}
  \hline
     Ablation & \cellcolor{col1}Abs Rel & \cellcolor{col1}Sq Rel & \cellcolor{col1}RMSE 
     & \cellcolor{col1}RMSE log & \cellcolor{col2}$\delta < 1.25 $ & \cellcolor{col2}$\delta < 1.25^{2}$ & \cellcolor{col2}$\delta < 1.25^{3}$ \\
  \hline
  ManyDepth full &   \textbf{0.098}  &   \textbf{0.770}  &   4.459  &   \textbf{0.176}  &   \textbf{0.900}  &   \textbf{0.965}  &   \textbf{0.983} \\

  ManyDepth (w/o motion masking) &   0.113  &   1.354  &   5.228  &   0.196  &   0.885  &   0.956  &   0.978  \\
  
  ManyDepth (w/o motion masking, w/o augmentation) & 0.284  &  11.240  &   8.516  &   0.297  &   0.813  &   0.896  &   0.934  \\ 
  
  ManyDepth (with motion masking but no teacher) &   0.154  &   2.682  &   6.573  &   0.236  &   0.842  &   0.933  &   0.965  \\
  
    Stack of 2 frames as input &   0.121  &   1.028  &   5.016 & 0.198  &   0.868  &   0.956  &   0.979\\
    
    ManyDepth (with motion masking from \cite{ranjan2018adversarial}) & 0.099  &   0.783  &   \textbf{4.447}  &   \textbf{0.176}  &   0.899  &   {0.964}  &   \textbf{0.983} \\
    
    ManyDepth (with motion masking from \cite{godard2019digging})  & 0.099  &   0.780  &   4.465  &   0.178  &   0.899  &   0.963  &   0.982  \\
    
    \hline
  ManyDepth with 3-frame input (-2, -1, 0) &   0.098  &   0.780  &   \textbf{4.430}  &   0.176  &   \textbf{0.902}  &   \textbf{0.964}  &   \textbf{0.983} \\
  ManyDepth with 3-frame input (-1, 0, +1) &   \textbf{0.097}  &   \textbf{0.768}  &   4.431  &   \textbf{0.175}  &   0.900  &   \textbf{0.964}  &   \textbf{0.983} \\
\hline
  \end{tabular}
  \vspace{1pt}
  \caption{\textbf{Full metrics from KITTI 2015 ablation (Table 4 in the main paper).} Here we ablate our method on KITTI 2015 \cite{Geiger2012CVPR} using the Eigen split, reporting all seven metrics. 
  Details for these experiments are given in the main paper. 
  \label{tab:kitti_eigen_ablations_sup_mat}}
\end{table*}

\begin{table*}
  \centering
  \vspace{-2pt}
  \footnotesize
  \begin{tabular}{|l||c|c|c|c|c|c|c|}
  \hline
     Ablation & \cellcolor{col1}Abs Rel & \cellcolor{col1}Sq Rel & \cellcolor{col1}RMSE 
     & \cellcolor{col1}RMSE log & \cellcolor{col2}$\delta < 1.25 $ & \cellcolor{col2}$\delta < 1.25^{2}$ & \cellcolor{col2}$\delta < 1.25^{3}$ \\
  \hline
  Monodepth2 HR (no pt)$\dagger$ & 0.131  &   1.063  &   5.112  &   0.208  &   0.851  &   0.950  &   0.977 \\

  Ours HR (no pt) & 0.104  &   0.844  &   4.598  &   0.185  &   0.889  &   0.959  &   0.981 \\

  Monodepth2 HR & 0.115  &   0.882  &   4.701  &   0.190  &   0.879  &   0.961  &   0.982 \\
  
    Ours HR & \textbf{0.093}  &   \textbf{0.715}  &   \textbf{4.245}  &   \textbf{0.172}  &   \textbf{0.909}  &   \textbf{0.966}  &   \textbf{0.983} \\
    
    \hline

  \end{tabular}
  \vspace{1pt}
  \caption{\textbf{Comparing the effect of ImageNet pretraining} on the KITTI 2015 depth dataset. For most of our experiments we follow \cite{godard2019digging, watson2019depthints, wang2020self, guo2018learning} etc.~ in using weights pretrained on ImageNet. As expected, pretraining improves scores, but we are still extremely competitive without it. $\dagger$ is trained by us using the authors' code. 
  \label{tab:kitti_nopt}}
\end{table*}

\begin{table*}
  \centering
  \vspace{-2pt}
  \footnotesize
  \begin{tabular}{|l||c|c|c|c|c|c|c|}
  \hline
     Ablation & \cellcolor{col1}Abs Rel & \cellcolor{col1}Sq Rel & \cellcolor{col1}RMSE 
     & \cellcolor{col1}RMSE log & \cellcolor{col2}$\delta < 1.25 $ & \cellcolor{col2}$\delta < 1.25^{2}$ & \cellcolor{col2}$\delta < 1.25^{3}$ \\
  \hline

  ManyDepth (w/o motion masking) & 0.185  &   4.080  &   8.499  &   0.240  &   0.803  &   0.921  &   0.960 \\
    
    ManyDepth (with motion masking from \cite{godard2019digging}) & 0.143  &   2.300  &   7.073  &   0.199  &   0.844  &   0.951  &   0.977 \\
    
    ManyDepth full & 0.114  &   1.193  &   6.223  &   0.170  &   0.875  &   0.967  &   0.989  \\ 
    \hline

  \end{tabular}
  \vspace{1pt}
  \caption{\textbf{Ablation on the Cityscapes dataset.} Here we ablate our method on the Cityscapes dataset, evaluating as described in the main paper. Our motion masking contributions make significantly larger improvements on Cityscapes compared to KITTI, as more moving objects are present in the Cityscapes training and test footage.
  \label{tab:cs_ablations_sup_mat}}
\end{table*}

\begin{table*}
  \centering
  \footnotesize

\begin{tabular}{|l|l||c|c|c|c|c|c|c|}
  \hline

     Test-time input & Model & \cellcolor{col1}Abs Rel & \cellcolor{col1}Sq Rel & \cellcolor{col1}RMSE   & \cellcolor{col1}RMSE log & \cellcolor{col2}$\delta < 1.25 $ & \cellcolor{col2}$\delta < 1.25^{2}$ & \cellcolor{col2}$\delta < 1.25^{3}$ \\
     \hline\hline
  \multirow{2}{*}{Standard: $(I_{t-1}, I_t)$} & No augmentation & 0.100  &   0.794  &   4.432  &   0.179  &   0.895  &   0.962  &   0.982 \\
   & ManyDepth &   {0.098}  &   0.770  &   4.459  &   {0.176}  &   {0.900}  &   {0.965}  &   0.983 \\
  
  \hline
  \multirow{3}{*}{Start-of-sequence: $(I_t)$} 
  & Monodepth2 \cite{godard2019digging} & 0.115 & 0.903 & 4.863 & 0.193 & 0.877 & 0.959 & 0.981 \\
  & No augmentation & 0.148  &   1.076  &   5.161  &   0.219  &   0.812  &   0.943  &   0.977 \\ 
   & ManyDepth & 0.118  &   0.892  &   4.764  &   0.192  &   0.871  &   0.959  &   0.982\\
  
  \hline
  \multirow{2}{*}{Static camera: $(I_t, I_t)$} & No augmentation & 0.158  &   1.132  &   5.228  &   0.225  &   0.794  &   0.939  &   0.977 \\
   & ManyDepth & 0.117  &   0.886  &   4.754 &   0.191  &   0.872  &   0.959  &   0.982 \\
\hline
  \end{tabular}
  \vspace{1pt}
  \caption{\textbf{Full metrics for `our augmentations help' (Table 5 in main paper).}
  We compare two variants of our model, one trained with our novel data augmentations (`Ours') and one without.
  We create two artificial scenarios to test each model's performance on start-of-sequence images (where we just input $I_t$) and static cameras (where both input frames are the exact same).
  \label{tab:kitti_augmentation_ablations_sup_mat}}
\end{table*}

\begin{table*}
  \centering
  \footnotesize
  \begin{threeparttable}
  \resizebox{1.0\textwidth}{!}{
  \begin{tabular}{|l|c|c||r|c|r|c|r|c|r|c||r|c|}
  \hline
																											
&		&		&	\multicolumn{2}{c|}{Encoder}			&	\multicolumn{2}{c|}{Decoder}			&	\multicolumn{2}{c|}{PoseCNN}			&	\multicolumn{2}{c||}{Losses}			&		\\								
\hline												
Method	&	WxH	&	Test frames	&	MACs	&	Passes	&	MACs	&	Passes	&	MACs	&	Passes	&	MACs	&	Passes	&	Total MACs & Abs.~rel.	\\						
\hline \hline 
ManyDepth (MR) single-frame	&	640x192	&	1	&	4.7	&	1	&	3.6	&	1	&	0.0	&	0	&	0.0	&	0	&	8.3	&	0.118	\\						
ManyDepth (MR) multi-frame	&	640x192	&	2	&	6.6	&	1\tnote{a}	&	3.6	&	1	&	4.9	&	1	&	0.0	&	0	&	15.1	&	0.098	\\						
\midline
ManyDepth (HR) single-frame	&	1024x320	&	1	&	12.6	&	1	&	9.5	&	1	&	0.0	&	0	&	0.0	&	0	&	22.2	&	0.11	\\						
ManyDepth (HR) multi-frame	&	1024x320	&	2	&	17.6	&	1\tnote{a}	&	9.5	&	1	&	13.1	&	1	&	0.0	&	0	&	40.2	&	0.093	\\						
\midline
Shu \etal~\cite{shu2020feature}	&	1024x320	&	1	&	17.6	&	1	&	58.3	&	1	&	0.0	&	0	&	0.0	&	0	&	75.8	&	0.104	\\						
Shu \etal~\cite{shu2020feature}	&	1024x320	&	3	&	26.8	&	40	&	58.3	&	40	&	4.9	&	40	&	26.8	&	20	&	4137.1	&	0.088	\\						
\midline
Struct2depth (M)  \cite{casser2018depth}	&	416x128	&	1	&	1.9	&	1	&	1.5	&	1	&	0.0	&	0	&	0.0	&	0	&	3.5	&	0.141	\\						
Struct2depth (M+R)  \cite{casser2018depth}	&	416x128	&	3	&	1.9	&	40	&	1.5	&	40	&	0.2	&	40	&	0.0	&	20	&	145.4	&	0.109	\\						
\midline
Chen \etal~\cite{chen2019self}	&	416x128	&	1	&	1.9	&	1	&	1.5	&	1	&	0.0	&	1	&	0.0	&	0	&	3.5	&	0.135	\\						
Chen \etal~\cite{chen2019self}	&	416x128	&	3	&	1.9	&	100	&	1.5	&	100	&	0.2	&	100	&	0.0	&	50	&	363.6	&	0.099	\\						
\midline
PacknetSfM~\cite{guizilini20203d}	&	640x192	&	1	&	186.8	&	1	&	18.5	&	1	&	0.0	&	0	&	0.0	&	0	&	205.3	&	0.111	\\						
\midline																															
Luo \etal~\cite{luo2020consistent}	&	384x112	&	N\tnote{c}	&	26.5	&	40	&	0.0	&	40	&	0.0	&	0	&	0.0\tnote{b}	&	0	&	1062.6	&	0.13	\\	
\midline
PyDNet \cite{poggi2018towards}	&	512x256	&	1	&	4.5	&	1	&	0.4	&	1	&	0.0	&	0	&	0.0	&	0	&	4.9	&	0.153	\\		\midline				
Wang \etal~\cite{wang2020self}	&	640x192	&	1	&	4.5	&	1	&	3.6	&	1	&	0.0	&	0	&	0.0	&	0	&	8.0	&	0.112	\\						
Wang \etal~\cite{wang2020self}	&	640x192	&	2	&	4.5	&	1	&	3.6	&	1	&	0.4	&	1	&	0.0	&	0	&	8.4	&	0.106	\\				\midline		
Monodepth2~\cite{godard2019digging}	&	640x192	&	1	&	4.5	&	1	&	3.6	&	1	&	0.0	&	0	&	0.0	&	0	&	8.0	&	0.115	\\					\midline	

Patil \etal~\cite{patil2020dont}	&	640x192	&	N	&	4.5	&	3\tnote{d}	&	3.6	&	1	&	0.0	&	0	&	0.0	&	0	&	16.9	&	0.112	\\						
Patil \etal~\cite{patil2020dont}	&	640x192	&	N	&	4.5	&	1\tnote{d}	&	3.6	&	1	&	0.0	&	0	&	0.0	&	0	&	8.0	&	0.112	\\						
\hline

	\end{tabular}	}
	\begin{tablenotes}
    \item[a] One pass through our multi-frame encoder includes feature extraction from source and target frames plus cost volume building and further convolutions. \\ 
    \item[b] Luo \etal \cite{luo2020consistent} doesn't use image reconstruction loss, so loss MACs negligible. \\
    \item[c] Luo \etal \cite{luo2020consistent} operates on whole sequences; we report numbers for a single frame of the sequence. \\
    \item[d] We give two numbers for LSTM method \cite{patil2020dont}; one with a single network pass, the other assuming three encoder passes needed (\ie for a sequence).
    \end{tablenotes}
    \end{threeparttable}
    \vspace{4pt}
	\caption{\textbf{Details on multiply-add computations (MACs) calculations}. For our multi-frame model, `Encoder' MACs includes performing feature extraction for both inputs frames, creating the cost volume and performing the final ResNet18 encoder layers.
	See text for more details.
	\label{tab:macs}}
\end{table*}

\newcommand{\splitlineblack}{\arrayrulecolor{black}\hhline{~-----------}}

\begin{table*}
  \centering
  \resizebox{1.0\textwidth}{!}{
  \begin{tabular}{|c|l|c|c|c||c|c|c|c|c|c|c|}
\arrayrulecolor{black}\hline
& Method & Test frames &  WxH & Evaluation crop & \cellcolor{col1}Abs Rel & \cellcolor{col1}Sq Rel & \cellcolor{col1}RMSE  & \cellcolor{col1}RMSE log & \cellcolor{col2}$\delta < 1.25 $ & \cellcolor{col2}$\delta < 1.25^{2}$ & \cellcolor{col2}$\delta < 1.25^{3}$ \\
  \hline
\arrayrulecolor{black}\hline\arrayrulecolor{gray}
\multirow{3}{2cm}{Semantic \\ supervision} 

& Struct2Depth 2 \cite{Casser_2019_CVPR_Workshops}  &  1  & 416\x128 &  A &
0.145  & 1.737  & 7.280  &  0.205 & 0.813 & 0.942 & 0.976 \\

& Struct2Depth 2 \cite{Casser_2019_CVPR_Workshops}  &  3 (-1,  0,  +1)  & 416\x128 &  A &
0.151 & 2.492 & 7.024 & 0.202 & 0.826 & 0.937 & 0.972 \\

& Videos in the Wild \cite{gordon2019depth} & 1 & 416\x128 &  A &
\textbf{0.127} & \textbf{1.330} & \textbf{6.960} & \textbf{0.195} & \textbf{0.830} & \textbf{0.947} & \textbf{0.981} \\

\arrayrulecolor{black}\hline\hline

\multirow{6}{*}{No semantics} 

& Monodepth2$\dagger$ \cite{godard2019digging} & 1  & 416\x128 &  A &
0.129  &   1.569  &   6.876  &   0.187  &   0.849  &   0.957  &   0.983 \\

&  Monodepth2$\dagger$ \cite{godard2019digging} & 1  & 512\x192 &  B &
0.159  &   2.016  &   8.074  &   0.221  &   0.787  &   0.941  &   0.978 \\

& Pilzer \etal \cite{pilzer2018unsupervised} & 1 & 512\x256 &  ? &
0.240 & 4.264 & 8.049 & 0.334 & 0.710 & 0.871 & 0.937 \\

& Struct2Depth 2 \cite{Casser_2019_CVPR_Workshops}  &  3 (-1,  0,  +1)    & 416\x128 &  A &
0.222  & 5.737  & 8.613  & 0.258  & 0.774  & 0.908 & 0.954 \\

& Li \etal \cite{li2020unsupervised} & 1 & 416\x128 & A & 0.119 &  1.290 & {6.980} &  {0.190} & {0.846} &  0.952 &  0.982 \\

\splitlineblack
& {\bf ManyDepth} & 2 (-1, 0)  & 416\x128 &   A &
\textbf{0.114}  &   \textbf{1.193}  &   \textbf{6.223}  &   \textbf{0.170}  &   \textbf{0.875}  &   \textbf{0.967}  &   \textbf{0.989} \\
& {\bf ManyDepth} & 2 (-1, 0)  & 512\x192 &   B &
{0.137} & {1.578} & {7.249} & {0.197} & {0.827}  & {0.954} & {0.985} \\
    
    \arrayrulecolor{black}\hline
  \end{tabular}
  }
  \vspace{-2pt}
  \caption{\textbf{Results on Cityscapes.} 
  Our method beats all competing models. $\dagger$ is trained by us using the authors' code, with the data preprocessing from \cite{zhou2017unsupervised}. See Section~\ref{sec:cityscapes_cropping} for more details. The two alternative cropping evaluation schemes `A' and `B' are described in the text.
  Numbers from Pilzer \etal~\cite{pilzer2018unsupervised} are from their Half-Cycle Mono variant, which is the only one of their models which does not rely on access to stereo pairs at test time.
  \label{tab:cs_train_test_sup_mat}}
  \vspace{-5pt}
\end{table*}

\renewcommand{\midline}{  }

\renewcommand{\splitline}{\arrayrulecolor{black}\hhline{~------------}}

\renewcommand*\rot{\rotatebox{90}}

\definecolor{Asectioncolor}{RGB}{255, 200, 200}
\definecolor{Bsectioncolor}{RGB}{255, 228, 196}
\definecolor{Csectioncolor}{RGB}{235, 255, 235}
\definecolor{Dsectioncolor}{RGB}{235, 235, 255}

\begin{table*}[t]
  \centering
  \footnotesize
  \resizebox{1.0\textwidth}{!}{
    \begin{tabular}{|l|c|l|c|c|c||c|c|c|c|c|c|c|}
        \arrayrulecolor{black}\hline
          &  TTR & Method & Test frames  & Semantics & WxH & \cellcolor{col1}Abs Rel & \cellcolor{col1}Sq Rel & \cellcolor{col1}RMSE  & \cellcolor{col1}RMSE log & \cellcolor{col2}$\delta < 1.25 $ & \cellcolor{col2}$\delta < 1.25^{2}$ & \cellcolor{col2}$\delta < 1.25^{3}$ \\
         
        \hline\hline
        \parbox[b]{2mm}{\multirow{9}{*}{\rotatebox[origin=c]{90}{LR + MR}}} & 
        \cellcolor{Asectioncolor} & Ranjan \cite{ranjan2018adversarial}  & 1 & & 832\x256  &
        0.123 & 0.881 & 4.834 & 0.181 & 0.860 & 0.959 & 0.985\\
        \midline
        & \cellcolor{Asectioncolor} &  EPC++ \cite{luo2019every} & 1  && 832\x256 & 
        0.120 & 0.789 &  4.755  & 0.177  & 0.856 &  0.961 &  0.987\\
        \midline

        &  \cellcolor{Asectioncolor}& Johnston \etal~\cite{johnston2020self} & 1 && 640\x192 &
        0.081 & 0.484  & 3.716  & 0.126  & 0.927  & 0.985  & 0.996 \\
        \midline
        &  \cellcolor{Asectioncolor}& Monodepth2 \cite{godard2019digging} & 1  &&  640\x192 &
         0.090 & 0.545 & 3.942 & 0.137 & 0.914 & 0.983 & 0.995
        \\ 
         \midline
        &  \cellcolor{Asectioncolor}& Packnet-SFM \cite{guizilini20203d} & 1 && 640\x192 &
         0.078 & 0.420 &  3.485 &  0.121 &  0.931 &  0.986 &  0.996
         \\
        \midline
        &  \cellcolor{Asectioncolor}& Patil \etal \cite{patil2020dont}  & N$^\dagger$    & & 640\x192 
        &   0.087  &   0.495  &   3.775  &   0.133  &   0.917  &   0.983  &   0.995  \\
        
        \midline
        &  \cellcolor{Asectioncolor}& Wang \etal \cite{wang2020self} & 2 (-1, 0)    & & 640\x192  & 
        0.082 & 0.462 & 3.739 & 0.127 &  0.923 &  0.984 & 0.996\\
        \midline

        
         & \cellcolor{Asectioncolor} &\textbf{ManyDepth (MR)} & 2 (-1, 0) & & 640\x192 &  \textbf{0.070}  &   \textbf{0.399}  &   \textbf{3.455}  &   \textbf{0.113}  &   \textbf{0.941}  &   \textbf{0.989}  &   \textbf{0.997}\\
        
        \splitline

        & $\bullet$ \cellcolor{Bsectioncolor} & \textbf{{ManyDepth (MR + TTR)}} & 2 (-1, 0) & & 640\x192 & \textbf{0.058}  &   \textbf{0.334}  &   \textbf{3.137}  &   \textbf{0.101}  &   \textbf{0.958}  &   \textbf{0.991}  &   \textbf{0.997} \\
            
        \arrayrulecolor{black}\hline\hline
                
        \parbox[b]{2mm}{\multirow{5}{*}{\rotatebox[origin=c]{90}{HR}}} 
        & \cellcolor{Csectioncolor} &Monodepth2 \cite{godard2019digging}   & 1 &  & 1024\x320 &  0.091  &   0.531  &   3.742  &   0.135  &   0.916  &   0.984  &   0.995  \\

        \midline
        
        & \cellcolor{Csectioncolor}& \textbf{ManyDepth (HR)} & 2 (-1, 0) && 1024\x320 & 0.064  &   0.345  &   \textbf{3.116}  &   0.103  &   0.949  &   0.989  &   \textbf{0.997}  \\
        
        \midline
        
        & \cellcolor{Csectioncolor} & \textbf{ManyDepth (HR ResNet50)} & 2 (-1, 0) && 1024\x320 &   \textbf{0.062}  &   \textbf{0.343}  &   3.139  &   \textbf{0.102}  &   \textbf{0.953}  &   \textbf{0.991}  &   \textbf{0.997} \\

        \splitline
        
        & $\bullet$ \cellcolor{Dsectioncolor} &\textbf{ManyDepth (HR + TTR)} & 2 (-1, 0) && 1024\x320 &   0.056  &   0.322  &   3.034  &   0.096  &   0.961  &   0.992  &   0.997 \\
        
        & $\bullet$ \cellcolor{Dsectioncolor} &\textbf{ManyDepth (HR ResNet50 + TTR)} & 2 (-1, 0) && 1024\x320 &   \textbf{0.055}  &   \textbf{0.305}  &   \textbf{2.945}  &   \textbf{0.094}  &   \textbf{0.963}  &  \textbf{ 0.992}  &   \textbf{0.997} \\

        \arrayrulecolor{black}\hline

    \end{tabular}
  }  
  \vspace{-1pt}
  \caption{\textbf{KITTI~\cite{Geiger2012CVPR} evaluation on improved ground truth from \cite{uhrig2017sparse}, as described in \cite{godard2019digging}.}   As in the main paper, at top we compare medium and low resolution results \colorbox{Asectioncolor}{without}  and \colorbox{Bsectioncolor}{with} test-time refinement (TTR).
  At bottom we compare high resolution results \colorbox{Csectioncolor}{without}  and \colorbox{Dsectioncolor}{with} TTR.
   Best results in each subsection are in \textbf{bold}.
  Our method outperforms all previous methods in all subsections across all metrics, whether or not the baselines use multiple frames at test time.
  We don't report results on the new ground truth using the methods \cite{mccraith2020monocular, casser2018depth, Casser_2019_CVPR_Workshops, gordon2019depth, guizilini2020semantically, chen2019self, luo2020consistent, shu2020feature} as they were not reported in the publications, and the raw predictions have not been made public by the authors. None of the methods presented here were trained with semantics.
  {  
        \footnotesize
        \newline
        \textbf{Legend:} \hspace{10pt} 
        TTR -- Uses test-time refinement \hspace{10pt}
        Semantics -- Semantic supervision \hspace{10pt}
        $^\dagger$ -- evaluated on whole sequences
    } 
    \label{tab:kitti_improved_ground_truth}} 
    
    \vspace{-5pt}

\end{table*}

\newcommand{\turnheightnew}{0.195\columnwidth}
\newcommand{\turnwidthnew}{0.8\columnwidth}

\centering


\newcommand{\imlabel}[2]{\includegraphics[width=\turnwidthnew]{#1}%
\raisebox{35pt}{\makebox[2pt][r]{\small #2}}}


\begin{figure*}
  \centering
  \resizebox{1.0\textwidth}{!}{
        \begin{tabular}{m{3.5cm}m{6cm}m{6cm}}
        
        Input and ground truth &
        \includegraphics[width=\turnwidthnew]{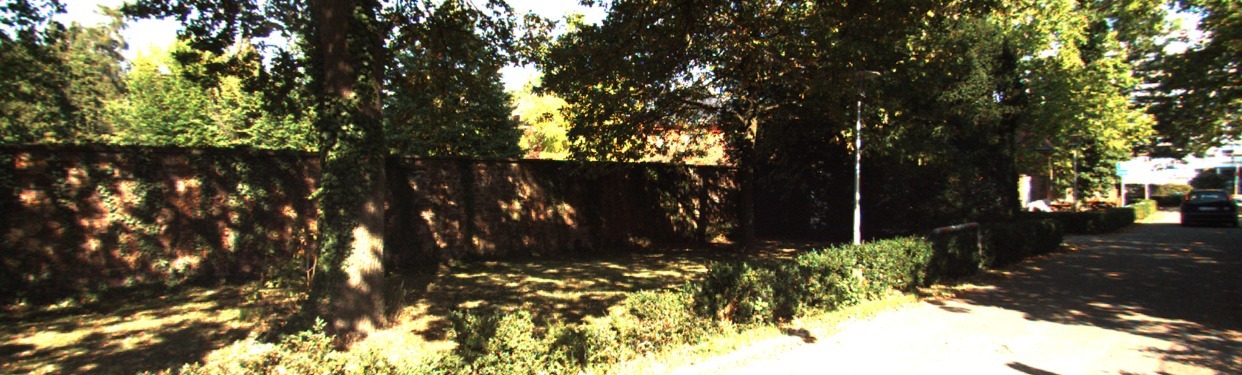} &
        \includegraphics[width=\turnwidthnew]{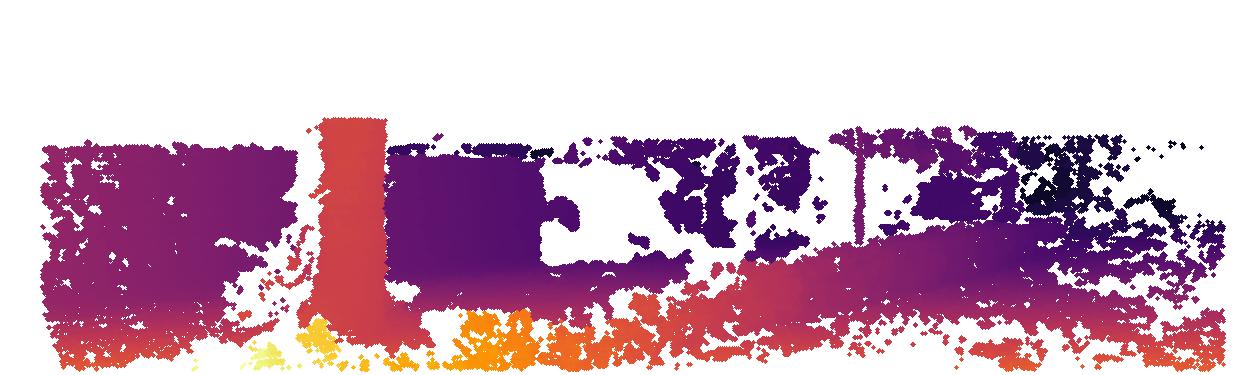} \\
        ManyDepth (MR) &
        \includegraphics[width=\turnwidthnew]{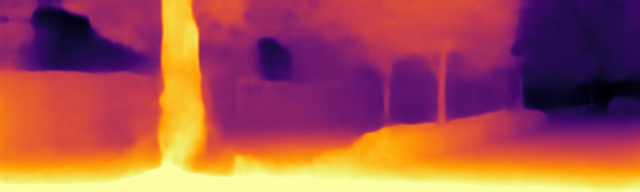} &
        \includegraphics[width=\turnwidthnew]{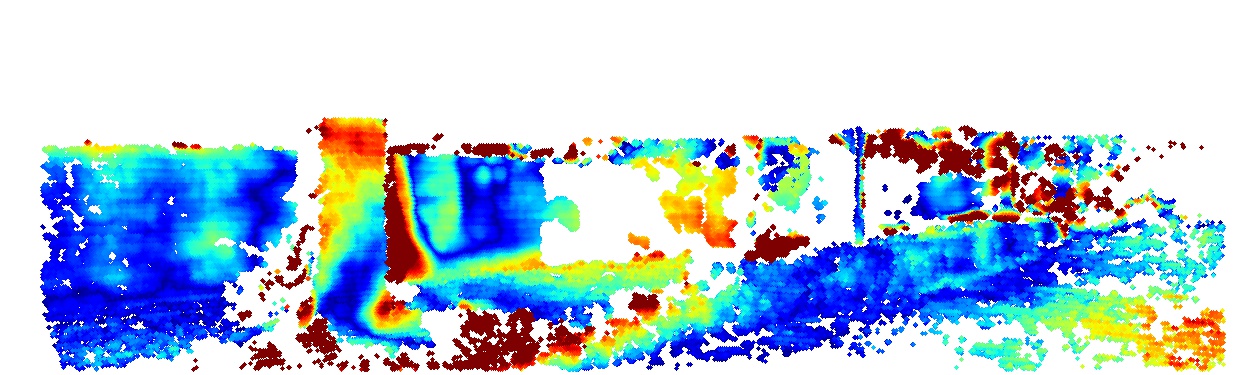} \\
        Monodepth2 \cite{godard2019digging} &
        \includegraphics[width=\turnwidthnew]{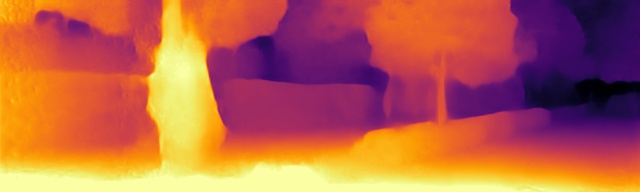} &
        \includegraphics[width=\turnwidthnew]{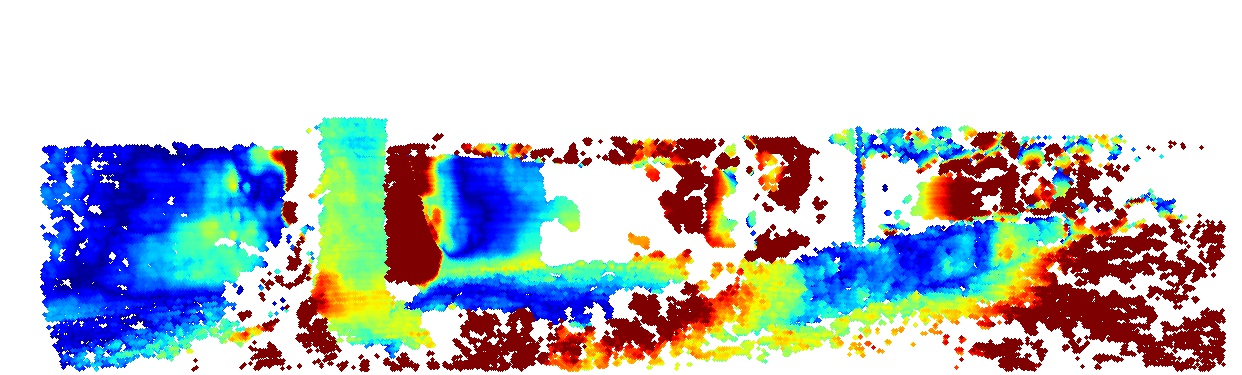} \\
        DFTP \cite{patil2020dont} &
        \includegraphics[width=\turnwidthnew]{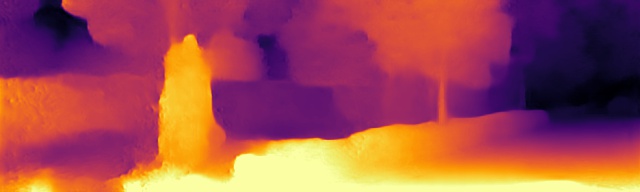} &
        \includegraphics[width=\turnwidthnew]{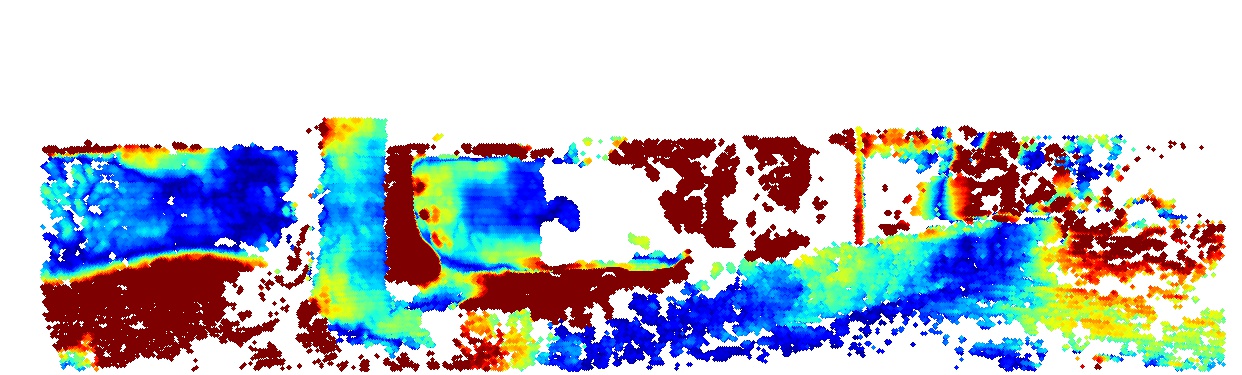} \\
        Packnet-SFM \cite{guizilini20203d} &
        \includegraphics[width=\turnwidthnew]{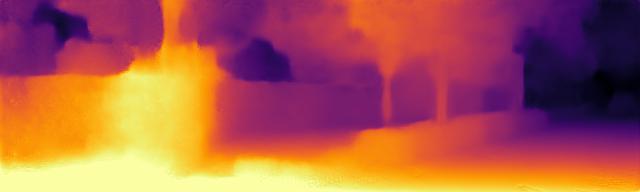} &
        \includegraphics[width=\turnwidthnew]{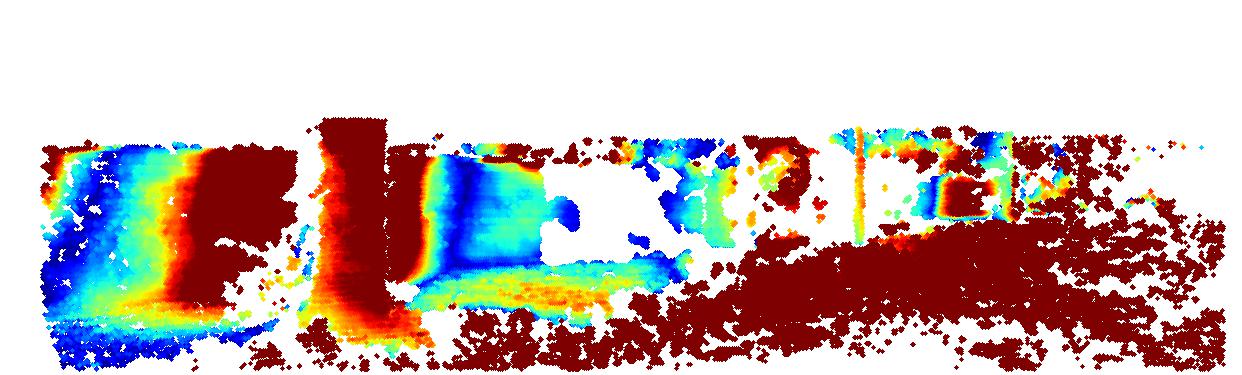} \\
        
        Input and ground truth &
        \includegraphics[width=\turnwidthnew]{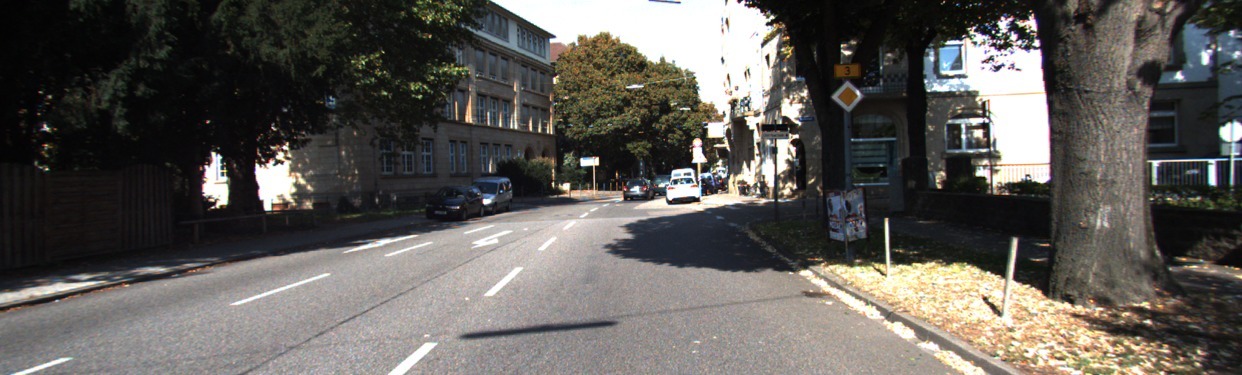} &
        \includegraphics[width=\turnwidthnew]{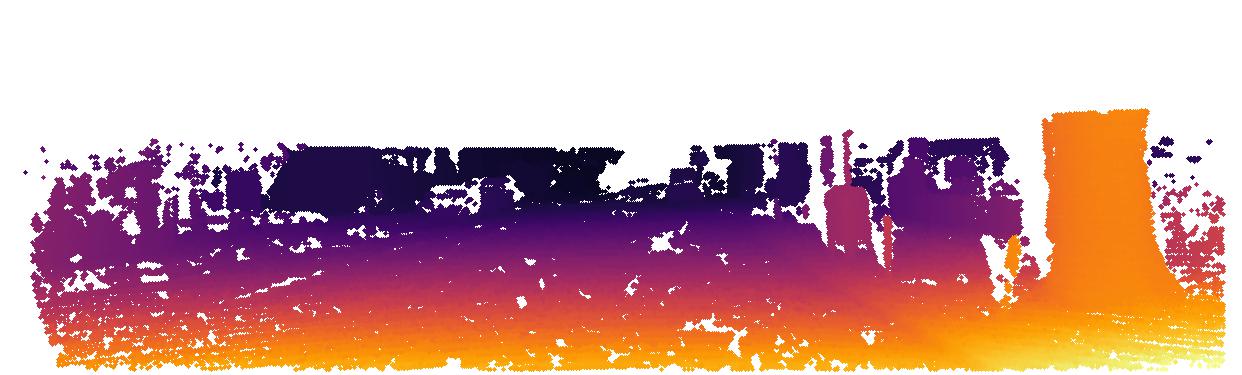} \\
        ManyDepth (MR) &
        \includegraphics[width=\turnwidthnew]{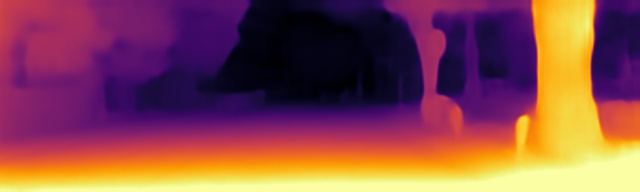} &
        \includegraphics[width=\turnwidthnew]{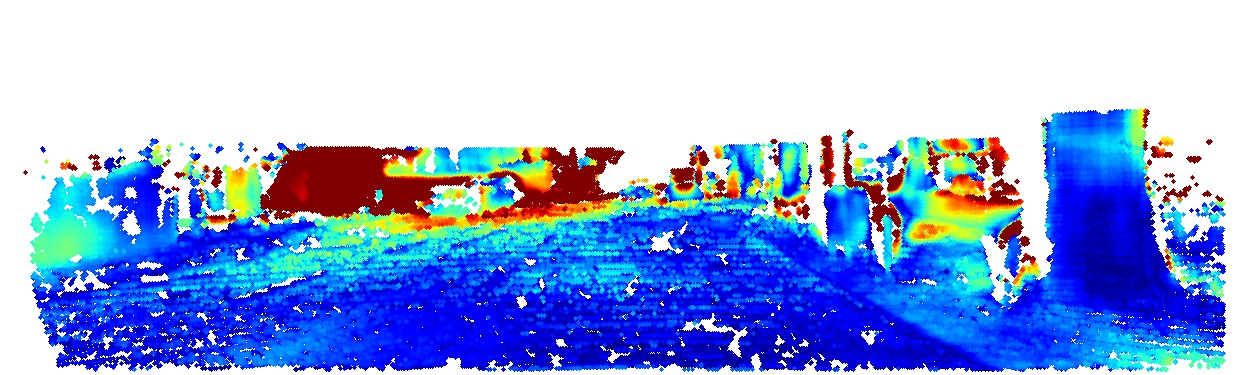} \\
        Monodepth2 \cite{godard2019digging} &
        \includegraphics[width=\turnwidthnew]{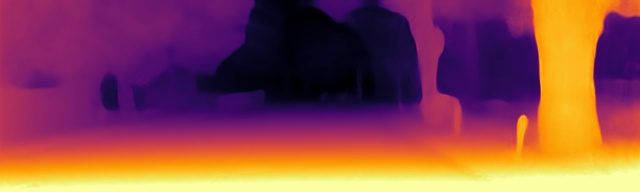} &
        \includegraphics[width=\turnwidthnew]{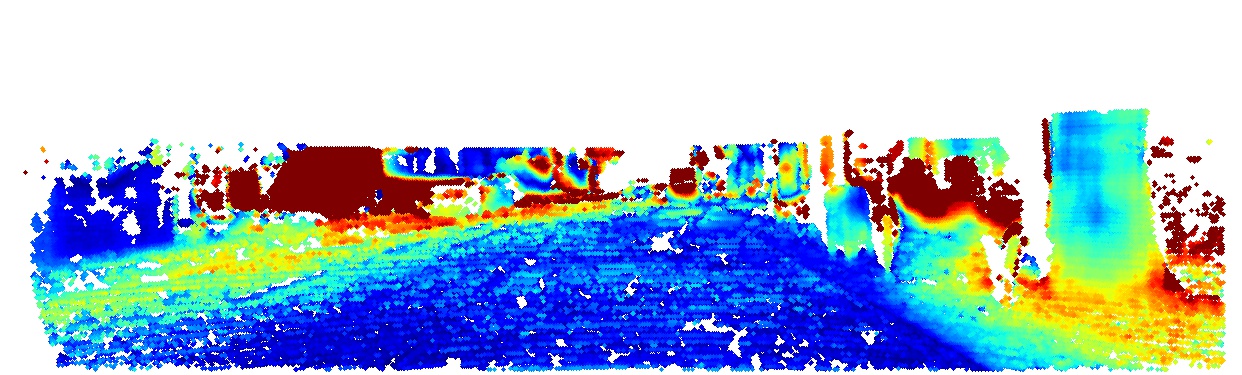} \\
        DFTP \cite{patil2020dont} &
        \includegraphics[width=\turnwidthnew]{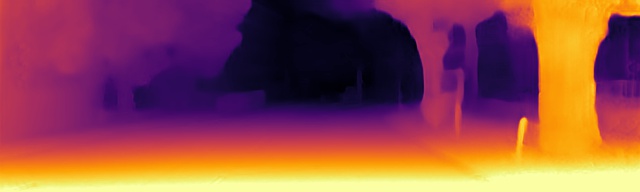} &
        \includegraphics[width=\turnwidthnew]{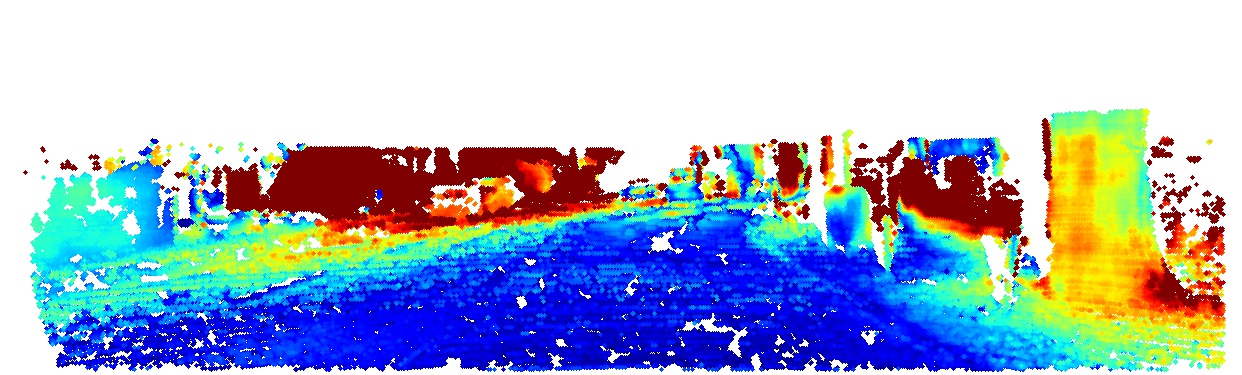} \\
        Packnet-SFM \cite{guizilini20203d} &
        \includegraphics[width=\turnwidthnew]{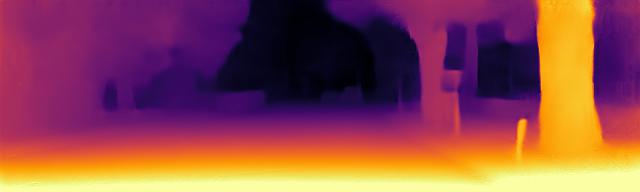} &
        \includegraphics[width=\turnwidthnew]{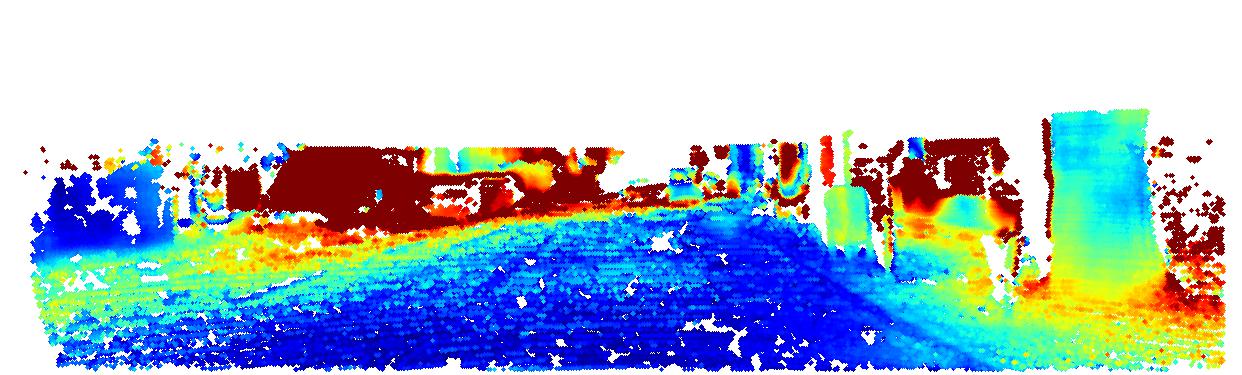} \\
    \end{tabular}
    }
    \caption{\textbf{Further qualitative results on the KITTI dataset (part 1)}. Error maps on the right measure the absolute relative error compared to the ground truth, after median scaling \cite{eigen2015predicting}. Errors range from blue (low error, abs.~rel.~$=0.0$) to red (high error, abs.~rel.~$=0.2$). The color mapping in all error maps are the same, and the colorbar is shown in Figure~\ref{fig:colormap}. \label{fig:extra_kitti_qual_1}}
    
\end{figure*}

\begin{figure*}
    \resizebox{1.0\textwidth}{!}{
        \begin{tabular}{m{3.5cm}m{6cm}m{6cm}}

            Input and ground truth &
            \includegraphics[width=\turnwidthnew]{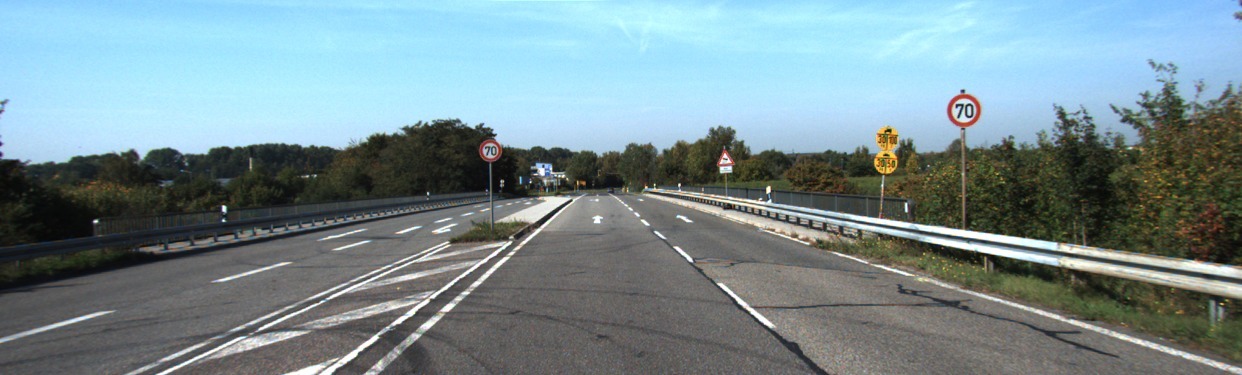} &
            \includegraphics[width=\turnwidthnew]{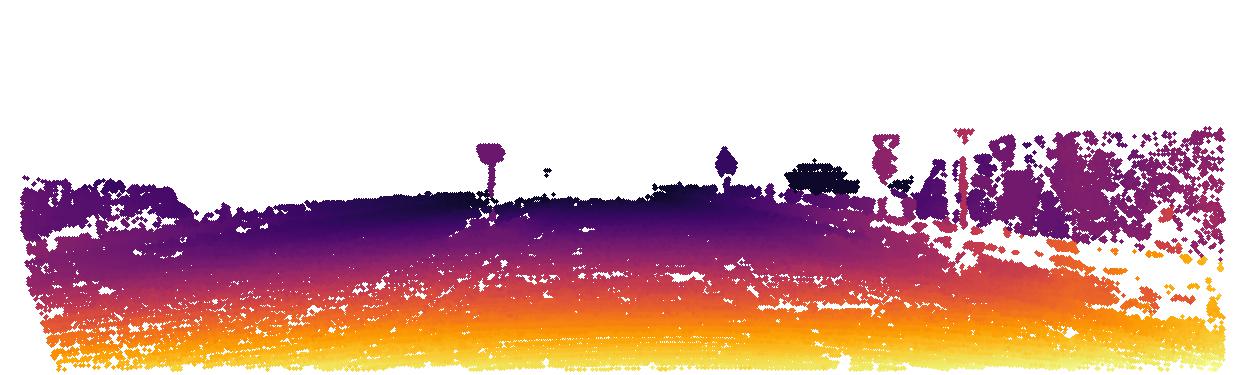} \\
            ManyDepth (MR) &
            \includegraphics[width=\turnwidthnew]{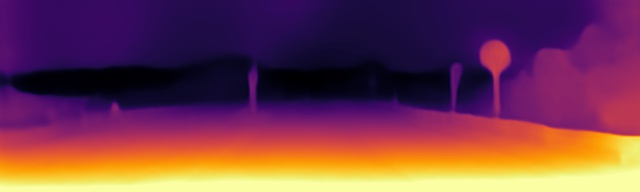} &
            \includegraphics[width=\turnwidthnew]{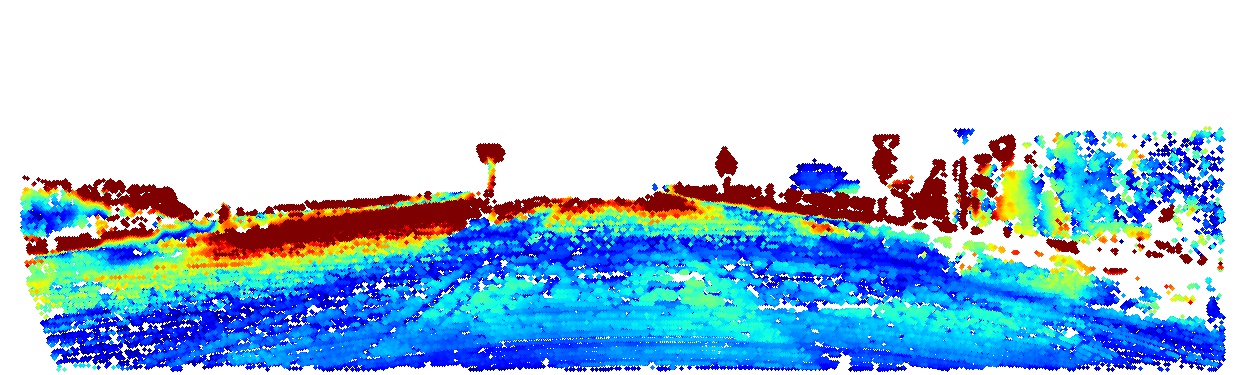} \\
            Monodepth2 \cite{godard2019digging} &
            \includegraphics[width=\turnwidthnew]{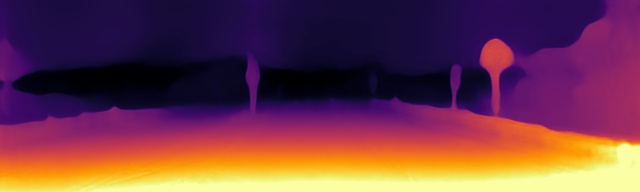} &
            \includegraphics[width=\turnwidthnew]{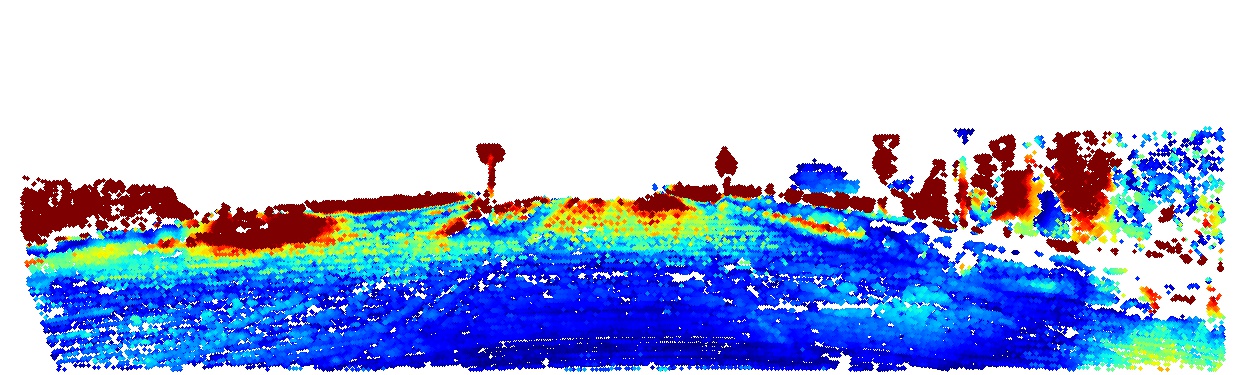} \\
            DFTP \cite{patil2020dont} &
            \includegraphics[width=\turnwidthnew]{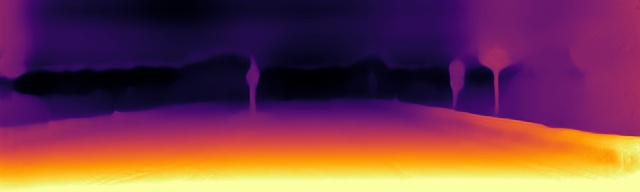} &
            \includegraphics[width=\turnwidthnew]{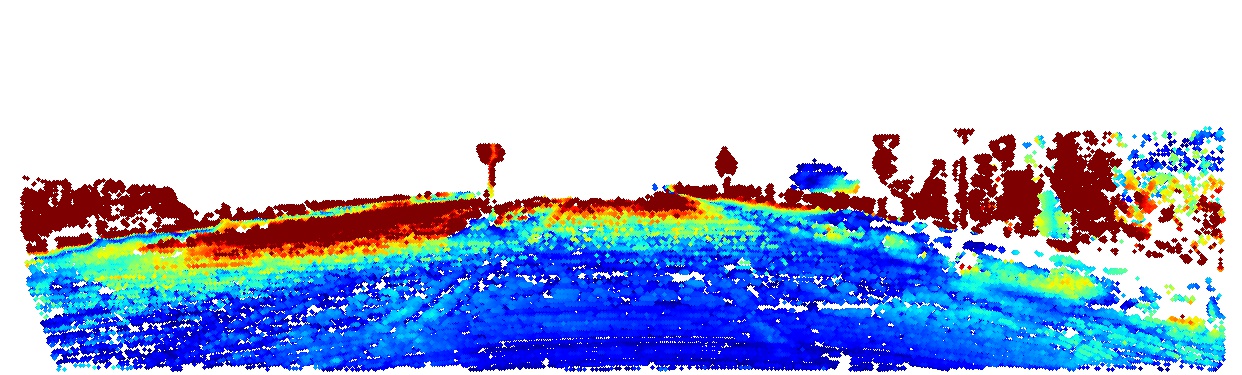} \\
            Packnet-SFM \cite{guizilini20203d} &
            \includegraphics[width=\turnwidthnew]{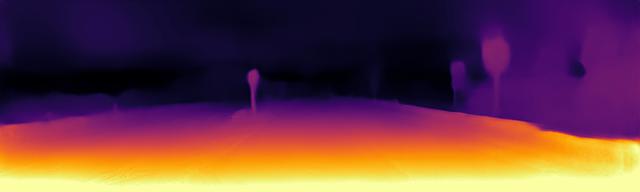} &
            \includegraphics[width=\turnwidthnew]{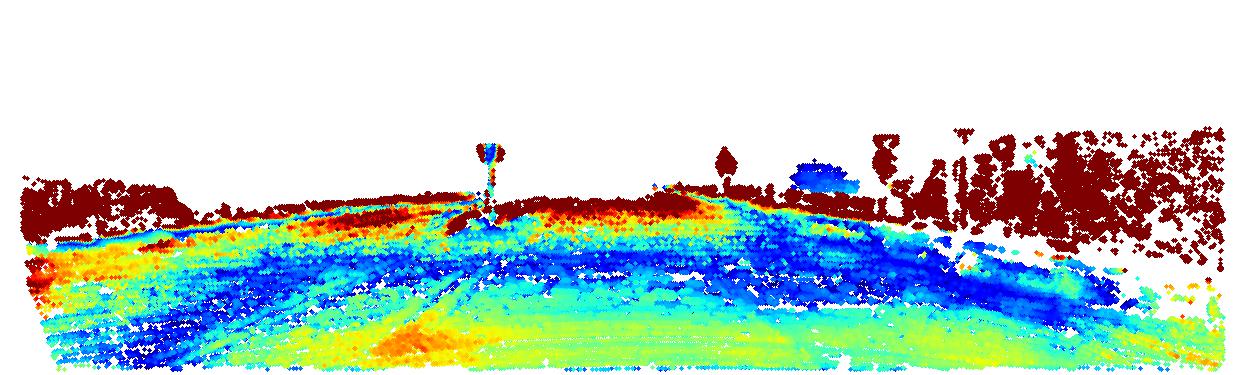} \\
            
            Input and ground truth &
            \includegraphics[width=\turnwidthnew]{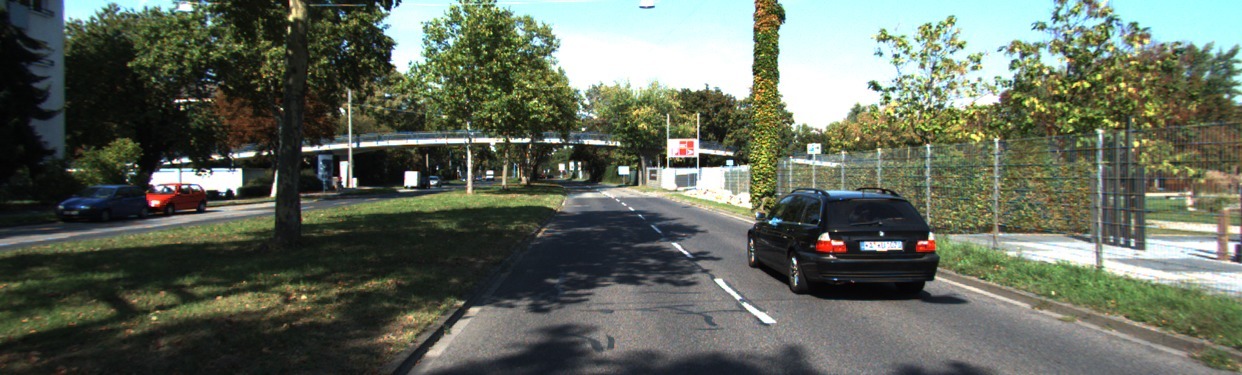} &
            \includegraphics[width=\turnwidthnew]{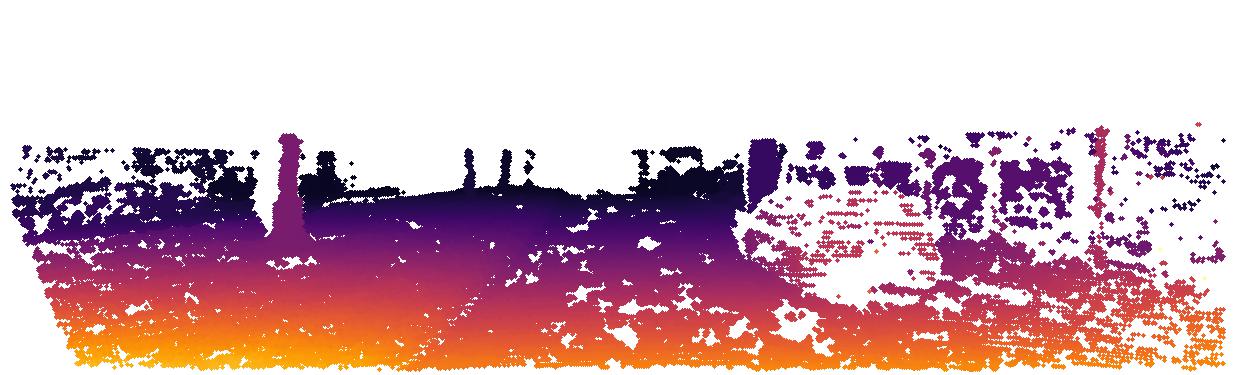} \\
            ManyDepth (MR) &
            \includegraphics[width=\turnwidthnew]{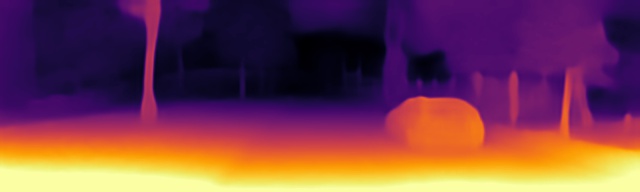} &
            \includegraphics[width=\turnwidthnew]{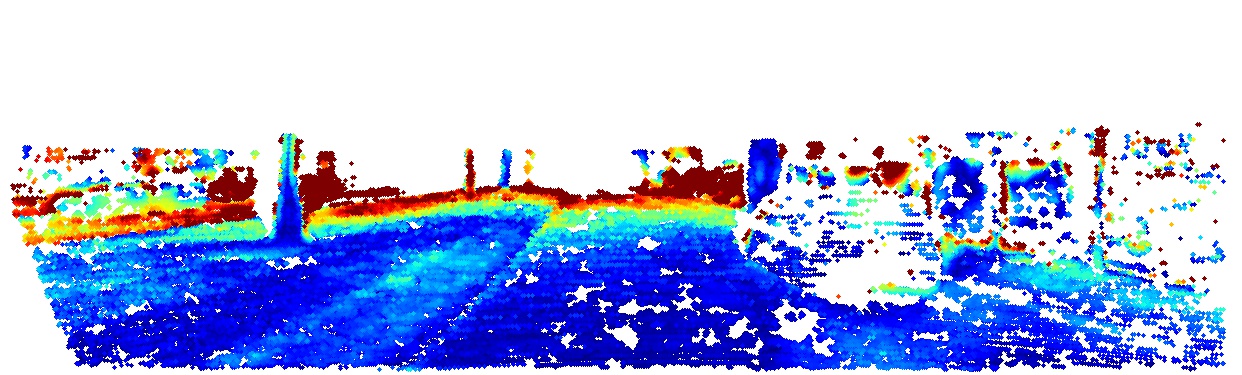} \\
            Monodepth2 \cite{godard2019digging} &
            \includegraphics[width=\turnwidthnew]{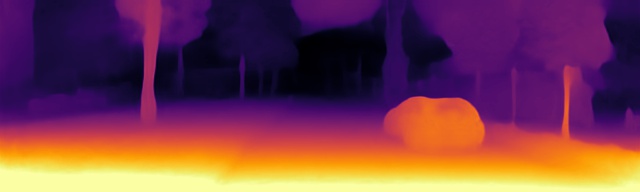} &
            \includegraphics[width=\turnwidthnew]{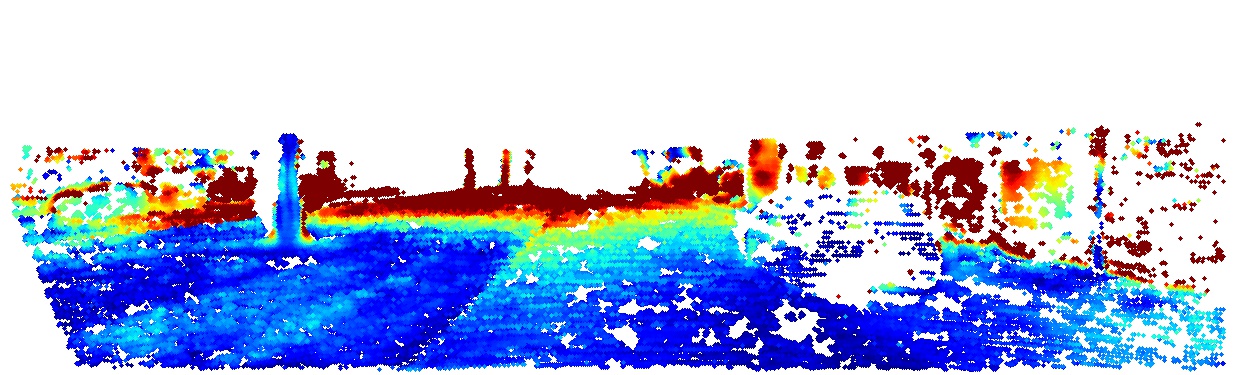} \\
            DFTP \cite{patil2020dont} &
            \includegraphics[width=\turnwidthnew]{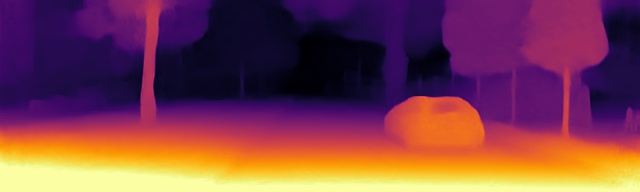} &
            \includegraphics[width=\turnwidthnew]{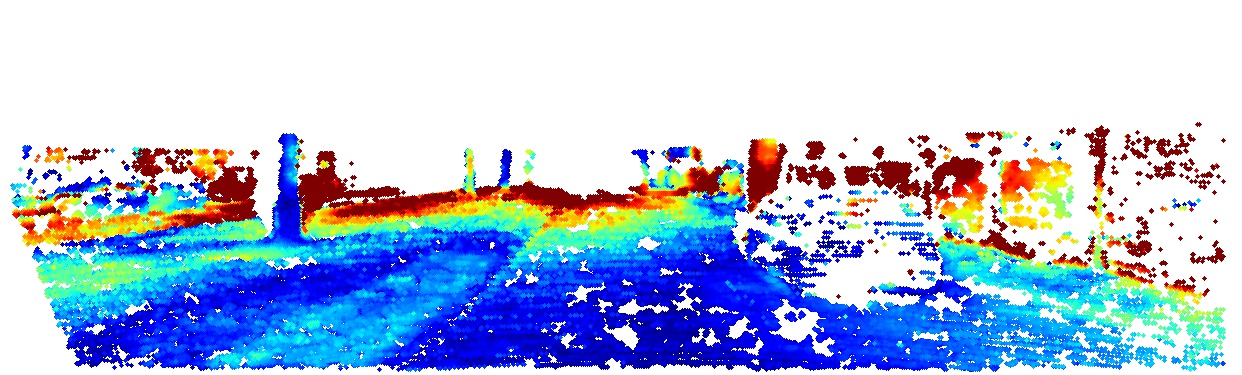} \\
            Packnet-SFM \cite{guizilini20203d} &
            \includegraphics[width=\turnwidthnew]{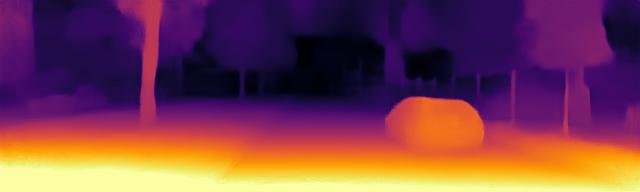} &
            \includegraphics[width=\turnwidthnew]{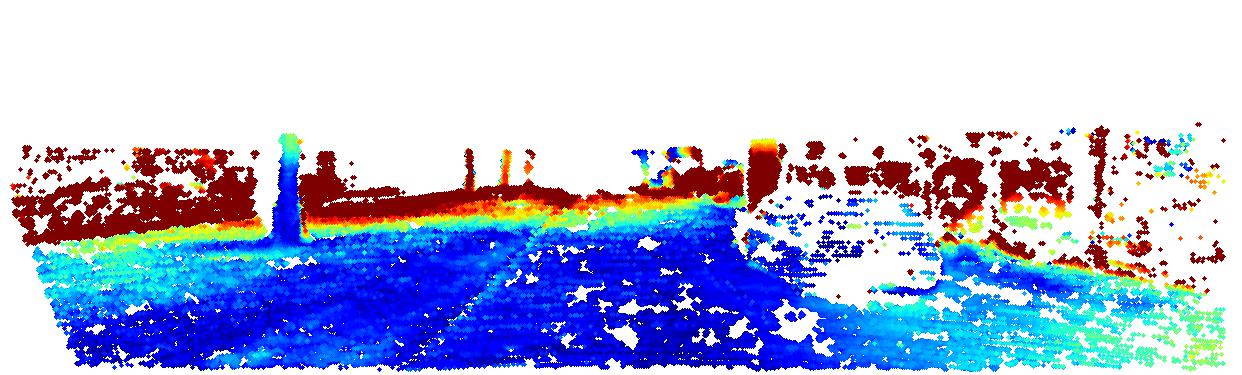} \\

        \end{tabular}
        }
        \caption{\textbf{Further qualitative results on the KITTI dataset (part 2)}. Error maps on the right measure the absolute relative error compared to the ground truth, after median scaling \cite{eigen2015predicting}. Errors range from blue (low error, abs.~rel.~$=0.0$) to red (high error, abs.~rel.~$=0.2$). The color mapping in all error maps are the same, and the colorbar is shown in Figure~\ref{fig:colormap}. \label{fig:extra_kitti_qual_2}}
\end{figure*}

\begin{figure*}
    \resizebox{1.0\textwidth}{!}{
        \begin{tabular}{m{3.5cm}m{6cm}m{6cm}}

            Input and ground truth &
            \includegraphics[width=\turnwidthnew]{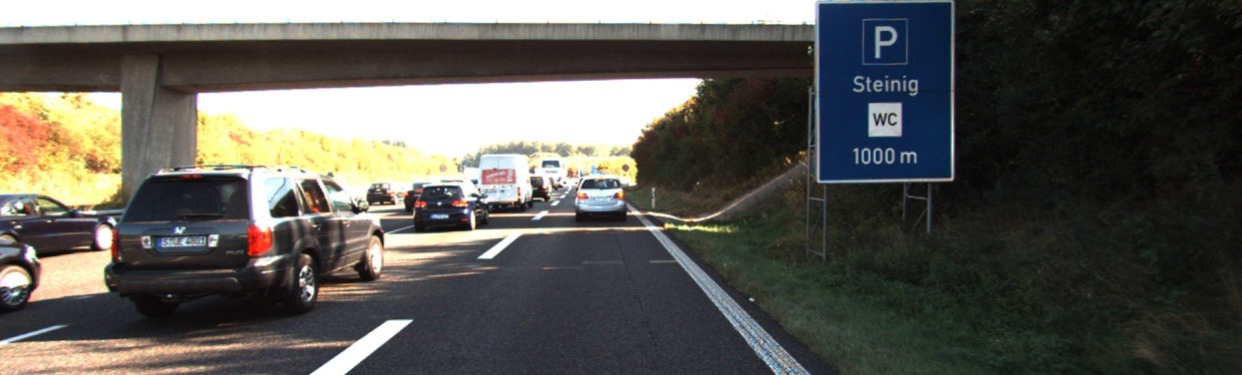} &
            \includegraphics[width=\turnwidthnew]{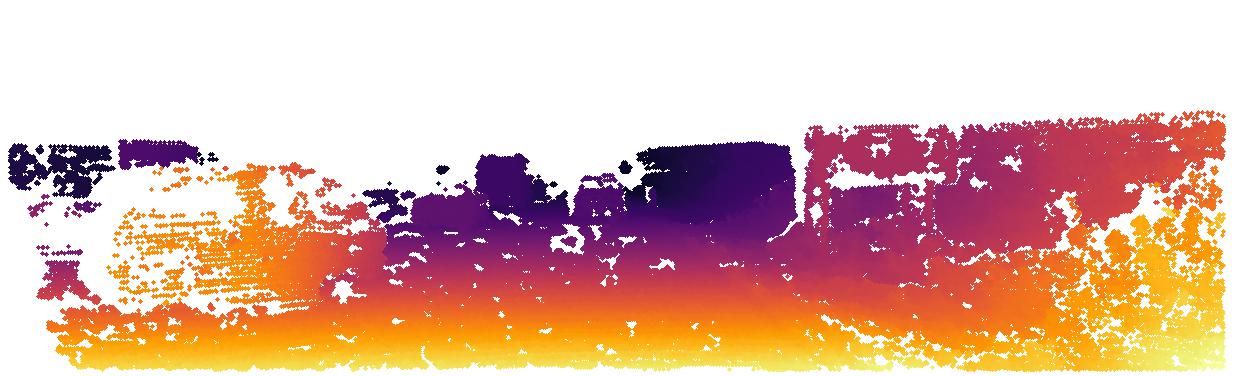} \\
            ManyDepth (MR) &
            \includegraphics[width=\turnwidthnew]{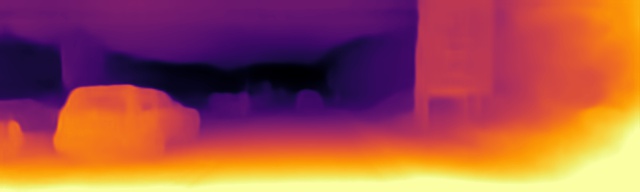} &
            \includegraphics[width=\turnwidthnew]{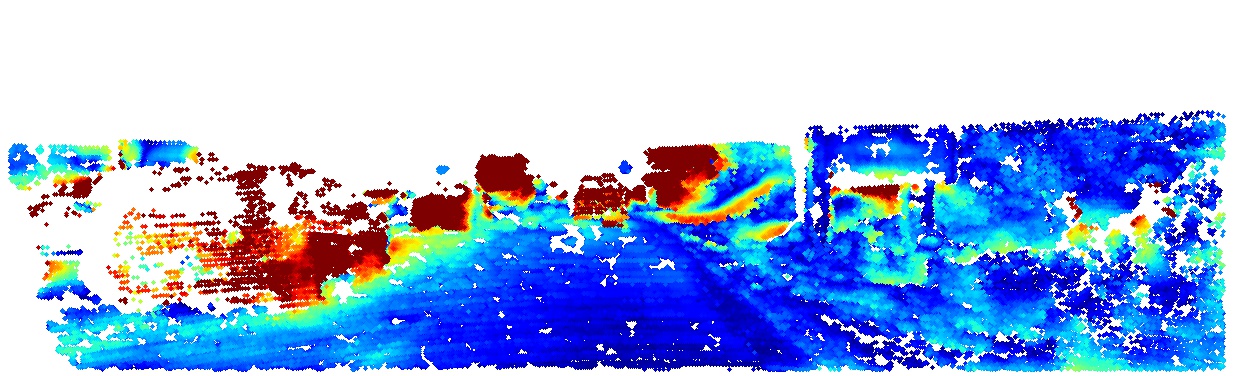} \\
            Monodepth2 \cite{godard2019digging} &
            \includegraphics[width=\turnwidthnew]{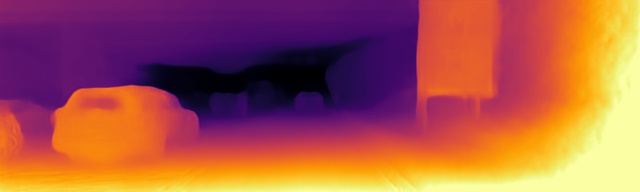} &
            \includegraphics[width=\turnwidthnew]{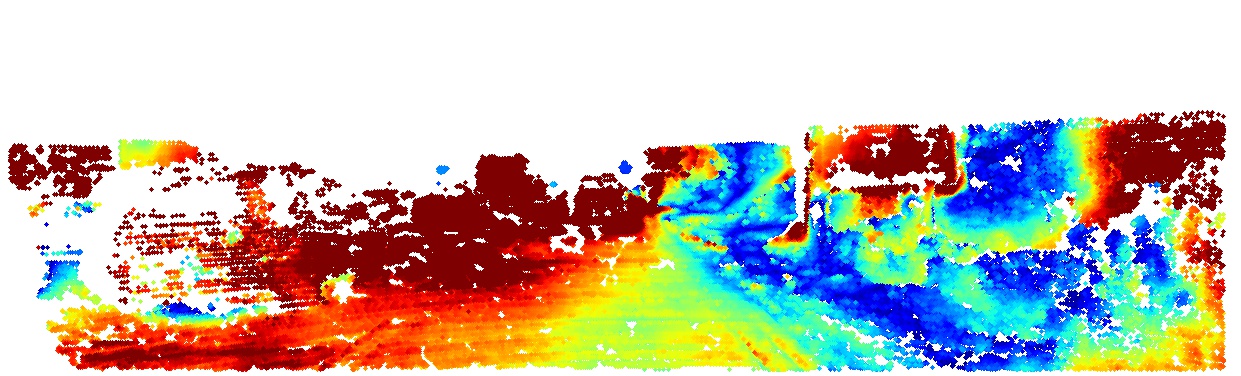} \\
            DFTP \cite{patil2020dont} &
            \includegraphics[width=\turnwidthnew]{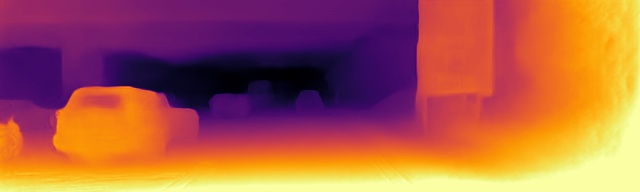} &
            \includegraphics[width=\turnwidthnew]{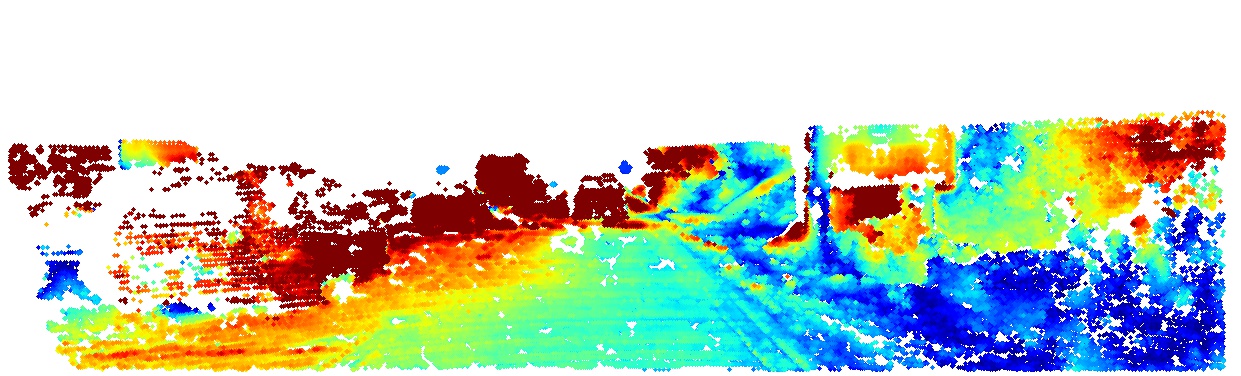} \\
            Packnet-SFM \cite{guizilini20203d} &
            \includegraphics[width=\turnwidthnew]{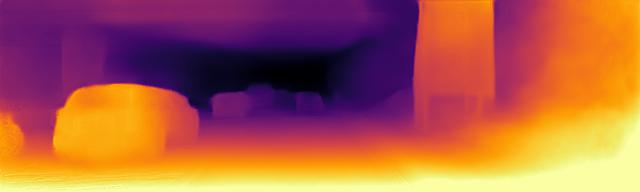} &
            \includegraphics[width=\turnwidthnew]{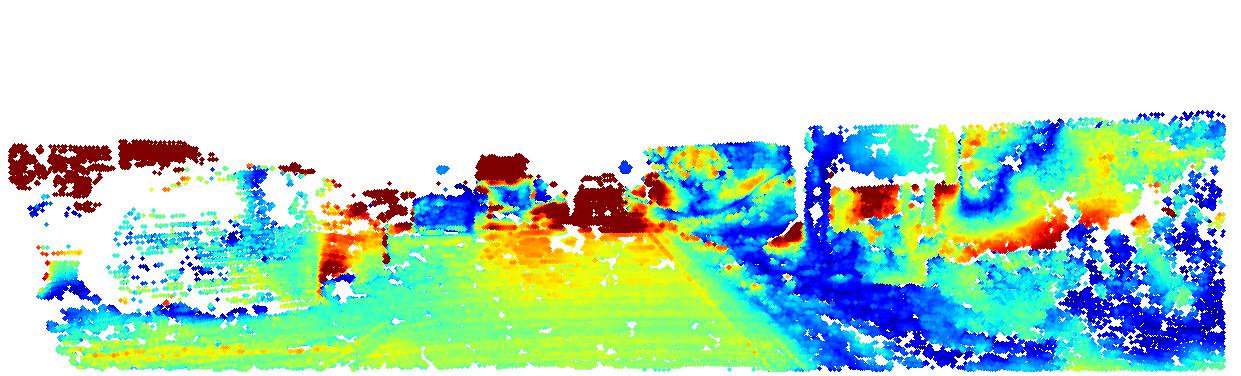} \\
    
        \end{tabular}
        }
        \caption{\textbf{Further qualitative results on the KITTI dataset (part 3)}. Error maps on the right measure the absolute relative error compared to the ground truth, after median scaling \cite{eigen2015predicting}. Errors range from blue (low error, abs.~rel.~$=0.0$) to red (high error, abs.~rel.~$=0.2$). The color mapping in all error maps are the same, and the colorbar is shown in Figure~\ref{fig:colormap}.  \label{fig:extra_kitti_qual_3}}
\end{figure*}

\end{appendices}


\end{document}